%% file: AIJ-SituatedConditionals.tex
\documentclass[review]{elsarticle}

\usepackage{lineno,hyperref}
\modulolinenumbers[5]

\journal{Journal of \LaTeX\ Templates}

\usepackage[utf8]{inputenc}
\usepackage[T1]{fontenc}

\usepackage[ruled,linesnumbered,vlined,scleft]{algorithm2e}
\usepackage{amsthm}
\usepackage{amsmath}
\usepackage{amssymb}
\usepackage{color}
\usepackage[svgnames]{xcolor} 
\usepackage{comment}
\usepackage{mathrsfs} 
\usepackage{url}

\usepackage[thinlines]{easytable}
\usepackage{latexsym}
\usepackage{stmaryrd}
\usepackage{subfigure}
\usepackage{thmtools}

\newcommand{\revised}[1]{{\color{black}{#1}}}

\input{AIJ-SituatedConditionals-Macros}

\begin{document}

\begin{frontmatter}

\title{Situated Conditional Reasoning\tnoteref{mytitlenote}}
\tnotetext[mytitlenote]{This article is an extended and elaborated version of a paper presented at the 35th AAAI Conference on Artificial Intelligence (AAAI 2021)~\cite{CasiniEtAl2021-AAAI}.}

\author[Pisa,CapeTown]{Giovanni Casini}
\author[CapeTown,Pisa]{Thomas Meyer}
\author[Paris,CapeTown,Pisa]{Ivan Varzinczak}

\address[Pisa]{CNR--ISTI, Italy}
\address[CapeTown]{CAIR, University of Cape Town, South Africa}
\address[Paris]{LIASD, Université Paris~8, France}

\begin{abstract}
Conditionals are useful for modelling many forms of everyday human reasoning but are not always sufficiently expressive to represent the information we want to reason about. In this paper, we make a case for a form of \emph{situated} conditional. By `situated', we mean that there is a context, based on an agent's beliefs and expectations, that works as background information in evaluating a conditional\revised{, and we allow such a context to vary}. These conditionals are able to  distinguish, for example, between \emph{expectations} and \emph{counterfactuals}. Formally, they are shown to generalise the conditional setting in the style of Kraus, Lehmann, and Magidor. We show that situated conditionals can be described in terms of a set of rationality postulates. We then propose an intuitive semantics for these conditionals and present a representation result which shows that our semantic construction corresponds exactly to the description in terms of postulates. With the semantics in place, we define a form of entailment for situated conditional knowledge bases, which we refer to as \emph{minimal closure}. 
Finally, we proceed to show that it is possible to reduce the computation of minimal closure to a series of propositional entailment and satisfiability checks. While this is also the case for rational closure, it is somewhat surprising that the result carries over to minimal closure. 
\end{abstract}


\end{frontmatter}


\section{Introduction}\label{Introduction}

Conditionals are at the heart of human everyday reasoning and play an important role in the logical formalisation of reasoning. They can usually be interpreted in many ways: as necessity~\cite{Kratzer1979,Bouchon-Meunier2002}, as presumption~\cite{KrausEtAl1990,LehmannMagidor1992,BOUTILIER199487}, normative~\cite{Hansson1969,MakinsonVandertorre2000}, causal~\cite{BellerKuhnmunch07,GIORDANO2004239}, probabilistic~\cite{SCHURZ199881,SNOW1999269,HawthorneM07}, counterfactual~\cite{sep-counterfactuals,Lewis1973}, and many others. Two very common interpretations that are also strongly interconnected are conditionals representing \emph{expectations} (`If it is a bird, then presumably it  flies') and conditionals representing \emph{counterfactuals} (`If Napoleon had won at Waterloo, the whole of Europe would be speaking French'). Although they are connected by virtue of being conditionals, the types of reasoning they aim to model differ somewhat. For instance, in the first example above, the premise of the conditional is consistent with what is believed, while in the second example, the premise is inconsistent with an agent's beliefs. That this point is problematic can be made concrete with an extended version of the (admittedly overused) penguin example.

\begin{example}\label{Ex:Introduction}
Suppose we know that birds usually fly, that penguins are birds that usually do not fly, that dodos were birds that usually did not fly, and that dodos do not exist anymore. As outlined in more detail in Example~\ref{Ex:PenguinDodo1} later on, the standard preferential semantic approach to representing conditionals~\cite{LehmannMagidor1992} is limited in that it allows for two forms of representation of an agent's beliefs. On the one, it would be impossible to distinguish between atypical (exceptional) entities, such as penguins, and non-existing entities, such as dodos (they are equally exceptional). On the other, it would be possible to draw this type of distinction, but at the expense of being unable to reason coherently about counterfactuals---the agent would be forced to conclude anything from the (nowadays absurd) existence of dodos.  
\end{example}

By `reasoning coherently' about counterfactuals, we mean being able to derive new information in a non-trivial way, formalising reasoning patterns  that can be recognised as `rational'. As mentioned in the above example and formally exemplified in Example~\ref{Ex:PenguinDodo1} later on, using the classical semantic solutions to reason simultaneously about what is actually plausible and what is counterfactually plausible has some strong limitations. For example, the formalism we use could  force us to reason about counterfactuals in a trivial way, relying on the \emph{ex falso quodlibet} principle. In this work, we introduce a logic of \emph{situated} conditionals to overcome precisely this problem. By `situated', we mean that there is a context, based on an agent's beliefs and expectations, which works as background information in evaluating a conditional. This is particularly important when we consider counterfactual conditionals. The central insight is that adding an explicit notion of situation to standard conditionals allows for a refined semantics of this enriched language in which the problems described in Example \ref{Ex:Introduction} can be dealt with adequately. It also allows us to reason coherently with counterfactual conditionals such as `Had Mauritius not been colonised, the dodo would not fly'\footnote{\revised{The extinction of dodos in the 17th century is considered to be a consequence of the colonisation of Mauritius.}}. That is, the premise of a counterfactual can be inconsistent with the agent's beliefs without lapsing into the triviality of the \emph{ex falso quodlibet}  principle. Moreover, it is possible to reason coherently with situated conditionals without knowing whether their premises are plausible or counterfactual. In the case of penguins and dodos, for example, it allows us to state that penguins usually do not fly in a situation where penguins exist and that dodos usually do not fly in a situation where dodos also exist while being unaware of whether or not penguins and dodos actually exist. At the same time, it remains possible to make classical statements specifying what necessarily holds (e.g., stating that penguins and dodos are birds as a necessary fact).

Counterfactual and defeasible reasoning have been important topics of research in knowledge representation and reasoning since the beginning of the AI endeavour \cite{BOBROW19801,GINSBERG198635}. Still, they have usually been formalised as the same form of conditional reasoning. While research on defeasible reasoning has always been quite active, the importance of counterfactuals in AI has become more apparent only recently, especially considering the role that counterfactuals play in causal reasoning \cite{Rott22} and in eXplainable AI (XAI)~\cite{WarrenEtAl2022,Byrne2019,DaiKSRB2022}. The increasing attention in the area of XAI to this topic means that a detailed formal analysis of counterfactuals and their associated reasoning systems is not just a timely research endeavour but a necessary one. In this context, we put forward a framework for managing both counterfactual reasoning and reasoning based on expectations---one that avoids some of the limitations associated with previous conditional approaches based on a preferential semantics.
\myskip

The remainder of the paper is organised as follows. Section~\ref{Preliminaries} outlines the formal background on propositional logic and on the preferential approach to conditionals on which our work is based. Section~\ref{CC} is the heart of the paper. It describes the language of situated conditionals, furnishes it with an appropriate and intuitive semantics, and motivates the  corresponding logic through examples, formal postulates, and a formal representation result. With the basics of the logic in place, Section~\ref{Entailment} defines a form of entailment for it that is based on the well-known notion of \emph{rational closure}~\cite{LehmannMagidor1992}. As such, it plays a role similar to the one that rational closure plays for reasoning with conditionals---it is a basic form of entailment on which other forms of logical consequence can be constructed. Section \ref{Algorithm} shows that, from a computational perspective, the version of entailment we propose in the previous section is reducible to classical propositional reasoning. Section~\ref{RelatedWork} reviews related work, while Section~\ref{Conclusion} concludes and considers future avenues to explore. Long proofs are in the appendix.

\section{Formal background}\label{Preliminaries}

In this paper, we assume a finite set of propositional \emph{atoms}~$\Prp$ and use $p,q,\ldots$ as metavariables to denote its elements. Sentences of the underlying propositional language are denoted by $\alpha,\beta,\ldots$, and are built up from the atomic propositions and the standard Boolean connectives in the usual way. The set of all propositional sentences is denoted by~$\Lang$.

A \emph{valuation} (alias \emph{world}) is a function from~$\Prp$ into~$\{0,1\}$. The set of all valuations is denoted~$\U$, and we use $u,v,\ldots$ to denote its elements. Whenever it eases presentation, we represent valuations as sequences of atoms (\eg, $p$) and barred atoms (\eg, $\bar{p}$), with the usual understanding. As an example, if $\Prp=\{\bird,\flies,\peng\}$, with the atoms standing for, respectively, `being a $\bird$ird', `being a $\flies$lying creature', and `being a $\peng$enguin', then the valuation~$\bird\bar{\flies}\peng$ conveys the idea that~$\bird$ is true, $\flies$ is false, and~$\peng$ is true.

With $v\sat\alpha$ we denote the fact that the valuation~$v$ \emph{satisfies} the sentence~$\alpha$. Given~$\alpha\in\Lang$, with $\Mod{\alpha}\defined\{v\in\U\mid v\sat\alpha\}$ we denote its \emph{models}. For $X\subseteq\Lang$, $\Mod{X}\defined\bigcap_{\alpha\in X}\Mod{\alpha}$. We say $X\subseteq\Lang$ (classically) \emph{entails} $\alpha\in\Lang$, denoted $X\entails\alpha$, if $\Mod{X}\subseteq\Mod{\alpha}$. Given a set of valuations $V$, $\fm{V}$ indicates a sentence characterising the set~$V$. That is, $\fm{V}$ is a propositional sentence satisfied by all, and only, the valuations in~$V$.

\subsection{KLM-style rational defeasible consequence}\label{KLM}

A \emph{defeasible consequence relation} $\twiddle$ is a binary relation on~$\Lang$. Intuitively, the fact that $(\alpha,\beta)\in\twiddle$, which is usually represented as the statement~$\alpha\twiddle\beta$, captures the idea that~``$\beta$ is a defeasible consequence of~$\alpha$'', or, in other words, that ``if~$\alpha$, then usually (alias normally, or typically)~$\beta$''. The relation~$\twiddle$ is said to be \emph{preferential}~\cite{KrausEtAl1990} if it satisfies the well-known~KLM preferential postulates below:
\[
\begin{array}{clccl}
(\text{Ref}) & \alpha \twiddle \alpha & \quad & (\text{\small LLE}) & {\displaystyle \frac{\entails \alpha\liff \beta,\ \alpha \twiddle \gamma}{\beta \twiddle \gamma}} \\[0.5cm]
(\text{And}) & {\displaystyle \frac{\alpha \twiddle \beta, \ \alpha \twiddle \gamma}{\alpha \twiddle \beta\land\gamma}} & \quad & (\text{Or}) & {\displaystyle \frac{\alpha \twiddle \gamma, \ \beta \twiddle \gamma}{\alpha \lor \beta \twiddle \gamma}} \\[0.5cm]
(\text{\small RW}) & {\displaystyle \frac{\alpha \twiddle \beta,
\ \entails \beta \limp \gamma}{\alpha \twiddle \gamma}} & \quad & 
(\text{\small CM}) & 
{\displaystyle \frac{\alpha \twiddle \beta, \ \alpha \twiddle \gamma}
{\alpha\land\gamma \twiddle \beta}} 
\end{array}
\]

If on top of the preferential postulates the relation~$\twiddle$ also satisfies the following rational monotonicity postulate, then~$\twiddle$ is said to be \emph{rational}:
\[
\begin{array}{cl}
(\text{\small RM}) & 
{\displaystyle \frac{\alpha \twiddle \beta, \ \alpha \ntwiddle \lnot\gamma}
{\alpha\land\gamma \twiddle \beta}}
\end{array}
\]

The merits of these postulates have been addressed extensively in the literature~\cite{KrausEtAl1990,Gabbay84}, and we shall not repeat them here.

A suitable semantics for rational consequence relations is provided by ordered structures called \emph{ranked interpretations} (alternative semantics have also been proposed in the literature, in particular, Spohn's ordinal conditional functions~\cite{Spohn1988}).

\begin{definition}[Ranked Interpretation]\label{Def:RankedInterpretation}
A \df{ranked interpretation}~$\RM$ is a total function from~$\U$ to $\mathbb{N}\cup\{\infty\}$, satisfying the following \df{convexity property}: for every~$u\in\U$ and every $i\in\mathbb{N}$, if $\RM(u)=i$, then, for every~$j$ \st\ $0\leq j<i$, there is a $u'\in\U$ for which $\RM(u')=j$.
\end{definition}

For a given ranked interpretation~$\RM$ and valuation~$v$, we denote with~$\RM(v)$ the \emph{rank of}~$v$. The number~$\RM(v)$ indicates the degree of \emph{atypicality} of~$v$. So the valuations judged most typical are those with rank~0, while those with an infinite rank are deemed so atypical as to be seen as implausible. We can therefore partition the set~$\U$ \wrt~$\RM$ into the set of \emph{plausible} valuations~$\U^{\f}_{\RM}\defined\{u\in\U\mid \RM(u)\in\mathbb{N}\}$, and \emph{implausible} valuations $\U^{\infty}_{\RM}\defined\U\setminus\U^{\f}_{\RM}$. It goes without saying that $\U^{\f}_{\RM}$ or $\U^{\infty}_{\RM}$ (but not both) can be empty. (Throughout the paper, we shall use the symbol~$\f$ to refer to finiteness.) With $\states{i}_{\RM}$, for $i\in\mathbb{N}\cup\{\infty\}$, we indicate all the valuations with rank~$i$ in~$\RM$ (we omit the subscript whenever it is clear from the context). 

Assuming $\Prp=\{\bird,\flies,\peng\}$, with the intuitions as above, Figure~\ref{Figure:RankedInterpretation} below shows an example of a ranked interpretation.

\begin{figure}[ht]
\begin{center}
\begin{TAB}(r,1cm,0.2cm)[3pt]{|c|c|}{|c|c|c|c|}%
 {$\infty$} & {$\peng\bar{\bird}\hspace{0.02cm}\bar{\flies}$ \quad $\peng\bar{\bird}\flies$} \\
 {$2$} & {$\peng\bird\flies$} \\
 {$1$} & {$\bar{\peng}\bird\bar{\flies}$ \quad $\peng\bird\bar{\flies}$}\\ 
 {$0$} & {$\bar{\peng}\bar{\bird}\hspace{0.02cm}\bar{\flies}$ \quad $\bar{\peng}\bar{\bird}\flies$ \quad $\bar{\peng}\bird\flies$} \\
\end{TAB}
\end{center}
\caption{A ranked interpretation for~$\Prp=\{\bird,\flies,\peng\}$ with both $\U^{\f}_{\RM}$ and $\U^{\infty}_{\RM}$ nonempty.}
\label{Figure:RankedInterpretation}
\end{figure}

Let~$\RM$ be a ranked interpretation and let~$\alpha\in\Lang$. Then $\states{\alpha}^{\f}_{\RM}\defined\U^{\f}_{\RM}\cap\states{\alpha}$, and  $\min\states{\alpha}^{\f}_{\RM}\defined\{u\in\states{\alpha}^{\f}_{\RM} \mid \RM(u)\leq\RM(v)$, for all $v\in\states{\alpha}^{\f}_{\RM}\}$. A defeasible consequence relation $\alpha\twiddle\beta$ can be given an intuitive semantics in terms of ranked interpretations as follows: $\alpha\twiddle\beta$ is \emph{satisfied in}~$\RM$ (denoted $\RM\sat\alpha\twiddle\beta$) if $\min\Mod{\alpha}^{\f}_{\RM}\subseteq\Mod{\beta}$, with $\RM$ referred to as a \emph{ranked model} of~$\alpha\twiddle\beta$. In the example in Figure~\ref{Figure:RankedInterpretation}, we have $\RM\sat\bird\twiddle\flies$,  $\RM\sat\lnot(\peng\limp\bird)\twiddle\bot$, $\RM\sat\peng\twiddle\lnot\flies$, $\RM\nsat\flies\twiddle\bird$, and $\RM\sat\peng\land\lnot\bird\twiddle\bird$. It is easily verified that $\RM\sat\lnot\alpha\twiddle\bot$ if and only if $\U^{\f}_{\RM}\subseteq\states{\alpha}$. Hence we frequently abbreviate $\lnot\alpha\twiddle\bot$~as~$\alpha$. Two defeasible statements $\alpha\twiddle\beta$ and $\gamma\twiddle\delta$ are said to be \emph{rank equivalent} if they have the same ranked models, \ie, if for every ranked interpretation~$\RM$, $\RM\sat\alpha\twiddle\beta$ if and only if~$\RM\sat\gamma\twiddle\delta$.

The correspondence between rational consequence relations and ranked interpretations is formalised by the following representation result.

\begin{theorem}[Lehmann \& Magidor~\cite{LehmannMagidor1992}; G\"{a}rdenfors \& Makinson~\cite{GardenforsMakinson1994}]\label{Theorem:RepresentationResult}
A defeasible consequence relation~$\twiddle$ is rational iff there is a ranked interpretation~$\RM$ such that\revised{, for every pair of formulae $\alpha$ and $\beta$,} $\alpha\twiddle\beta$ iff $\RM\sat\alpha\twiddle\beta$.
\end{theorem}

\subsection{Rational closure}\label{RationalClosure}

It is possible to represent knowledge as a set of defeasible statements and to use such a set to infer other defeasible statements from it. This is the stance adopted by Lehmann and Magidor~\cite{LehmannMagidor1992}. A \emph{conditional knowledge base}~$\C$ is a finite set of defeasible statements of the form $\alpha\twiddle\beta$, with $\alpha,\beta\in\Lang$. As before, in knowledge bases, we shall also abbreviate $\lnot\alpha\twiddle\bot$ with~$\alpha$. As an example, let $\C=\{\bird\twiddle\flies,\peng\limp\bird,\peng\twiddle\lnot\flies\}$. 

Given a conditional knowledge base~$\C$, a \emph{ranked model} of~$\C$ is a ranked interpretation satisfying all statements in~$\C$. As it turns out, the ranked interpretation in Figure~\ref{Figure:RankedInterpretation} is a ranked model of the above~$\C$. It is not hard to see that, in every ranked model of~$\C$, the valuations $\bar{\bird}\hspace{0.02cm}\bar{\flies}\peng$ and $\bar{\bird}\flies\peng$ are deemed implausible---note, however, that they are still possible from a logical point of view, which is the reason why they feature in all ranked interpretations.

A conditional knowledge base $\C$ is \emph{consistent} if it has a ranked model $\RM$ \st\ $\states{0}_{\RM}\neq\emptyset$. That is, $\C$ is consistent if it has a ranked model~$\RM$ that does not satisfy~$\top\twiddle\bot$. Two conditional knowledge bases are \emph{rank equivalent} if they have exactly the same ranked models. 

An important reasoning task in this setting is determining which conditionals follow from a conditional knowledge base. Of course, even when interpreted as a conditional in (and under) a given knowledge base~$\C$, $\twiddle$ is expected to adhere to the postulates of Section~\ref{KLM}. Intuitively, that means whenever appropriate instantiations of the premises in a postulate are sanctioned by~$\C$, so should the suitable instantiation of its conclusion.

To be more precise, we can take the defeasible conditionals in~$\C$ as the core elements of a defeasible consequence relation~$\twiddle^{\C}$. By closing the latter under the preferential postulates (in the sense of exhaustively applying them as rules), we get a \emph{preferential extension} of~$\twiddle^{\C}$. Since there can be more than one such extension, the most cautious approach consists in taking their intersection. The resulting set, which also happens to be closed under the preferential postulates, is the \emph{preferential closure} of~$\twiddle^{\C}$, which we denote by~$\twiddle^{\C}_{PC}$. It turns out that the preferential closure of~$\twiddle^{\C}$ contains exactly the conditionals entailed by~$\C$. (Hence, the notions of closure of and entailment from a conditional knowledge base are two sides of the same coin.)

The same process and definitions from above carry over when one requires the defeasible consequence relations also to be closed under the rule~RM, in which case we talk of \emph{rational} extensions of~$\twiddle^{\C}$. Nevertheless, as pointed out by Lehmann and Magidor~\cite[Section~4.2]{LehmannMagidor1992}, the intersection of all such rational extensions does not generally yield a rational consequence relation: it coincides with preferential closure and, therefore, may fail~RM. Among other things, this means that the corresponding entailment relation, which is called \emph{rank entailment} and defined as $\C\entails_{\RM}\alpha\twiddle\beta$ if every ranked model of~$\C$ also satisfies $\alpha\twiddle\beta$, is \emph{monotonic} (in that it is defined as a standard Tarskian entailment relation). Therefore rank entailment falls short of being a suitable form of entailment in a defeasible reasoning setting. As a result, several alternative notions of entailment from conditional knowledge bases have been explored in the literature on non-monotonic reasoning~\cite{Lehmann1995,BoothParis1998,Weydert03,GiordanoEtAl2012,GiordanoEtAl2015,BoothEtAl2019,CasiniEtAl2019-JELIA}, with \emph{rational closure}~\cite{LehmannMagidor1992} commonly acknowledged as the `gold standard' in the matter.

Rational closure (RC) is a form of inferential closure extending the notion of rank entailment above. It formalises the principle of \emph{presumption of typicality}~\cite[p.~63]{Lehmann1995}, which, informally, specifies that a situation (in our case, a valuation) should be assumed to be as typical as possible (\wrt\ background information in a knowledge base).

Multiple equivalent characterisations of RC have been proposed~\cite{LehmannMagidor1992, Pearl1990,BoothParis1998,HillParis2003,BritzEtAl2020}, and here we rely on the one by Giordano and others \cite{GiordanoEtAl2015}.  Assume an ordering~$\preceq_{\C}$ on all ranked models of a knowledge base~$\C$, which is defined as follows: $\RM_{1}\preceq_{\C}\RM_{2}$, if, for every $v\in\U$, $\RM_{1}(v)\leq\RM_{2}(v)$. Intuitively, ranked models lower down in the ordering correspond to descriptions of the world in which the typicality of each situation (valuation) is maximised. It is easy to see that~$\preceq_{\C}$ is a weak partial order. Giordano~\etal.~\cite{GiordanoEtAl2015} showed that there is a unique $\preceq_{\C}$-minimal element. The rational closure of~$\C$ is defined in terms of this minimum ranked model of~$\C$.

\begin{definition}[Rational Closure]\label{Def:RC}
Let $\C$ be a conditional knowledge base, and let $\RM^{\C}_{RC}$ be the minimum element of~$\preceq_{\C}$ on ranked models of~$\C$. The \df{rational closure} of~$\C$ is the defeasible consequence relation $\twiddle^{\C}_{RC}\defined\{\alpha\twiddle\beta \mid \RM^{\C}_{RC}\sat\alpha\twiddle\beta\}$.
\end{definition}

As an example, Figure~\ref{Figure:RankedInterpretation} shows the minimum ranked model of $\C=\{\bird\twiddle\flies,\peng\limp\bird,\peng\twiddle\lnot\flies\}$ \wrt~$\preceq_{\C}$. Hence we have that $\lnot\flies\twiddle\lnot\bird$ is in the rational closure of~$\C$ (but note it is not in the preferential closure of~$\C$).
\myskip

Observe that there are two levels of typicality at work for rational closure, namely \emph{within} ranked models of~$\C$, where valuations lower down are viewed as more typical, but also \emph{between} ranked models of~$\C$, where ranked models lower down in the ordering are viewed as more typical. The most typical ranked model~$\RM^{\C}_{RC}$ is the one in which valuations are as typical as $\C$~allows them to be (the principle of presumption of typicality we alluded to above).

Rational closure is commonly viewed as the \emph{basic} (although certainly 
not the only acceptable) form of non-monotonic entailment, on which other, more venturous forms can be and have been constructed~\cite{Lehmann1995,Kern-Isberner2001,CasiniEtAl2014,BoothEtAl2019,CasiniEtAl2019-JELIA}.

\section{Situated conditionals}\label{CC}

We now turn to the heart of the paper, the introduction of a logic-based formalism for the specification of and reasoning with situated conditionals. For a more detailed motivation, let us consider a more technical version of the penguin-dodo example introduced in Section~\ref{Introduction}.

\begin{example}\label{Ex:PenguinDodo1}
We know that birds usually fly ($\bird\twiddle\flies$), and that penguins are birds ($\peng\rightarrow\bird$
) that usually do not fly ($\peng\twiddle\lnot\flies$). Also, we know that dodos were birds ($\dodo\rightarrow\bird$
) that usually did not fly ($\dodo\twiddle\lnot\flies$), and that dodos do not exist anymore. Using the standard ranked semantics (Definition~\ref{Def:RankedInterpretation}), we have two ways of modelling the information above. 

The first option is to formalise what an agent believes by referring to the valuations with rank~$0$ in a ranked interpretation. That is, the agent believes~$\alpha$ is true if and only if $\top\twiddle\alpha$ holds. In such a case, $\top\twiddle\lnot\dodo$ means that the agent believes that dodos do not exist. The minimal model for this conditional knowledge base is shown in Figure~\ref{Figure:PenguinDodo1} (left). The main limitation of this representation is that all exceptional entities have the same status as dodos since they cannot be satisfied at rank~$0$. Hence, one of the consequences of the agent's beliefs is the statement $\top\twiddle\lnot\peng$, just as we have $\top\twiddle\lnot\dodo$, and, as a result, we are not able to distinguish between the status of the dodos (they do not exist anymore) and the status of the penguins (they do exist and are simply exceptional birds).

The second option is to represent what an agent believes in terms of all valuations with finite ranks. That is, an agent believes~$\alpha$ to hold if and only if $\lnot\alpha\twiddle\bot$ holds. If dodos do not exist, we add the statement $\dodo\twiddle\bot$. The minimal model for this case is depicted in Figure~\ref{Figure:PenguinDodo1} (right). Here we can distinguish between what is considered false (dodos exist) and what is exceptional (penguins), but we are unable to reason coherently about counterfactuals since from $\dodo\twiddle\bot$ we can conclude anything about dodos (via $\entails\bot\limp\alpha $ and RW, for any $\alpha\in\Lang$).
\end{example}

\begin{figure}[h]
\begin{minipage}{0.50\textwidth}
\begin{center}
\begin{TAB}(r,1cm,0.2cm)[3pt]{|c|c|}{|c|c|c|c|}%
{ $\infty$} & { $\U\setminus(\states{0}\cup\states{1}\cup\states{2}$) }  \\
{ $2$} & {$\peng\bar{\dodo}\bird\flies$\quad $\bar{\peng}\dodo\bird\flies$\quad $\peng\dodo\bird\flies$ }\\
{ $1$} & {$\bar{\peng}\bar{\dodo}\bird\bar{\flies}$\quad $\peng\bar{\dodo}\bird\bar{\flies}$\quad $\bar{\peng}\dodo\bird\bar{\flies}$\quad $\peng\dodo\bird\bar{\flies}$ }\\
{ $0$} & {$\bar{\peng}\bar{\dodo}\bird\flies$\quad $\bar{\peng}\bar{\dodo}\hspace{0.02cm}\bar{\bird}\flies$}\quad {$\bar{\peng}\bar{\dodo}\hspace{0.02cm}\bar{\bird}\hspace{0.02cm}\bar{\flies}$} \\
\end{TAB}
\end{center}
\end{minipage}\quad\begin{minipage}{0.50\textwidth}
\begin{center}
\begin{TAB}(r,1cm,0.2cm)[3pt]{|c|c|}{|c|c|c|c|}%
{ $\infty$} & { $\U\setminus(\states{0}\cup\states{1}\cup\states{2}$) }  \\
{ $2$} & {$\peng\bar{\dodo}\bird\flies$ }\\
{ $1$} & {$\bar{\peng}\bar{\dodo}\bird\bar{\flies}$\quad $\peng\bar{\dodo}\bird\bar{\flies}$\quad}\\
{ $0$} & {$\bar{\peng}\bar{\dodo}\bird\flies$\quad $\bar{\peng}\bar{\dodo}\hspace{0.02cm}\bar{\bird}\flies$}\quad {$\bar{\peng}\bar{\dodo}\hspace{0.02cm}\bar{\bird}\hspace{0.02cm}\bar{\flies}$} \\
\end{TAB}
\end{center}
\end{minipage}
\caption{Left: minimal ranked model of the KB in Example~\ref{Ex:PenguinDodo1} satisfying~$\top\twiddle\lnot\dodo$. Right: minimal ranked model of the KB expanded with~$\dodo\twiddle\bot$.}
\label{Figure:PenguinDodo1}
\end{figure}

A \emph{situated conditional} (SC for short) is a statement of the form $\alpha\twiddle_{\gamma}\beta$, with~$\alpha,\beta,\gamma\in\Lang$, which is read as `given the situation~$\gamma$, $\beta$ usually holds on condition that~$\alpha$ holds'. Formally, a situated conditional $\twiddle$ is a ternary relation on~$\Lang$. We shall write $\alpha\twiddle_{\gamma}\beta$ as an abbreviation for $\tuple{\alpha,\beta,\gamma}\in\ \twiddle$.  
To provide a suitable semantics for SCs, we define a refined version of the ranked interpretations of Section~\ref{Preliminaries} that we refer to as \emph{epistemic interpretations}. \revised{Following the provision of the semantics, we illustrate it with a representative example in Example~\ref{Ex:PenguinDodo2}}.

A ranked interpretation can differentiate between plausible valuations (those in $\U^{\f}_{\RM}$) but not between implausible ones (those in $\U^{\infty}_{\RM}$). In contrast, an epistemic interpretation can also tell implausible valuations apart. 
We thus distinguish between two classes of valuations: plausible valuations with a \emph{finite rank}, and 
implausible valuations with an \emph{infinite rank}. Within implausible valuations, we further distinguish between those considered as \emph{possible} and those that would be \emph{impossible}. This is formalised by assigning to each valuation $u$ a tuple of the form $\tuple{\f,i}$, where $i\in\mathbb{N}$, or $\tuple{\infty,i}$, where $i\in\mathbb{N}\cup\{\infty\}$. The~$\f$ in $\tuple{\f,i}$ is meant to indicate that~$u$ has a \emph{finite rank}, while the~$\infty$ in $\tuple{\infty,i}$ is intended to denote that~$u$ has an \emph{infinite rank}, where finite ranks are viewed as more typical than infinite ranks. Implausible valuations that are considered possible have an infinite rank $\tuple{\infty,i}$, where $i\in\mathbb{N}$, while those considered impossible have the infinite rank $\tuple{\infty,\infty}$, where $\tuple{\infty,\infty}$ is taken to be less expected than any of the other infinite ranks. 

To capture this formally, let $\R\defined\{\tuple{\f,i}\mid i\in\mathbb{N}\}\cup \{\tuple{\infty,i}\mid i\in\mathbb{N}\cup\{\infty\}\}$ denote henceforth the set of all possible~\emph{ranks}. We define the total ordering $\preceq$ over $\R$ as follows: $\tuple{x_1,y_1}\preceq\tuple{x_2,y_2}$ if $x_1=x_2$ and $y_1\leq y_2$, or $x_1=\f$ and $x_2=\infty$, where $i<\infty$ for all $i\in\mathbb{N}$. 

\begin{definition}[Epistemic Interpretation]\label{Def:EpistemicInterpretation}
An \df{epistemic interpretation}~$\EI$ is a total function from~$\U$ to~$\R$ for which the following \df{convexity} property holds: (\emph{i})~for every~$u\in\U$ and every~$i\in\mathbb{N}$, if $\EI(u)=\tuple{\f,i}$, then, for all~$j$ \st~$0\leq j<i$, there is a $u_j\in\U$ \st~$\EI(u_j)=\tuple{\f,j}$, and (\emph{ii})~for every~$u\in\U$ and every~$i\in\mathbb{N}$, if $\EI(u)=\tuple{\infty,i}$, then, for all~$j$ \st~$0\leq j<i$, there is a $u_j\in\U$ \st~$\EI(u_j)=\tuple{\infty,j}$.
\end{definition}

Observe that the version of convexity satisfied by epistemic interpretations is a straightforward extension of the convexity of ranked interpretations (Definition~\ref{Def:RankedInterpretation}). Figure~\ref{Fig:PenguinDodo2} depicts an epistemic interpretation in our running example.

\begin{figure}[h]
\begin{center}
\begin{TAB}(r,1cm,0.2cm)[3pt]{|c|c|}{|c|c|c|c|c|c|}%
{ $\tuple{\infty,\infty}$} & { $\states{\peng\land\lnot\bird}\cup\states{\dodo\land \lnot\bird}$}  \\
{ $\tuple{\infty,1}$} & { $\bar{\peng}\dodo\bird\flies$\quad $\peng\dodo\bird\flies$}\\
{ $\tuple{\infty,0}$} & {$\bar{\peng}\dodo\bird\bar{\flies}$\quad $\peng\dodo\bird\bar{\flies}$}\\
{ $\tuple{\f,2}$} & {$\peng\bar{\dodo}\bird\flies$ }\\
{ $\tuple{\f,1}$} & {$\bar{\peng}\hspace{0.02cm}\bar{\dodo}\bird\bar{\flies}$\quad $\peng\bar{\dodo}\bird\bar{\flies}$}\\
{ $\tuple{\f,0}$} & {$\bar{\peng}\hspace{0.02cm}\bar{\dodo}\bird\flies$\quad $\bar{\peng}\hspace{0.02cm}\bar{\dodo}\hspace{0.02cm}\bar{\bird}\flies$}\quad {$\bar{\peng}\hspace{0.02cm}\bar{\dodo}\hspace{0.02cm}\bar{\bird}\hspace{0.02cm}\bar{\flies}$} \\
\end{TAB}
\end{center}
\caption{Epistemic interpretation for~$\Prp=\{\bird,\dodo,\flies,\peng\}$.}
\label{Fig:PenguinDodo2}
\end{figure}

Casini~\etal.~\cite{CasiniEtal2020} have a similar definition of epistemic interpretations, but they do not allow for the rank $\tuple{\infty,\infty}$.
\myskip

We let $\U^{\f}_{\EI}\defined\{u\in\U\mid \EI(u)=\tuple{\f,i}$, for some $i\in\mathbb{N}\}$ and $\U^{\infty}_{\EI}\defined\{u\in\U\mid \EI(u)=\tuple{\infty,i}$, for some $i\in\mathbb{N}\}$. 
Note that $\U^{\infty}_{\EI}$ does \emph{not} contain valuations with rank~$\tuple{\infty,\infty}$. We let $\min\states{\alpha}_{\EI}\defined\{u\in\states{\alpha}\mid \EI(u)\preceq\EI(v)$, for all $v\in\states{\alpha}\}$,
$\min\states{\alpha}^{\f}_{\EI}\defined\{u\in\states{\alpha}\cap\U^{\f}_{\EI}\mid \EI(u)\preceq\EI(v)$, for all $v\in\states{\alpha}\cap\U^{\f}_{\EI}\}$, and
$\min\states{\alpha}^{\infty}_{\EI}\defined\{u\in\states{\alpha}\cap\U^{\infty}_{\EI}\mid \EI(u)\preceq\EI(v)$, for all $v\in\states{\alpha}\cap\U^{\infty}_{\EI}\}$.

Observe that epistemic interpretations are allowed to have no plausible valuations ($\U^{\f}_{\EI}=\emptyset$), as well as no implausible valuations that are possible ($\U^{\infty}_{\EI}=\emptyset$). This means it is possible that $\EI(u)=\tuple{\infty,\infty}$ for all $u\in\U$, in which case $\EI\sat\alpha\twiddle_{\gamma}\beta$, for all $\alpha,\beta,\gamma$ (\cf\ Definition~\ref{Def:SatisfactionEpistemicInterpretation} below). Epistemic interpretations also allow for cases where all valuations are possible (that is, either plausible or implausible but possible). This corresponds to the case where an epistemic interpretation does not have any valuation with rank ~$\tuple{\infty,\infty}$. 

Armed with the notion of epistemic interpretation, we can provide an intuitive semantics to situated conditionals.

\begin{definition}[Satisfaction and Generation of Situated Conditionals]\label{Def:SatisfactionEpistemicInterpretation}
Let~$\EI$ be an epistemic interpretation and let~$\alpha,\beta,\gamma\in\Lang$. We say~$\EI$ \df{satisfies} $\alpha\twiddle_{\gamma}\beta$, denoted as  $\EI\sat\alpha\twiddle_{\gamma}\beta$ and often abbreviated as $\alpha\twiddle^{\EI}_{\gamma}\beta$, if
\[
\left\{\begin{array}{cl}
     \min\states{\alpha\land\gamma}^{\f}_{\EI}\subseteq\states{\beta},& \text{ if } \states{\gamma}\cap\U^{\f}_{\EI}\neq\emptyset;\\
     \min\states{\alpha\land\gamma}^\infty_{\EI}\subseteq\states{\beta},& \text{otherwise.}
\end{array}\right. 
\]
We say that $\EI$ \df{generates} the situated conditional $\twiddle$ if, for every $\alpha, \beta, \gamma\in\Lang$, $\tuple{\alpha, \beta, \gamma}\in\twiddle$ iff $\alpha\twiddle^{\EI}_{\gamma}\beta$.
\end{definition}

Intuitively, the satisfaction of situated conditionals works as follows. If the situation~$\gamma$ is compatible with the plausible part of~$\EI$ (the valuations in~$\U^{\f}_{\EI}$), then $\alpha\twiddle_{\gamma}\beta$ holds if the most typical plausible models of~$\alpha\land\gamma$ are also models of~$\beta$. On the other hand, if the situation~$\gamma$ is not compatible with the plausible part of~$\EI$, \ie, all models of~$\gamma$ have an infinite rank, then $\alpha\twiddle_{\gamma}\beta$ holds if the most typical implausible (but possible) models of~$\alpha\land\gamma$ are also models of~$\beta$. 

An immediate corollary of Definition~\ref{Def:SatisfactionEpistemicInterpretation} is that the rational conditionals defined in terms of ranked interpretations can be simulated with SCs by setting the situation to~$\top$.

\begin{definition}[Extracted Ranked Interpretation]\label{Def:ExtractedRanked}
For an epistemic interpretation~$\EI$, we define the \df{ranked interpretation} $\RM^{\EI}$ \df{extracted from}~$\EI$ as follows: for $u\in\U^{\f}_{\EI}$, $\RM^{\EI}(u)=i$, where $\EI(u)=\tuple{\f,i}$, and $\RM^{\EI}(u)=\infty$ for $u\in\U\setminus\U^{\f}_{\EI}$.
\end{definition}

\begin{corollary}\label{Corollary:ClassicalConditionals}
Let $\EI$ be an epistemic interpretation. Then $\RM^{\EI}\sat\alpha\twiddle\beta$ iff $\EI\sat\alpha\twiddle_{\top}\beta$. 
\end{corollary}
\begin{proof}
Assume $\EI\sat\alpha\twiddle_{\top}\beta$. Then, by definition, we have $\min\states{\alpha\land\top}^\f_{\EI}\subseteq\states{\beta}$ if $\U^{\f}_{\EI}\neq\emptyset$, and $\min\states{\alpha\land\top}^\infty_{\EI}\subseteq\states{\beta}$ otherwise. If the former is the case, then, by the construction of~$\RM^{\EI}$, we have $\min\states{\alpha}^{\f}_{\RM^\EI}\subseteq\states{\beta}$, and therefore $\RM^{\EI}\sat\alpha\twiddle\beta$. If, instead, the latter holds, then $\states{\alpha}^{\f}_{\EI}=\emptyset$, from which it follows that $\states{\alpha}^{\f}_{\RM^\EI}=\emptyset$, and therefore $\RM^{\EI}\sat\alpha\twiddle\beta$. For the other direction, assume $\RM^{\EI}\sat\alpha\twiddle\beta$. If $\states{\alpha}^{\f}_{\RM^\EI}=\emptyset$, then, from the construction of~$\RM^{\EI}$, we have $\states{\alpha}^{\f}_{\EI}=\emptyset$, from which we get $\EI\sat\alpha\twiddle_{\top}\beta$. If $\states{\alpha}^{\f}_{\RM^\EI}\neq\emptyset$, then, since $\min\states{\alpha}^{\f}_{\RM^\EI}\subseteq\states{\beta}$, we must have $\min\states{\alpha}^{\f}_{\EI}\subseteq\states{\beta}$, too. From the latter it follows that $\min\states{\alpha\land\top}^{\f}_{\EI}\subseteq\states{\beta}$, and therefore $\EI\sat\alpha\twiddle_{\top}\beta$.
\end{proof}

The principal advantage of situated conditionals and their associated enriched semantics in terms of epistemic interpretations is that they allow us to represent different degrees of epistemic involvement, with the finite ranks (the plausible valuations) representing the expectations of an agent. So $\top\twiddle_{\top}\alpha$ being satisfied in $\EI$ indicates that $\alpha$ is expected. What an agent believes to be true corresponds to what is true in all the valuations with finite ranks. That is, the agent believes $\alpha$ to be true if and only if~$\EI\sat\neg\alpha\twiddle_{\top}\bot$, and we will abbreviate $\neg\alpha\twiddle_{\top}\bot$ with $\alpha$, extending to epistemic interpretations the convention introduced above for ranked interpretations (see Section \ref{RationalClosure}).

Another advantage of our framework is that it also allows for reasoning counterfactual: we can express that dodos would not fly if they existed in a coherent way. We can talk about dodos in a counterfactual situation or context, for example, assuming that Mauritius had never been colonised ($\neg \mc$): the conditional $\dodo\twiddle_{\lnot\mc}\lnot\flies$ is read as `In the situation of Mauritius not having been colonised, the dodo would not fly'. Importantly, we can reason coherently with a situated conditional, even when not knowing whether its premises are plausible or counterfactual. To do so, it is sufficient to introduce statements of the form $\alpha\twiddle_{\alpha}\beta$. If $\alpha$ is plausible, this conditional is evaluated in the context of the finite ranks, exactly as if $\alpha\twiddle_{\top}\beta$ were being evaluated. On the other hand, if  $\alpha\twiddle_{\top}\bot$ holds, $\alpha\twiddle_{\alpha}\beta$ will be evaluated referring to the infinite ranks. So, in the case of penguins and dodos, $\peng\twiddle_{\peng}\lnot\flies$ and $\dodo\twiddle_{\dodo}\lnot\flies$ express the information that penguins usually do not fly in the situation of penguins existing, and that dodos usually do not fly in the situation of dodos existing, regardless of whether the agent is aware of penguins or dodos existing or not. In contrast, a statement such as $\dodo\twiddle_{\top}\lnot\flies$ cannot be used to reason counterfactually about dodos, once we are aware that they do not exist (that is, $\dodo\twiddle_{\top}\bot$): given the latter, once we  consider all the valuations satisfying $\top$ (that is, all the valuations), we have to evaluate every  defeasible conditional $\dodo\twiddle\alpha$ (for any $\alpha$) looking at the valuations with finite ranks. In all such valuations, the sentence~$\dodo$ is not satisfied; hence any SC $\dodo\twiddle_{\top}\alpha$, for any $\alpha$, would be satisfied. Also, note that it is still possible to impose that something necessarily holds, both in plausible and counterfactual situations. The conditional $\alpha\twiddle_{\alpha}\bot$ holds only in epistemic interpretations in which all valuations satisfying $\alpha$ have $\tuple{\infty,\infty}$ as their rank. The following example illustrates these claims more concretely.

\begin{example}\label{Ex:PenguinDodo2}
Consider the following rephrasing of the statements in Example~\ref{Ex:PenguinDodo1}. `Birds usually fly' becomes $\bird\twiddle_{\top}\flies$. Defeasible information about penguins and dodos are modelled using $\peng\twiddle_{\peng}\lnot\flies$ and $\dodo\twiddle_{\dodo}\lnot\flies$. Given that dodos don't exist anymore, the statement $\dodo\twiddle_{\top}\bot$ leaves open the existence of dodos in the infinite ranks, which allow for coherent reasoning under the assumption that dodos exist (the situation~$\dodo$). Moreover, information such as dodos and penguins necessarily being birds can be modelled by the conditionals $\peng\land\lnot\bird\twiddle_{\peng\land\lnot\bird}\bot$ and $\dodo\land\lnot\bird\twiddle_{\dodo\land\lnot\bird}\bot$, relegating the valuations in $\states{\peng\land\lnot\bird}\cup\states{\dodo\land\lnot\bird}$ to the rank $\tuple{\infty,\infty}$. Figure~\ref{Fig:PenguinDodo2} (below Definition~\ref{Def:EpistemicInterpretation}) shows a model of these statements. (We shall address how certain models of given conditionals are excluded from the picture in Section~\ref{Entailment}, where we define a suitable form of entailment from a set of situated conditionals.)
\end{example}

Next, we consider the class of situated conditionals from the perspective of a list of \emph{situated} rationality postulates in the KLM style. We start with the following ones:
\[
\begin{array}{clccl}
(\text{Ref}) & \alpha \twiddle_{\gamma} \alpha & \quad & (\text{\small LLE}) & {\displaystyle \frac{\entails \alpha\liff \beta,\ \alpha \twiddle_{\gamma} \delta}{\beta \twiddle_{\gamma} \delta}} \\[0.5cm]
(\text{And}) & {\displaystyle \frac{\alpha \twiddle_{\gamma} \beta, \ \alpha \twiddle_{\gamma} \delta}{\alpha \twiddle_{\gamma} \beta\land\delta}} & \quad & (\text{Or}) & {\displaystyle \frac{\alpha \twiddle_{\gamma} \delta, \ \beta \twiddle_{\gamma} \delta}{\alpha \lor \beta \twiddle_{\gamma} \delta}} \\[0.5cm]
(\text{\small RW}) & {\displaystyle \frac{\alpha \twiddle_{\gamma} \beta,
\ \entails \beta \limp \delta}{\alpha \twiddle_{\gamma} \delta}} & \quad & 
(\text{\small RM}) & 
{\displaystyle \frac{\alpha \twiddle_{\gamma} \beta, \ \alpha \ntwiddle_{\gamma} \lnot\delta}
{\alpha\land\delta \twiddle_{\gamma} \beta}} 
\end{array}
\]

Observe that they correspond exactly to the original KLM postulates, except that the notion of situation has been added. As for $\alpha$ and $\beta$, the $\gamma$ occurring in the postulates should be viewed as a meta-variable ranging over~$\Lang$.

\begin{definition}[Basic Situated Conditional]
An SC~$\twiddle$ is a \df{basic situated conditional} (BSC, for short) if it satisfies the situated rationality postulates.
\end{definition}

An immediate corollary of this definition is that for a BSC with the situation~$\gamma$ fixed, $\twiddle_{\gamma}$ is a rational conditional. We then get the following result.

\begin{restatable}{theorem}{restatableTheoremBSC}\label{Theorem:BSC}
Every epistemic interpretation generates a~BSC (see Definition~\ref{Def:SatisfactionEpistemicInterpretation}). Nevertheless, the converse does not hold, \ie, some BSCs cannot be generated by any epistemic state. 
\end{restatable}

The reason why the converse of Theorem~\ref{Theorem:BSC} does not hold is that the structure of a~BSC is completely independent of the situation~$\gamma$ referred to in the situated~KLM postulates. As a very simple instance of this problem, observe that~BSCs are not even syntax-independent \wrt\ the situation. That is, we may have $\alpha\twiddle_{\gamma}\beta$ but $\alpha\ntwiddle_{\delta}\beta$, where $\gamma\equiv\delta$. To put it another way, a BSC is simply a rational defeasible consequence relation with the situation playing no role in determining the BSC's structure. To remedy this, we require~BSCs to satisfy the following additional postulates:
\[
\begin{array}{clccl}
(\text{Inc}) & \dfrac{\alpha\twiddle_{\gamma}\beta}{\alpha\land\gamma\twiddle_{\top} \beta} & & (\text{Vac}) & \dfrac{\top\ntwiddle_{\top}\lnot\gamma,\ \alpha\land\gamma\twiddle_{\top}\beta}{\alpha\twiddle_{\gamma}\beta}\\[0.5cm]
(\text{Ext}) & \dfrac{\gamma\equiv\delta}{\alpha\twiddle_{\gamma}\beta\textrm{ iff } \alpha\twiddle_{\delta}\beta} & & (\text{SupExp}) & \dfrac{\alpha\twiddle_{\gamma\land\delta}\beta}{\alpha\land\gamma\twiddle_{\delta} \beta} \\[0.5cm]
\end{array}
\]\[
\begin{array}{cl}
(\text{SubExp}) & \dfrac{\delta\twiddle_{\top}\bot,\ \alpha\land\gamma \twiddle_{\delta}\beta}{\alpha \twiddle_{\gamma\land\delta}\beta} 
\end{array}
\]

We shall refer to these as the \emph{situated AGM postulates} for reasons to be outlined below.

\begin{definition}[Full Situated Conditional]
A BSC is a \df{full SC} (FSC) if it satisfies the situated AGM postulates. 
\end{definition}

One way to interpret the addition of a situation to conditionals, from a technical perspective, is to think of it as similar to \emph{belief revision}. That is, $\alpha\twiddle_{\gamma}\beta$ can be thought of as stating that if a revision with $\gamma$ has taken place, then~$\beta$ will hold on condition that~$\alpha$ holds. With this view of situated conditionals, the situated AGM postulates above are seen as versions of the AGM postulates for belief revision~\cite{AlchourronEtAl1985}. The names of these postulates were chosen with the names of their AGM analogues in mind. The situated AGM postulates can be motivated intuitively as follows.

Together, Inc and Vac require that when the situation (or revision with)~$\gamma$ is compatible with what is currently plausible, then a conditional \wrt\ the situation~$\gamma$ (a `revison by'~$\gamma$) is the same as a conditional where the situation is~$\top$ (where there is no `revision' at all), but with~$\gamma$ added to the premise of the conditional. Ext ensures that BSCs are syntax-independent of the situation. Finally, SupExp and SubExp together require that if the situation~$\delta$ is implausible (that is, the `revision' with~$\delta$ is incompatible with what is plausible), then a conditional \wrt\ the situation~$\gamma\land\delta$ (a `revision by'~$\gamma\land\delta$) is the same as a conditional where the situation (or `revision') is~$\delta$, but with~$\gamma$ added to the premise of the conditional.

It turns out that FSCs are characterised by epistemic interpretations, resulting in the following representation result.

\begin{restatable}{theorem}{restatableTheoremFSC}\label{Theorem:FSC}
Every epistemic interpretation generates an FSC. Every FSC can be generated by an epistemic interpretation.
\end{restatable}

The AGM-savvy reader may have noticed that the following two obvious analogues of the suite of situated AGM postulates are missing from our list above.
\[
\begin{array}{clccl}
(\text{Succ}) & \alpha\twiddle_{\gamma}\gamma & & (\text{Cons}) & \top\twiddle_{\gamma}\bot \text{ iff } \gamma\equiv\bot
\end{array}
\]

Succ requires situations to matter: a `revision' by~$\gamma$ will always be successful. Cons states that we obtain an inconsistency only when the situation is inconsistent. 

It turns out that Succ holds for epistemic interpretations: it follows from the combination of the situated~KLM and~AGM postulates. On the other hand, just one direction of Cons holds.

\begin{corollary}
Every FSC satisfies Succ, but there are FSCs for which Cons does not hold. However, the right-to-left direction of Cons holds: If $\gamma\equiv\bot$, then $\top\twiddle_{\gamma}\bot$.
\end{corollary}
\begin{proof}
To prove that~Succ holds, it suffices, by Theorem~\ref{Theorem:BSC}, to show that $\EI\sat\alpha\twiddle_{\gamma}\gamma$ for all epistemic interpretations~$\EI$ and all~$\alpha,\gamma$. To see that this holds, observe that~$\states{\alpha\land\gamma}_{\EI}\subseteq\states{\gamma}_{\EI}$. 

To prove that~Cons does not hold, it  suffices, by Theorem~\ref{Theorem:BSC}, to show that there is an epistemic interpretation~$\EI$ such that~$\EI\sat\top\twiddle_{\gamma}\bot$ but~$\gamma\not\equiv\bot$. To construct such an~$\EI$, let $\U^{\f}_{\EI} = \U^{\infty}_{\EI}=\emptyset$ (and so $\EI(u)=\tuple{\infty,\infty}$ for all $u\in\U)$. It is easy to see that by picking any~$\gamma$ \st~$\gamma\not\equiv\bot$ the result follows.

To prove that if $\gamma\equiv\bot$ then $\top\twiddle_{\gamma}\bot$, note that, by Definition \ref{Def:SatisfactionEpistemicInterpretation} and Theorem \ref{Theorem:FSC}, $\top\twiddle_{\gamma}\bot$ iff $\min\states{\top\land\gamma}^\infty_{\EI}\subseteq\states{\bot}$ whenever $\gamma\equiv\bot$, which holds since $\min\states{\top\land\gamma}^\infty_{\EI} = \states{\bot} = \emptyset$.  
\end{proof}

 
We conclude this section by considering the following two properties.
\[
\begin{array}{clccl}
(\text{Incons}) & \alpha\twiddle_{\bot}\beta & & (\text{Cond}) & \text{If } \gamma\ntwiddle_{\top}\bot, \text{ then } \alpha\land\gamma\twiddle_{\top}\beta \text{ iff }\alpha\twiddle_{\gamma}\beta
\end{array}
\]

Incons requires that all conditionals hold when the situation is inconsistent, while Cond requires that conditionals \wrt\ the situation~$\gamma$ be equivalent to the same conditional with~$\gamma$ added to the premise and with a tautologous situation (\ie, the situation is~$\top$), provided that $\gamma$ is not inconsistent \wrt\ the tautologous situation.

\begin{proposition}\label{Prop:InconsCond}
Every FSC satisfies Incons and Cond.
\end{proposition}
\begin{proof}
To prove that Incons holds, it suffices, by Theorem~\ref{Theorem:BSC}, to show that~$\EI\sat\alpha\twiddle_{\bot}\beta$ for all epistemic interpretations~$\EI$, and all~$\alpha,\beta$. To see that this holds, observe that~$\states{\alpha\land\bot}_{\EI}=\emptyset$. 

To prove that Cond holds, it suffices, by Theorem~\ref{Theorem:BSC}, to show that if~$\EI\nsat\gamma\twiddle_{\top}\bot$, then~$\EI\sat\alpha\land\gamma\twiddle_{\top}\beta$ iff~$\EI\sat\alpha\twiddle_{\gamma}\beta$ for all epistemic interpretations~$\EI$, and all~$\alpha,\beta,\gamma$. So, suppose that~$\EI\nsat\gamma\twiddle_{\top}\bot$. By Definition~\ref{Def:SatisfactionEpistemicInterpretation}, this means that~$\U^{\f}_{\EI}\cap\states{\gamma}\neq\emptyset$ and also that~$\U^{\f}_{\EI}\cap\states{\top}\neq\emptyset$. From this, by Definition~\ref{Def:SatisfactionEpistemicInterpretation}, we need to show that~$\states{\alpha\land\gamma\land\top}^{\f}_{\EI}\subseteq\states{\beta}$ iff $\states{\alpha\land\gamma}^{\f}_{\EI}\subseteq\states{\beta}$ for the result to hold, which follows immediately.
\end{proof}

\section{Reasoning with Situated Conditionals}\label{Entailment}

The previous section provides a framework for characterising the class of full situated conditionals in terms of epistemic interpretations. In this section, we move to an investigation of how we can reason within this framework. More precisely, the question of interest is the following: given a finite set of situated conditionals, or a \emph{situated conditional knowledge base} (SCKB) $\KB$, which situated conditionals can be said to be \emph{entailed} from it? That is, for example, given an SCKB consisting of the conditionals `birds typically fly' ($\bird\twiddle_{\top}\flies$), `penguins are birds' ($\peng\land\lnot\bird\twiddle_{\peng\land\lnot\bird}\bot$), `emperor penguins are penguins' ($\epeng\land\lnot\peng\twiddle_{\epeng\land\lnot\peng}\bot$), and `penguins typically do not fly' ($\peng\twiddle_{\peng}\neg \flies$), should or should we not derive that emperor penguins typically do not fly ($\epeng\twiddle_{\epeng}\neg \flies$)? 
In a non-monotonic framework, it is generally not appropriate to consider entailment relations that are Tarskian in nature, \ie, that determine the consequences of a knowledge base by looking at what holds in \emph{all} the models of the knowledge base. This is because such entailment relations are, by definition, monotonic: let $\KB$ and $\KB'$ be two SCKB's \st~$\KB\subseteq \KB'$, and let $\alpha\twiddle_{\gamma}\beta$ be a consequence of $\KB$. That is, it is satisfied by every model of $\KB$. Since every model of $\KB'$ is also a model of $\KB$, $\alpha\twiddle_{\gamma}\beta$ is satisfied by every model of $\KB'$ too. That is, $\alpha\twiddle_{\gamma}\beta$ is also a consequence of $\KB'$.
Because of monotonicity, if we reason with a Tarskian approach in a conditional setting, we are relegated to very weak inferences. For example, if we only know that `birds typically fly' ($\bird\twiddle_{\top}\flies$) and `robins are birds' ($\rob\land\lnot\bird\twiddle_{\rob\land\lnot\bird}\bot$), we could not even draw a simple conclusion based on property inheritance such as `robins typically fly' ($\rob\twiddle_{\rob}\flies$), since the initial information can also be satisfied by interpretations in which robins are atypical non-flying birds. Hence in a monotonic framework, we cannot, for example, reason under the principle of \emph{presumption of typicality} \cite{Lehmann1995}, assuming that everything behaves according to our expectations unless we are explicitly informed that this is not the case.

Since Tarskian inference relations tend to be too weak, inferentially speaking~\cite{LehmannMagidor1992}, in the framework of non-monotonic reasoning, more suitable entailment relations can be defined by choosing a single model of the knowledge base that satisfies some desirable postulates. It is generally accepted that there is not a unique entailment relation for defeasible reasoning, with different forms of entailment being dependent on the kind of reasoning one wants to model~\cite{Lehmann1995,CasiniEtAl2019-JELIA}. In the framework of preferential semantics, rational closure, recalled in Section~\ref{Preliminaries}, is generally recognised as a core form of entailment, with other suitable forms of entailment being refinements of rational closure.

We now define a form of entailment for situated conditionals, which we call \emph{minimal closure}~(MC). It is based on a semantic construction that reformulates in the framework of situated conditionals the semantic construction characterising rational closure for defeasible conditionals: we adapt the notion of a minimal model~\cite{GiordanoEtAl2015}, recalled in Section~\ref{Preliminaries}, for our framework, and show that for any SCKB the minimal model is unique.

In the rest of the section, we proceed as follows:  we first define the notion of consistency in the present setting. Then, we connect the notions of satisfaction in epistemic states for situated and classic conditionals, respectively. Such a connection, expressed in particular by the content of Corollary~\ref{Corollary:SituationAsPremise} below, allows us to use known results, regarding entailment relations for defeasible conditionals in ranked interpretations, for the definition of an entailment relation for situated conditionals and epistemic interpretations. In particular, using known results about minimal ranked models \cite{GiordanoEtAl2015}, we can define our minimal closure on top of the well-known rational closure. Moving from that, in Section~\ref{Algorithm}, we define a decision procedure for entailment that is based on a series of decision steps for classical propositional logic.

Starting from  rational closure, which is defined for defeasible conditionals and ranked interpretations, we define minimal closure, defined for situated conditionals and epistemic interpretations, using the connections between the former framework and the latter one. First, we can extend the notion of consistency for defeasible conditionals to situated conditionals. We have seen (Section \ref{RationalClosure}) that a set~$\C$ of defeasible conditionals is \emph{consistent} if and only if it has a ranked model~$\RM$ \st\ $\states{0}_{\RM}\neq\emptyset$. Such a condition indicates that the agent has a consistent set of expectations since such a model does not satisfy the conditional~$\top\twiddle\bot$, which captures absurdity in the conditional framework.  This condition can easily be translated into our framework.

\begin{definition}[SCKB Consistency]\label{Def:SCKBconsistency}
An SCKB is \df{consistent} if it has an epistemic model $\EI$ \st\ $\states{\tuple{\f,0}}_{\EI}\neq\emptyset$.
\end{definition}

In other words, an SCKB is consistent if it has an epistemic model~$\EI$ that does not satisfy~$\top\twiddle_{\top}\bot$. $\states{\tuple{\f,0}}_{\EI}$ is a notation for epistemic interpretations that mirrors the notation $\states{0}_{\RM}$ for ranked interpretations, that is, $\states{\tuple{x,y}}_{\EI}$ represents the set of worlds that have rank $\tuple{x,y}$ in $\EI$. On the other hand, given Corollary~\ref{Corollary:ClassicalConditionals}, $\top\twiddle_{\top}\bot$ is a situated conditional that has the same meaning as the  defeasible conditional $\top\twiddle\bot$, that is, an  agent believing $\top\twiddle_{\top}\bot$  believes to be in an inconsistent situation since it expects $\bot$ to hold.

Given Corollary~\ref{Corollary:ClassicalConditionals}, we can intuitively introduce a notion of satisfaction of defeasible conditionals also for epistemic interpretations:
\[
\EI\sat \alpha\twiddle\beta\quad\text{ iff }\quad\EI\sat \alpha\twiddle_{\top}\beta
\]

Note that an epistemic interpretation~$\EI$ satisfies exactly the same defeasible conditionals of its extracted ranked interpretation~$\RM^{\EI}$ (see Definition~\ref{Def:ExtractedRanked}). That is, the ranks specified in the interval $\U^{\infty}_{\EI}\cup\states{\tuple{\infty,\infty}}$ are totally irrelevant \wrt\ the satisfaction of the defeasible conditionals of the form~$\alpha\twiddle\beta$. We can also intuitively define the converse operation \wrt~the extraction of a ranked interpretation from an epistemic one (Definition~\ref{Def:ExtractedRanked}): we can \emph{extract} an epistemic interpretation from a given ranked interpretation. Such an extraction is simply a direct translation of the ranks of the ranked interpretations into the formalism of the epistemic interpretations, simply associating the value $\tuple{\infty,\infty}$ to all the worlds that have the rank $\infty$ in the ranked interpretation.

\begin{definition}[Extracted Epistemic Interpretation]\label{Def:ExtractedEpistemic}
For a ranked interpretation~$\RM$, we define the \df{epistemic interpretation} $\EI^{\RM}$ \df{extracted from}~$\RM$ as follows: for $u\in\U^{\f}_{\RM}$, $\EI^{\RM}(u)=\tuple{\f,i}$, where $\RM(u)=i$, and $\EI^{\RM}(u)=\tuple{\infty,\infty}$, for $u\in\U\setminus\U^{\f}_{\RM}$.
\end{definition}

It is easy to see that~$\RM$ and~$\EI^{\RM}$ are equivalent \wrt\ the satisfaction of defeasible conditionals.

The following corollary of Proposition~\ref{Prop:InconsCond}, which is simply a semantic reformulation of the postulate~Cond, will be central in connecting the satisfaction of situated conditionals to that of defeasible ones.

\begin{corollary}\label{Corollary:SituationAsPremise}
For every epistemic interpretation~$\EI$, if $\U^{\f}_{\EI}\cap\states{\gamma}\neq\emptyset$, then $\EI\sat\alpha\twiddle_{\gamma}\beta$ iff $\EI\sat\alpha\land\gamma\twiddle\beta$.
\end{corollary}
\begin{proof}
Since it is just a semantic reformulation of the postulate~Cond, it follows directly from the proof that~Cond holds (Proposition~\ref{Prop:InconsCond}).
\end{proof}

Given Corollary~\ref{Corollary:SituationAsPremise}, we define a simple transformation of situated conditional knowledge bases.

\begin{definition}
Let~$\KB$ be an~SCKB; with~$\KB^{\land}$ we denote its \emph{conjunctive classical form}, defined as follows: $\KB^{\land}\defined\{\alpha\land\gamma\twiddle \beta\mid \alpha\twiddle_{\gamma} \beta \in\KB\}$.
\end{definition}

We can use the conjunctive classical form to define two models for an SCKB: the \emph{classical epistemic model} and the \emph{minimal epistemic model}.  The former will allow us to prove that checking  the logical consistency of an~SCKB can be reduced to a consistency check in propositional logic (see Corollary \ref{Corollary:Consistency} below). The latter is the epistemic model, which characterises the~MC of an SCKB. Given an SCKB~$\KB$, both its classical epistemic model and its minimal epistemic model are defined starting from the minimal ranked model of its conjunctive classical form $\KB^{\land}$ (see Definitions \ref{Def:ClassicalEpistemicModel} and \ref{Def:ConstructionMinimalModel} below).

\begin{definition}[Classical Epistemic Model]\label{Def:ClassicalEpistemicModel}
Let~$\KB$ be an~SCKB, $\KB^{\land}$ its conjunctive classical form, and~$\RM$ the minimal ranked model of~$\KB^{\land}$. The \df{classical epistemic model} of~$\KB$ is the epistemic interpretation~$\EI^{\RM}$ extracted from~$\RM$ (see Definition~\ref{Def:ExtractedEpistemic}).
\end{definition}

Since~$\RM$ is a ranked model of~$\KB^{\land}$, so is~$\EI^{\RM}$. We need to check whether~$\EI^{\RM}$ is also a model of~$\KB$.

\begin{proposition}\label{Prop:ClassicalModel}
Let~$\KB$ be an~SCKB, and let $\EI^{\RM}$ be defined as in Definition~\ref{Def:ClassicalEpistemicModel}. Then, we have that~$\EI^{\RM}$ is a model of~$\KB$.
\end{proposition}
\begin{proof}
Let~$\alpha\twiddle_{\gamma}\beta\in\KB$. Since~$\EI^{\RM}$ is an epistemic model of $\KB^{\land}$ and we have Corollary~\ref{Corollary:SituationAsPremise}, if~$\states{\gamma}\cap \U^{\f}_{\EI^{\RM}}\neq\emptyset$, then we conclude~$\EI^{\RM}\sat\alpha\twiddle_{\gamma}\beta$. Otherwise, suppose~$\states{\gamma}\cap \U^{\f}_{\EI^{\RM}}=\emptyset$. Since~$\EI^{\RM}$ is an extracted epistemic interpretation (Definition \ref{Def:ExtractedEpistemic}), its only infinite rank is $\tuple{\infty,\infty}$, and we have $\states{\gamma}\subseteq \states{\tuple{\infty,\infty}}$, which implies $\states{\alpha\land\gamma}\subseteq \states{\tuple{\infty,\infty}}$, which in turn implies~$\EI^{\RM}\sat\alpha\twiddle_{\gamma}\beta$.
\end{proof}

From Proposition~\ref{Prop:ClassicalModel} and Corollary~\ref{Corollary:SituationAsPremise}, we can prove the following result.

\begin{proposition}\label{Prop:Consistency}
Let~$\KB$ be an~SCKB. $\KB$ has an epistemic model with $\states{\tuple{\f,0}}\neq\emptyset$ iff~$\KB^{\land}$ has a ranked model with $\states{0}\neq\emptyset$.
\end{proposition}
\begin{proof}
Proposition~\ref{Prop:ClassicalModel} and Definitions \ref{Def:ExtractedEpistemic} and \ref{Def:ClassicalEpistemicModel} show that if~$\KB^{\land}$ has a ranked model with $\states{0}\neq\emptyset$, then~$\KB$ has an epistemic model with $\states{\tuple{\f,0}}\neq\emptyset$. For the opposite direction, assume that~$\KB$ has an epistemic model~$\EI$ \st~$\states{\tuple{\f,0}}_{\EI}\neq\emptyset$. From~$\EI$, we define an epistemic model~$\EI_{rk}$ in the following way:
\[
\EI_{rk}(u)=
\left\{\begin{array}{ll}
    \EI(u), & \text{if } \EI(u)=\tuple{\f,i} \text{ for some }i;\\
    \tuple{\infty, \infty}, & \text{otherwise.}
\end{array}\right.
\]

Clearly $\states{\tuple{\f,0}}_{\EI_{rk}}\neq\emptyset$. It is easy to check that~$\EI_{rk}$ is an epistemic model of~$\KB$. Moreover, thanks to Corollary~\ref{Corollary:SituationAsPremise}, we can prove that it is also an epistemic model of~$\KB^{\land}$: for every~$\alpha\twiddle_{\gamma}\beta\in\KB$, 
if $\states{\gamma}\cap \U^{\f}_{\EI_{rk}}\neq\emptyset$, then~$\EI_{rk}\sat\alpha\land\gamma\twiddle\beta$ by Corollary~\ref{Corollary:SituationAsPremise}; 
if $\states{\gamma}\cap \U^{\f}_{\EI_{rk}}=\emptyset$, then~$\states{\alpha\land\gamma}\subseteq\states{\tuple{\infty,\infty}}$, and we can conclude~$\EI_{rk}\sat\alpha\land\gamma\twiddle\beta$.

Let~$\RM$ be the ranked model corresponding to~$\EI_{rk}$, that is,
\[
\RM(u)=
\left\{\begin{array}{ll}
    i, & \text{if } \EI_{rk}(u)=\tuple{\f,i} \text{ for some }i;\\
    \infty, & \text{otherwise.}
\end{array}\right.
\]

We have that $\states{\tuple{\f,0}}_{\EI_{rk}}\neq\emptyset$ implies $\states{0}\neq\emptyset$. Since for every pair of valuations $u,v$ in $\U$, $u$ is preferred to $v$ in~$\EI_{rk}$ iff  $u$ is preferred to $v$ in~$\RM$, it is easy to see that if~$\EI_{rk}$ is an epistemic model of~$\KB^{\land}$, then $\RM$ is a ranked model of~$\KB^{\land}$.
\end{proof}

Proposition \ref{Prop:Consistency} tells us that the consistency of an SCKB~$\KB$ corresponds to the consistency of the conditional knowledge base~$\KB^{\land}$, the conjunctive form of~$\KB$. By linking the satisfaction of an SCKB~$\KB$ to the satisfaction of its conjunctive form~$\KB^\land$, we can define a simple method for checking the consistency of an~SCKB, based on the \emph{materialisation} $\overline{\KB^\land}$ of~$\KB^\land$. The materialisation~$\overline{\C}$ of a set of defeasible conditionals~$\C$ is the set of material implications corresponding to the conditionals in~$\C$, defined in the following way:
\[
\overline{\C}\defined\{\alpha\rightarrow\beta\mid \alpha\twiddle\beta\in\C\}
\]

\begin{corollary}\label{Corollary:Consistency}
An SCKB~$\KB$ is consistent iff $\overline{\KB^\land}\not\entails\bot$.
\end{corollary}

This corollary is an immediate consequence of Proposition~\ref{Prop:Consistency} and the well-known property that a finite set of defeasible conditionals is consistent if and only if its materialisation is a consistent propositional knowledge base~\cite[Lemma 5.21]{LehmannMagidor1992}.

\begin{example}\label{Ex:Consistency}
Consider an SCKB $\KB=\{\alpha\twiddle_{\alpha}\bot, \top\twiddle_{\top}\alpha\land \beta\}$. The meaning of $\alpha\twiddle_{\alpha}\bot$ is that~$\alpha$ is necessarily false, while $\top\twiddle_{\top}\alpha\land \beta$ indicates that the agent presumes that $\alpha\land \beta$ holds (see Example \ref{Ex:PenguinDodo2}). Clearly, this is a simple inconsistent knowledge base since it is not rational to consider $\alpha$ as presumably true and necessarily false at the same time. In fact, its only epistemic model is the epistemic model in which all the worlds have rank $\tuple{\infty, \infty}$.

We can actually check the inconsistency of $\KB$ easily: according to Proposition \ref{Prop:Consistency}, $\KB$ has an  epistemic model with $\states{\tuple{\f,0}}\neq\emptyset$ iff $\KB^{\land}$ has a ranked model with $\states{0}\neq\emptyset$, and, in turn, $\KB^{\land}$ has a ranked model with $\states{0}\neq\emptyset$ iff $\overline{\KB^\land}\not\entails\bot$ (Corollary \ref{Corollary:Consistency}). $\KB^{\land}=\{\alpha\land\alpha \twiddle\bot, \top\land \top\twiddle\alpha\land \beta\}$, and $\overline{\KB^{\land}}=\{(\alpha\land\alpha) \rightarrow\bot, (\top\land \top)\rightarrow\alpha\land \beta\}$, which is logically equivalent to the set $\{\neg \alpha, \alpha\land \beta\}$, which is clearly inconsistent (that is, $\overline{\KB^{\land}}\entails \bot$).
\end{example}

Hence, the classical epistemic model allows reducing SCKB consistency checking to a simple propositional satisfiability checking. This is because it is a direct translation of a ranked interpretation into an equivalent epistemic interpretation. At the same time, since classical epistemic models do not cater for an immediate definition of appropriate forms of entailment (at least in a non-monotonic setting), we now move to the definition of the \emph{minimal epistemic model}, referring to the minimality order introduced for ranked interpretations in Section~\ref{Preliminaries}. We need to adapt, in an intuitive way, the notion of minimality defined for ranked interpretations to the present framework. In Section~\ref{CC}, we  defined a total ordering~$\preceq$ over the tuples~$\tuple{x,y}$ representing the ranks in epistemic interpretations. Let the ordering~$\preceq_{\KB}$ on all the epistemic models of an SCKB~$\KB$ be defined as follows: $\EI_{1}\preceq_{\KB}\EI_{2}$, if, for every $v\in\U$, $\EI_{1}(v)\preceq\EI_{2}(v)$. We use $\EI_{1}\prec_{\KB}\EI_{2}$ to denote its strict counterpart ($\EI_{1}\preceq_{\KB}\EI_{2}$ and $\EI_{2}\not\preceq_{\KB}\EI_{1}$).

\begin{definition}[Minimal Epistemic Model]\label{Def:MinimalEpistemicModel}
Let~$\KB$ be a consistent SCKB, and $\E_{\KB}$ be the set of its epistemic models. $\EI\in \E_{\KB}$ is a \df{minimal epistemic model of}~$\KB$ if there is no $\EI'\in\E_{\KB}$ \st~$\EI'\prec_{\KB}\EI$.
\end{definition}

We first define the construction of a model, given a consistent SCKB~$\KB$. Then we prove that it is actually the unique minimal epistemic model of~$\KB$ \wrt\ the ordering~$\prec_{\KB}$.

\begin{definition}[Construction of a Minimal Epistemic Model]\label{Def:ConstructionMinimalModel}
Let~$\KB$ be a consistent SCKB, $\KB^{\land}$ its conjunctive classical form, and let~$\RM$ be the minimal ranked model of~$\KB^{\land}$. We pick out in a set $\KB_{\infty}$ the conditionals in~$\KB$ associated with a situation that has infinite rank in~$\RM$, that is, 
\begin{itemize}
    \item $\KB_{\infty}\defined\{\alpha\twiddle_{\gamma}\beta\in \KB\mid \RM(\gamma)=\infty\}$.
\end{itemize}

And from $\KB_{\infty}$ we define the set $\KB_{\infty\downarrow}^{\land}$:    
    
\begin{itemize}
    \item $\KB_{\infty\downarrow}^{\land}\defined\{\alpha\land \gamma\twiddle\beta\mid \alpha\twiddle_{\gamma}\beta\in \KB_{\infty}\}\cup\{\fm{\U_{\RM}^{\f}}\twiddle\bot\}$.
\end{itemize}
We construct the interpretation~$\EI_{\KB}$ in the following way:
\begin{enumerate}
\item  For every $u\in\U_{\RM}^{\f}$, if $\RM(u)=i$, then $\EI_{\KB}(u)=\tuple{\f,i}$;
\item Let $\RM'$ be the minimal ranked model of~$\KB_{\infty\downarrow}^{\land}$. For every $u\in\U_{\RM}^{\infty}$, if~$\RM'(u)=i$, with $i\in \mathbb{N}\cup\{\infty\}$, then~$\EI_{\KB}(u)=\tuple{\infty,i}$.
\end{enumerate}
\end{definition}

Definition~\ref{Def:ConstructionMinimalModel} proceeds as follows. First, we want to partition the conditionals that can be considered plausible (that is, the associated situation can be satisfied by valuations with a finite rank) from those that must be considered implausible (that is, the associated situation can be satisfied only by valuations with infinite ranks). This is the set $\KB_{\infty}$. According to Definition \ref{Def:SatisfactionEpistemicInterpretation}, given an epistemic interpretation, a conditional $\alpha\twiddle_{\gamma}\beta$ is evaluated \wrt~plausible valuations if and only if $\gamma$ is satisfied by some plausible valuation. It is evaluated \wrt~implausible valuations otherwise. Given an SCKB $\KB$, $\gamma$ is not satisfied by any plausible valuation in any model of $\KB$ if and only if $\gamma\twiddle_{\top}\bot$ is satisfied by every model of~$\KB$, which, by Corollaries~\ref{Corollary:SituationAsPremise} and~\ref{Corollary:ClassicalConditionals}, justifies the use of the minimal ranked model~$\RM$ of the conjunctive form~$\KB^{\land}$ for  identifying $\KB_{\infty}$. We then specify the minimal configuration satisfying~$\KB$, considering first the finite ranks, and then the infinite ones. Corollary~\ref{Corollary:SituationAsPremise} tells us that, \wrt\ the plausible situations (\ie, finite ranks), the minimal configuration is associated with the conjunctive classical form. Hence, we refer again to the minimal ranked model~$\RM$ of~$\KB^{\land}$ to decide the configuration of the plausible valuations (Point~1 in Definition~\ref{Def:ConstructionMinimalModel}). We move to configure the infinite ranks, which need to have the minimal configuration satisfying $\KB_{\infty}$, the counterfactual conditionals in our knowledge base. In order to decide such a configuration, we consider $\KB_{\infty\downarrow}^{\land}$: all the conditionals in $\KB_{\infty}$, plus the conditional negating  the formula characterising all the plausible valuations ($\fm{\U_{\RM}^{\f}}\twiddle\bot$). The idea behind the use of $\KB_{\infty\downarrow}^{\land}$ is as follows. We want to construct a minimal ranking of the counterfactual situations. In order to do that, we need to consider matters from the perspective of being in the counterfactual situations. To do that, we introduce $\fm{\U_{\RM}^{\f}}\twiddle\bot$. In this  way,  all the situations that are plausible \wrt~$\KB$ are now considered impossible, and the plausible situations \wrt~$\KB_{\infty\downarrow}^{\land}$ are the ones that were counterfactual \wrt~$\KB$. The rank of the implausible valuations in $\EI_{\KB}$ is then determined by the rank  of the same valuations in the minimal model of $\KB_{\infty\downarrow}^{\land}$:
$\RM'$ defines the minimal configuration satisfying the conditionals in~$\KB_{\infty\downarrow}^{\land}$, and, at Point~2 in Definition~\ref{Def:ConstructionMinimalModel}, we put such a configuration `on top' of the finite ranks to define~$\EI_{\KB}$. There is the possibility that the conditional knowledge base $\KB_{\infty\downarrow}^{\land}$ is not consistent (see Section~\ref{RationalClosure}). In such a case, Definition~\ref{Def:ConstructionMinimalModel} still holds: the only model of $\KB_{\infty\downarrow}^{\land}$  is the one associating to every valuation the rank $\infty$, and consequently for every  $u\in\U_{\RM}^{\infty}$, $\EI_{\KB}(u)=\tuple{\infty,\infty}$.

We need to prove that~$\EI_{\KB}$ is an epistemic model of~$\KB$, and that, moreover, it is the unique minimal epistemic model of~$\KB$.

Let~$\EI$ be an epistemic interpretation. We build an interpretation~$\EI^{\infty}_{\downarrow}$, the \emph{counterfactual shifting}\label{countshift} of~$\EI$, in the following way. For every valuation~$u$,
\[
\EI^{\infty}_{\downarrow}(u)\defined
\left\{\begin{array}{ll}
     \tuple{\f,i}, & \text{if } \EI(u)= \tuple{\infty,i}, \text{ with } i<\infty;\\
     \tuple{\infty,\infty}, & \text{otherwise}.
\end{array}\right.
\]

Intuitively, $\EI^{\infty}_{\downarrow}$ simply shifts the infinite ranks in~$\EI$ to the finite ranks. For~$\EI^{\infty}_{\downarrow}$, we can prove a lemma corresponding to Corollary~\ref{Corollary:SituationAsPremise}.

\begin{restatable}{lemma}{restatableLemmaShifting}\label{Lemma:Shifting}
For every epistemic interpretation~$\EI$, if~$\U^{\f}_{\EI}\cap\states{\gamma}=\emptyset$, then $\EI\sat\alpha\twiddle_{\gamma}\beta$ iff $\EI^{\infty}_{\downarrow}\sat\alpha\land\gamma\twiddle\beta$.
\end{restatable}

Using Corollary~\ref{Corollary:SituationAsPremise} and Lemma~\ref{Lemma:Shifting}, it becomes easy to prove that~$\EI_{\KB}$ is indeed an epistemic model of~$\KB$.

\begin{proposition}\label{Prop:ModelOfE}
Let~$\KB$ be a consistent SCKB, and let~$\EI_{\KB}$ be the epistemic interpretation built as in Definition~\ref{Def:ConstructionMinimalModel}. Then, $\EI_{\KB}$ is an epistemic model of~$\KB$.
\end{proposition}
\begin{proof}
Let $\KB_{\infty}$ be defined as in Definition~\ref{Def:ConstructionMinimalModel}. We  distinguish two possible cases.
\begin{itemize}
\item $\alpha\twiddle_{\gamma}\beta\in\KB\setminus\KB_{\infty}$, that is, $\EI_{\KB}(\gamma)=\tuple{\f,i}$, for some~$i$. By the construction of~$\EI_{\KB}$ (Definition~\ref{Def:ConstructionMinimalModel}), $\EI_{\KB}$ is an epistemic model of~$\KB^{\land}$, that is, it is an epistemic model of~$\alpha\land\gamma\twiddle\beta$. From Corollary~\ref{Corollary:SituationAsPremise}, it follows that~$\EI_{\KB}\sat\alpha\twiddle_{\gamma}\beta$.
\item $\alpha\twiddle_{\gamma}\beta\in\KB_{\infty}$, that is, $\EI_{\KB}(\gamma)=\tuple{\infty,i}$, for some~$i$. By the construction of~$\EI_{\KB}$ (Definition~\ref{Def:ConstructionMinimalModel}), $\EI_{\KB}$ is an epistemic model of~$\KB^{\land}$, that is, it is an epistemic model of~$\alpha\land\gamma\twiddle\beta$. Let~$\EI_{\KB\downarrow}^{\infty}$ be the counterfactual shifting of~$\EI_{\KB}$. From Lemma~\ref{Lemma:Shifting}, we know that, since~$\EI_{\KB\downarrow}^{\infty}\sat\alpha\land\gamma\twiddle\beta$, $\EI_{\KB\downarrow}^{\infty}\sat\alpha\twiddle_{\gamma}\beta$ holds. Since~$\states{\alpha\land\gamma}_{\EI_{\KB}}=\states{\alpha\land\gamma}_{\EI_{\KB}}^{\infty}=\states{\alpha\land\gamma}_{\EI_{\KB\downarrow}^\infty}$, for every $u\in\U$, we have~$u\in\states{\alpha\land\gamma}_{\EI_{\KB\downarrow}^\infty}$ iff~$u\in \states{\alpha\land\gamma}_{\EI_{\KB}}$, that is, $\EI_{\KB}\sat \alpha\twiddle_{\gamma}\beta$.
\end{itemize}

Therefore, for every~$\alpha\twiddle_{\gamma}\beta\in\KB$, we have~$\EI_{\KB}\sat \alpha\twiddle_{\gamma}\beta$, and the result follows.
\end{proof}

We proceed by showing that~$\EI_{\KB}$ above is actually the only minimal epistemic model of~$\KB$.

\begin{restatable}{proposition}{restatablePropUniqueMinimalModel}
\label{Prop:UniqueMinimalModel}
Let~$\KB$ be a consistent SCKB, and let~$\EI_{\KB}$ be the epistemic interpretation built as in Definition~\ref{Def:ConstructionMinimalModel}. Then $\EI_{\KB}$ is the only minimal epistemic model of~$\KB$.
\end{restatable}


\begin{example}\label{Ex:MinimalModel}
Assume the SCKB $\KB=\{\bird\twiddle_{\top}\flies,\peng\twiddle_{\peng}\lnot\flies,\dodo\twiddle_{\dodo}\lnot\flies,\dodo\twiddle_{\top}\bot,\peng\land\lnot\bird\twiddle_{\peng\land\lnot\bird}\bot,\dodo\land\lnot\bird\twiddle_{\dodo\land\lnot\bird}\bot\}$ from Example~\ref{Ex:PenguinDodo2}. Then we have $\KB^{\land}=\{\bird\land\top\twiddle\flies,\peng\land\peng\twiddle\lnot\flies,\dodo\land\dodo\twiddle\lnot\flies,\dodo\land\top\twiddle\bot,\peng\land\lnot\bird\land\peng\land\lnot\bird\twiddle\bot,\dodo\land\lnot\bird\land\dodo\land\lnot\bird\twiddle\bot\}$, which is rank equivalent to $\{\bird\twiddle\flies,\peng\twiddle\lnot\flies,\dodo\twiddle\lnot\flies,\dodo\twiddle\bot,\peng\land\lnot\bird\twiddle\bot,\dodo\land\lnot\bird\twiddle\bot\}$. Figure~\ref{Figure:PenguinDodoRepeated} depicts the minimal ranked model of~$\KB^{\land}$. Following Definition~\ref{Def:ConstructionMinimalModel}, we have $\KB_\infty=\{\dodo\twiddle_{\dodo}\lnot\flies,\peng\land\lnot\bird\twiddle_{\peng\land\lnot\bird}\bot,\dodo\land\lnot\bird\twiddle_{\dodo\land\lnot\bird}\bot\}$. From $\KB_\infty$, we get $\KB_{\infty\downarrow}^{\land}=\{\dodo\land\dodo\twiddle\lnot\flies,\peng\land\lnot\bird\land\peng\land\lnot\bird\twiddle\bot,\dodo\land\lnot\bird\land\dodo\land\lnot\bird\twiddle\bot,(\peng\limp\bird)\land \lnot\dodo\twiddle\bot\}$, which is rank equivalent to $\{\dodo\twiddle\lnot\flies,\peng\land\lnot\bird\twiddle\bot,\dodo\land\lnot\bird\twiddle\bot,(\peng\limp\bird)\land \lnot\dodo\twiddle\bot\}$ (note that $(\peng\limp\bird)\land \lnot\dodo\twiddle\bot\in\KB_{\infty\downarrow}^{\land}$ since $(\peng\limp\bird)\land \lnot\dodo\twiddle\bot$ corresponds to the conditional $\fm{\U_{\RM}^{\f}}\twiddle\bot$, as indicated in Definition \ref{Def:ConstructionMinimalModel}). Following Steps~1 and~2 in Definition~\ref{Def:ConstructionMinimalModel}, we construct the minimal epistemic model of the original knowledge base, which is shown in Figure~\ref{Fig:PenguinDodoRepeated}.
\end{example}

\begin{figure}[h]
\begin{center}
\begin{TAB}(r,1cm,0.2cm)[3pt]{|c|c|}{|c|c|c|c|}%
{ $\infty$} & { $\U\setminus(\states{0}\cup\states{1}\cup\states{2}$) }  \\
{ $2$} & {$\peng\bar{\dodo}\bird\flies$ }\\
{ $1$} & {$\bar{\peng}\bar{\dodo}\bird\bar{\flies}$,\hspace{0.2cm} $\peng\bar{\dodo}\bird\bar{\flies}$,\hspace{0.2cm}  }\\
{ $0$} & {$\bar{\peng}\bar{\dodo}\bird\flies$,\hspace{0.2cm} $\bar{\peng}\hspace{0.02cm}\bar{\dodo}\hspace{0.02cm}\bar{\bird}\flies$},\hspace{0.2cm} {$\bar{\peng}\bar{\dodo}\hspace{0.02cm}\bar{\bird}\hspace{0.02cm}\bar{\flies}$} \\
\end{TAB}
\end{center}
\caption{Minimal ranked model of~$\KB^{\land}$ in Example~\ref{Ex:MinimalModel}.}
\label{Figure:PenguinDodoRepeated}
\end{figure}

\begin{figure}[t]
\begin{center}
\begin{TAB}(r,1cm,0.2cm)[3pt]{|c|c|}{|c|c|c|c|c|c|}%
{ $\tuple{\infty,\infty}$} & { $\states{\peng\land\lnot\bird}\cup\states{\dodo\land \lnot\bird}$}  \\
{ $\tuple{\infty,1}$} & { $\bar{\peng}\dodo\bird\flies$,\hspace{0.2cm} $\peng\dodo\bird\flies$}\\
{ $\tuple{\infty,0}$} & {$\bar{\peng}\dodo\bird\bar{\flies}$,\hspace{0.2cm} $\peng\dodo\bird\bar{\flies}$}\\
{ $\tuple{\f,2}$} & {$\peng\bar{\dodo}\bird\flies$ }\\
{ $\tuple{\f,1}$} & {$\bar{\peng}\bar{\dodo}\bird\bar{\flies}$,\hspace{0.2cm} $\peng\bar{\dodo}\bird\bar{\flies}$}\\
{ $\tuple{\f,0}$} & {$\bar{\peng}\bar{\dodo}\bird\flies$,\hspace{0.2cm} $\bar{\peng}\bar{\dodo}\hspace{0.02cm}\bar{\bird}\flies$},\hspace{0.2cm} {$\bar{\peng}\bar{\dodo}\hspace{0.02cm}\bar{\bird}\hspace{0.02cm}\bar{\flies}$} \\
\end{TAB}
\end{center}
\caption{Minimal epistemic model of the knowledge base in Example~\ref{Ex:MinimalModel}.}
\label{Fig:PenguinDodoRepeated}
\end{figure}

The minimal closure of~$\KB$ is defined in terms of the minimum epistemic model of~$\KB$ constructed in this way.

\begin{definition}[Minimal Entailment and Closure]\label{Def:MinimalClosure}
$\alpha\twiddle_{\gamma}\beta$ is \df{minimally entailed} by an SCKB~$\KB$, denoted as $\KB\miniment\alpha\twiddle_{\gamma}\beta$, if $\EI_{\KB}\sat\alpha\twiddle_{\gamma}\beta$, where $\EI_{\KB}$ is the minimal model of~$\KB$. The corresponding closure operation
\[
\C_{m}(\KB)\defined\{\alpha\twiddle_{\gamma}\beta\mid \KB\miniment\alpha\twiddle_{\gamma}\beta\}
\]
is the \df{minimal closure} of~$\KB$.
\end{definition}

\begin{example}\label{Ex:MinimalClosure}
We proceed from Example~\ref{Ex:MinimalModel}. Looking at the model in~Figure~\ref{Fig:PenguinDodoRepeated}, we are able to check what is minimally entailed. For every $\alpha\twiddle_{\gamma}\beta\in\KB$, $\KB\miniment\alpha\twiddle_{\gamma}\beta$. In particular, while $\KB\miniment \dodo\twiddle_{\top}\bot$, we do not have $\KB\miniment\dodo\twiddle_{\dodo}\bot$, that is, it is possible to reason counterfactually about dodos. From the point of view of the actual situation (that is, in the situation~$\top$), we can conclude anything about dodos, since they do not exist. Indeed, we have both $\KB\miniment \dodo\twiddle_{\top}\lnot\flies$ and $\KB\miniment\dodo\twiddle_{\top}\flies$. Nevertheless, we are able to reason coherently about dodos once we assume a point of view in which they would exist. To witness, we have $\KB\miniment\dodo\twiddle_{\dodo}\lnot\flies$, but $\KB\not\miniment \dodo\twiddle_{\dodo}\flies$.
\end{example}

Definition~\ref{Def:ConstructionMinimalModel} shows that the minimal epistemic model can be defined using the minimal ranked models for two sets of defeasible conditionals, $\KB^{\land}$ and~$\KB^{\land}_{\infty\downarrow}$. If a valuation is associated with a finite rank $i$ in the minimal ranked model of $\KB^{\land}$, then we associate to it the corresponding rank $\tuple{\f,i}$ in the minimal epistemic model. All the other valuations, those that have rank $\infty$ in the minimal model of $\KB^{\land}$, will have a rank determined by the minimal ranked model of $\KB^{\land}_{\infty\downarrow}$: for each one of such valuations, if its rank in the minimal model of $\KB^{\land}_{\infty\downarrow}$ is $i$ ($i\in\mathbb{N}\cup\{\infty\}$), it will have the rank $\tuple{\infty,i}$ in the minimal epistemic model.

As we are going to see in the next section, since the construction of the minimal epistemic model relies on the construction of two minimal ranked models, it is possible to decide whether an SC is in the minimal entailment of an SCKB fully relying on a series of propositional decision steps.


\section{Computing entailment from situated conditional knowledge bases}\label{Algorithm}

In this section, we define a procedure to decide whether a conditional is in the minimal closure of an~SCKB. The procedure is described by Algorithm~\ref{Func:MinClosure}, $\mathtt{MinimalClosure}$, and it relies on a series of propositional entailment checks. Hence, it can be implemented on top of any propositional reasoner. 

We will start by looking at Algorithms  $\mathtt{Exceptional}$ (\ref{Func:Exceptional}), $\mathtt{ComputeRanking}$~ (\ref{Func:Ranking}), $\mathtt{Rank}$ (\ref{Func:InfiniteRank}), and $\mathtt{RationalClosure}$ (\ref{Func:RationalClosure}), which formalise known procedures (see the work of Freund~\cite{Freund1998} and of Casini and Straccia~\cite[Section 2]{CasiniStraccia2010}) that together define a decision procedure for rational closure (RC). As indicated in Section~\ref{Preliminaries}, on the semantic side, the~RC of a knowledge base~$\C$ containing defeasible conditionals can be characterised using the \emph{minimal ranked model} $\RM^{\C}_{RC}$~\cite{GiordanoEtAl2015}, that is, $\alpha\twiddle\beta$ is in the RC of a set of defeasible conditionals $\C$ if and only if $\RM^{\C}_{RC}\sat\alpha\twiddle\beta$ (Definition~\ref{Def:RC}).

It has been proved~\cite{Freund1998,CasiniStraccia2010} that $\alpha\twiddle\beta$ is in the RC of $\C$, that is, $\RM^{\C}_{RC}\sat\alpha\twiddle\beta$, if and only if $\mathtt{RationalClosure}(\C,\alpha\twiddle\beta)$ returns $\mathtt{true}$. In what follows, we provide an explanation of all the algorithms involved in the process. We shall often refer to Figure~\ref{Figure:RankedInterpretation} (repeated in Figure~\ref{Figure:RankedInterpretation-Repeated} below for the reader's convenience), which is the minimal ranked model of the knowledge base $\C=\{\bird\twiddle\flies, \peng\twiddle\lnot\flies, \peng\land\lnot\bird\twiddle\bot\}$.

\begin{figure}[ht]
\begin{center}
\begin{TAB}(r,1cm,0.2cm)[3pt]{|c|c|}{|c|c|c|c|}%
 {$\infty$} & {$\bar{\bird}\hspace{0.02cm}\bar{\flies}\p$, \quad $\bar{\bird}\flies\p$} \\
 {$2$} & {$\bird\flies\p$} \\
 {$1$} & {$\bird\bar{\flies}\bar{\p}$, \quad $\bird\bar{\flies}\p$}\\ 
 {$0$} & {$\bar{\bird}\hspace{0.02cm}\bar{\flies}\bar{\p}$, \quad $\bar{\bird}\flies\bar{\p}$, \quad $\bird\flies\bar{\p}$} \\
\end{TAB}
\end{center}
\vspace*{-0.2cm}
\caption{Minimal ranked model of the knowledge base $\C=\{\bird\twiddle\flies, \peng\twiddle\lnot\flies, \peng\land\lnot\bird\twiddle\bot\}$.}
\label{Figure:RankedInterpretation-Repeated}
\end{figure}

\begin{itemize}
    \item $\mathtt{Exceptional}(\C)$ (Algorithm \ref{Func:Exceptional}) takes as input a finite set $\C$ of defeasible conditionals and gives back the exceptional elements, that is, the conditionals $\alpha\twiddle \beta$ \st\ $\top\twiddle\neg\alpha$ holds in the minimal ranked model of~$\C$. 
    For example, from Figure~\ref{Figure:RankedInterpretation-Repeated}, one can check that the conditionals $\peng\twiddle\lnot\flies$ and $\peng\land\lnot\bird\twiddle\bot$ are exceptional, since none of the valuations in layer~$0$ satisfies~$\peng$, and in fact $\mathtt{Exceptional}($\C$)=\{\peng\twiddle\lnot\flies,\peng\land\lnot \bird\twiddle\bot\}$. The procedure fully relies on a series of decision steps in classical propositional logic, since it uses the \emph{materialisation} of the KB~$\C$ (see Section~\ref{Entailment}).
    \item $\mathtt{ComputeRanking}(\C)$ (Algorithm \ref{Func:Ranking}) ranks each conditional in the KB~$\C$ \wrt\ its exceptionality level. $\E_{0}$ contains all the conditionals, $\E_{1}$ the exceptional ones \wrt\ $\E_{0}$, and so on. $\E_{\infty}$ contains the fixed point of the exceptionality procedure, that is, the conditionals having antecedents that cannot be satisfied in any   valuation that is ranked as finite in any ranked model of~$\C$. $\mathtt{ComputeRanking}(\C)$ returns $\E_{0}=\C=\{\bird\twiddle\flies, \peng\twiddle\lnot\flies,\peng\land\lnot\bird\twiddle\bot\}$, $\E_{1}=\{\peng\twiddle\lnot\flies,\peng\land \lnot\bird\twiddle\bot\}$, $\E_{\infty}=\{  \peng\land\lnot\bird\twiddle\bot\}$.
    \item $\mathtt{Rank}(\C,\alpha)$ (Algorithm \ref{Func:InfiniteRank}) decides the rank of a proposition, that is, the lowest rank in the minimal ranked model containing a valuation that satisfies the proposition. For example, the reader can check that $\mathtt{Rank}(\C, \lnot\peng)=0$, $\mathtt{Rank}(\C, \peng)=1$, $\mathtt{Rank}(\C, \peng\land\flies)=2$, $\mathtt{Rank}(\C, \peng\land\lnot\bird)=\infty$, values that, for each of the propositions, correspond exactly to the lowest layer in the minimal ranked model in which there is a valuation satisfying the proposition (see Figure~\ref{Figure:RankedInterpretation-Repeated}). 
    \item $\mathtt{RationalClosure}(\C,\alpha\twiddle\beta)$ (Algorithm \ref{Func:RationalClosure}) tells us whether $\alpha\twiddle\beta$ is in the~RC of~$\C$, that is, whether $\RM^{\C}_{RC}\sat\alpha\twiddle\beta$. For example, $\mathtt{RationalClosure}(\C,\peng\twiddle\lnot\flies)$ is $\mathtt{true}$, since: $\mathtt{Rank}(\C, \peng)=1$,  $\E_{1}=\{ \peng\twiddle\lnot\flies, \peng\land\lnot\bird\twiddle\bot\}$, and $\overline{\E_{1}}\cup\{\peng\}\entails\lnot\flies$.
\end{itemize}

Note that all the procedures fully rely on a series of decision steps in classical propositional logic.

\begin{algorithm}[h]
\SetAlgoLined
\SetKwData{Left}{left}\SetKwData{This}{this}\SetKwData{Up}{up}
\SetKwFunction{Union}{Union}\SetKwFunction{FindCompress}{FindCompress}
\SetKwInOut{Input}{input}\SetKwInOut{Output}{output}
\SetKw{Return}{return}

\Input{a set  of defeasible conditionals $\C$}
\Output{$\E\subseteq\C$ \st\ $\E$ is exceptional \wrt\ $\C$}
\BlankLine
 $\E\assigned\emptyset$ 
 
$\overline{\C}\assigned\{\alpha\rightarrow\beta\mid \alpha\twiddle \beta\in\C\}$ 

\ForEach{$\alpha\twiddle \beta\in\C$}{	
  	\If{$\overline{\C}\entails\neg\alpha$}{$\E\assigned\E\cup\{\alpha\twiddle \beta\}$}
 }
\Return{$\E$}
\caption{$\mathtt{Exceptional}(\C)$}\label{Func:Exceptional}
\end{algorithm}
\begin{algorithm}[h]
\caption{$\mathtt{ComputeRanking}(\C)$\label{Func:Ranking}}

\SetAlgoLined
\SetKwData{Left}{left}\SetKwData{This}{this}\SetKwData{Up}{up}
\SetKwFunction{Union}{Union}\SetKwFunction{FindCompress}{FindCompress}
\SetKwInOut{Input}{input}\SetKwInOut{Output}{output}
\SetKw{Return}{return}
\Input{a set of defeasible conditionals $\C$}
\Output{an exceptionality ranking $r_{\C}$}
\BlankLine
	$i\assigned 0$ 
	
	$\E_{0}\assigned\C$ 
	
	$\E_{1}\assigned \mathtt{Exceptional}(\E_{0}$) 
	
	\While{$\E_{i+1}\neq\E_{i}$}{
		$i\assigned i + 1$ 
		
		$\E_{i+1}\assigned \mathtt{Exceptional}(\E_{i}$) 
	}
	$\E_{\infty}\assigned\E_{i}$ 
	
	$r_{\C}\assigned (\E_0,\ldots,\E_{i-1}, \E_{\infty})$ 
	
\Return{$r_{\C}$}
\end{algorithm}
\begin{algorithm}[h]
\SetAlgoLined
\SetKwData{Left}{left}\SetKwData{This}{this}\SetKwData{Up}{up}
\SetKwFunction{Union}{Union}\SetKwFunction{FindCompress}{FindCompress}
\SetKwInOut{Input}{input}\SetKwInOut{Output}{output}
\SetKw{Return}{return}

\caption{$\mathtt{Rank}(\C,\alpha)$\label{Func:InfiniteRank}}
\Input{a set of defeasible conditionals $\C$, a proposition $\alpha$}
\Output{the rank $rk_{\C}(\alpha)$ of $\alpha$}
\BlankLine
$r_{\C}=(\E_0,\ldots,\E_n,\E_{\infty})\assigned \mathtt{ComputeRanking}(\C)$ 

\ForEach{$0\leq i\leq n$}{
$\overline{\E_{i}}\assigned\{\alpha\limp\beta\mid \alpha\twiddle \beta\in\E_{i}\}$
 }
$\overline{\E_{\infty}}\assigned\{\alpha\limp\beta \mid \alpha\twiddle\beta\in\E_{\infty}\}$

$i\assigned 0$ 

\While {$\overline{\E_{i}}\entails\neg\alpha \text{ and }i\leq n$}
{$i\assigned i + 1$ 
}
\If{$i\leq n$}
	{$rk_{\C}(\alpha)\assigned i$ }
\Else{\If{$\overline{\E_{\infty}}\not\entails\neg\alpha$}{$rk_{\C}(\alpha)\assigned i+1$ }	
        \Else{$rk_{\C}(\alpha)\assigned\infty$ }
        }
\Return{$rk_{\C}(\alpha)$}
\end{algorithm}

\begin{algorithm}[h]
\SetAlgoLined
\SetKwData{Left}{left}\SetKwData{This}{this}\SetKwData{Up}{up}
\SetKwFunction{Union}{Union}\SetKwFunction{FindCompress}{FindCompress}
\SetKwInOut{Input}{input}\SetKwInOut{Output}{output}
\SetKw{Return}{return}

\caption{$\mathtt{RationalClosure}(\C, \alpha\twiddle\beta)$\label{Func:RationalClosure}}
\Input{a set of defeasible conditionals $\C$, a query $\alpha\twiddle\beta$}
\Output{$\mathtt{true}$, if $\C\entails_{RC}\alpha\twiddle\beta$, $\mathtt{false}$ otherwise}
\BlankLine
$r_{\KB}=(\E_0,\ldots,\E_n,\E_{\infty})\assigned \mathtt{ComputeRanking}(\C)$ 

$r\assigned \mathtt{Rank}(\C,\alpha)$ 



\Return{$\overline{\E_{r}}\cup\{\alpha\}\entails\beta$ }
\end{algorithm}

Algorithms $\mathtt{Partition}$ (\ref{Func:Partition}) and $\mathtt{MinimalClosure}$ (\ref{Func:MinClosure}) are novel. They define a procedure to decide minimal entailment $\miniment$, given an~SCKB, and they are built on top of $\mathtt{ComputeRanking}$, $\mathtt{Rank}$, and $\mathtt{RationalClosure}$. Let us go through them:

\begin{itemize}
    \item $\mathtt{Partition}(\KB)$ (Algorithm \ref{Func:Partition}) takes as input an SCKB~$\KB$ and identifies the set $\KB_{\infty}$ and the set of defeasible conditionals $\KB^{\land}_{\infty\downarrow}$, in a way that, as we shall prove, corresponds to Definition~\ref{Def:ConstructionMinimalModel}. That is, $\KB_{\infty}$ is the set of conditionals of which the situations are ranked as infinite \wrt~$\KB^{\land}$.
    \item $\mathtt{MinimalClosure}(\KB,\alpha\twiddle_{\gamma}\beta)$  (Algorithm \ref{Func:MinClosure}) tells us whether $\alpha\twiddle_{\gamma}\beta$ is in the minimal closure of $\KB$. First, the algorithm checks if~$\KB$ is a consistent SCKB (see Definition~\ref{Def:SCKBconsistency}): by Corollary \ref{Corollary:Consistency}, it is sufficient to check whether $\overline{\KB^{\land}}\models\bot$. Then, in case it is consistent, it checks the rank of the situation~$\gamma$. If the situation's rank is finite, then it checks whether the conjunctive form  $\alpha\land\gamma\twiddle\beta$ is in the~RC of~$\KB^{\land}$. Otherwise, it checks whether the conjunctive form $\alpha\land\gamma\twiddle\beta$ is in the~RC of~$\KB^{\land}_{\infty\downarrow}$.
\end{itemize}

\begin{algorithm}[h]
\caption{$\mathtt{Partition}(\KB)$\label{Func:Partition}}

\SetAlgoLined
\SetKwData{Left}{left}\SetKwData{This}{this}\SetKwData{Up}{up}
\SetKwFunction{Union}{Union}\SetKwFunction{FindCompress}{FindCompress}
\SetKwInOut{Input}{input}\SetKwInOut{Output}{output}
\SetKw{Return}{return}

\Input{an SCKB $\KB$}
\Output{the conjunctive forms $\KB^{\land}$ and $\KB_{\infty\downarrow}^{\land}$}
\BlankLine
$\KB^{\land}\assigned\{\alpha\land\gamma\twiddle \beta\mid\alpha\twiddle_{\gamma} \beta\in\KB\}$ 

	$r_{\KB^{\land}}=(\E_0,\ldots,\E_n,\E_{\infty})\assigned \mathtt{ComputeRanking}(\KB^{\land})$ 
	
	$\KB_{\infty}\assigned\emptyset$ 
	
	\ForEach{$\alpha\twiddle_{\gamma} \beta\in\KB$}{	
  	\If{$\mathtt{Rank} (\KB^{\land},\gamma)=\infty $}{$\KB_{\infty}\assigned\KB_{\infty}\cup\{\alpha\twiddle_{\gamma}\beta\}$}
 }
	

$\mu\assigned\bigwedge\{\neg \alpha\mid \alpha\twiddle\beta\in\E_{\infty}\} $ 

$\KB^{\land}_{\infty\downarrow}\assigned\{\alpha\land\gamma\twiddle \beta\mid\alpha\twiddle_{\gamma} \beta\in\KB_{\infty}\}\cup\{\mu\twiddle\bot\}$ 

 \Return{$\KB^{\land},\KB_{\infty\downarrow}^{\land}$}
\end{algorithm}




	
	




\begin{algorithm}[h]
\caption{$\mathtt{MinimalClosure}(\KB,\alpha\twiddle_{\gamma}\beta)$\label{Func:MinClosure}}

\SetAlgoLined
\SetKwData{Left}{left}\SetKwData{This}{this}\SetKwData{Up}{up}
\SetKwFunction{Union}{Union}\SetKwFunction{FindCompress}{FindCompress}
\SetKwInOut{Input}{input}\SetKwInOut{Output}{output}
\SetKw{Return}{return}

\Input{an SCKB $\KB$, a query $\alpha\twiddle_{\gamma}\beta$}
\Output{$\mathtt{true}$, if $\KB\miniment\alpha\twiddle_{\gamma}\beta$, $\mathtt{false}$ otherwise}
\BlankLine
$\KB^{\land},\KB_{\infty\downarrow}^{\land}\assigned \mathtt{Partition(\KB)}$ 
$\overline{\KB^{\land}}=\{(\alpha\land\gamma)\rightarrow\beta\mid\alpha\land\gamma\twiddle\beta\in\KB^{\land}\}$ 

\If{$\overline{\KB^{\land}}\entails\bot$}
    {\Return{$\mathtt{true}$} }
\Else{
    \If{$\mathtt{Rank}(\KB^{\land},\gamma)<\infty$}
    {\Return{$\mathtt{RationalClosure}(\KB^{\land},\alpha\land\gamma\twiddle\beta)$} }
    \Else{\Return{$\mathtt{RationalClosure}(\KB^{\land}_{\infty\downarrow},\alpha\land\gamma\twiddle\beta)$} }
    }
\end{algorithm}


We need to prove that Algorithm~\ref{Func:MinClosure} is complete and correct \wrt~minimal entailment $\miniment$. Before the main theorem, we need to prove the following lemma.

\begin{lemma}\label{lemma_knowledge}
Let $\KB$ be a consistent SCKB, let~$\KB^\land$ be its conjunctive classical form, and let~$\RM$ be the minimal ranked model of~$\KB^\land$. Moreover, let~$\mu$ be defined as in Algorithm~\ref{Func:Partition}, and let~$\fm{\U^{\f}_\RM}$ be as in Definition~\ref{Def:ConstructionMinimalModel}. Then we have that~$\mu$ is logically equivalent to $\fm{\U^{\f}_\RM}$.
\end{lemma}

\begin{proof}
First, we prove that $\fm{\U^{\f}_\RM}\entails\mu$. Let $\alpha\twiddle\beta\in\E_{\infty}$. This implies that $rk_{\KB^\land}(\alpha)=\infty$, that is, all the valuations satisfying~$\alpha$ have rank~$\infty$. That is, $\U^{\f}_\RM\subseteq\states{\neg\alpha}$ for every $\alpha$ s.t. $\alpha\twiddle\beta\in\E_{\infty}$. That implies 
\[
\U^{\f}_\RM\subseteq\bigcap\{\states{\neg\alpha}_{\RM}\mid\alpha
\twiddle\beta\in\E_{\infty}\},
\]
and, consequently, $\fm{\U^{\f}_\RM}\entails\mu$.

Now we prove that $\mu\entails\fm{\U^{\f}_\RM}$. Assume this is not the case. That is, there is a valuation $w\in\U^\infty_\RM$ \st~$w\sat\mu$. Let $n$ be the highest finite rank in~$\RM$, and consider the ranked model~$\RM'$ obtained from~$\RM$ just by re-assigning  the valuation~$w$ from the rank $\infty$ to the rank~$n+1$ (note that if the valuation~$w$ is the only valuation in $\U^{\infty}_{\RM}$ and, consequently, $\U^{\infty}_{\RM'}=\emptyset$, then $\RM'$ is still a ranked interpretation since it is compatible with Definition \ref{Def:RankedInterpretation}). $\RM'$ is preferred to~$\RM$, and it is easy to see that~$\RM'$ is a ranked model of~$\KB$: for every $\alpha\twiddle \beta\in\E_i$, for some $i<\infty$, there is a valuation in a lower rank satisfying $\alpha\land\beta$, while for every $\alpha\twiddle\beta\in\E_\infty$, $w\sat\neg\alpha$, and consequently $w$ is irrelevant \wrt\ the satisfaction of $\alpha\twiddle\beta$ by $\RM'$, since it is not in $\min\states{\alpha}^{\f}_{\RM'}$. Hence, we have that $\RM'\prec_{\KB}\RM$, against the hypothesis that~$\RM$ is the minimal element in~$\prec_{\KB}$, which leads to a contradiction. Therefore, $\mu\models\fm{\U^{\f}_{\RM}}$.
\end{proof}

Now we can state the main result of the present section.

\begin{restatable}{theorem}{restatableTheoremCompleteProcedure}\label{th_complete_procedure}
Let $\KB$ be an SCKB. $\mathtt{MinimalClosure}(\KB,\alpha\twiddle_{\gamma}\beta)$ returns $\mathtt{true}$ iff $\KB\miniment\alpha\twiddle_{\gamma}\beta$.
\end{restatable}

\newcommand{\CK}{\ensuremath{\mathsf{ck}}}
\newcommand{\CL}{\ensuremath{\mathsf{cl}}}
\newcommand{\ST}{\ensuremath{\mathsf{st}}}
\newcommand{\SI}{\ensuremath{\mathsf{si}}}
\newcommand{\CA}{\ensuremath{\mathsf{cb1}}}
\newcommand{\CB}{\ensuremath{\mathsf{cb2}}}

\begin{example}\label{ex_kitchen}
Let us model a more practically-oriented scenario. The agent knows that the Kitchen has been cleaned ($\neg \CK \twiddle_{\top} \bot$), and has a series of (defeasible) expectations: the pan is clean ($\CL$) and positioned in Cupboard1 ($\CA$) ($\top \twiddle_{\top} \CL$ and $\top \twiddle_{\top} \CA$), but in case the pan is in Cupboard2 ($\CB$), the agent will need a stool ($\ST$) to reach the pan ($\CB \twiddle_{\top} \ST$). We can also model the agent’s expectations about counterfactual situations, that is, situations that are not compatible with the information the agent has about the actual situation: if the kitchen has not been cleaned, the pan will presumably be in the sink ($\top\twiddle_{\neg \CK} \SI$), and it will be dirty ($\top\twiddle_{\neg \CK} \neg \CL$). 
Also, we have some constraints that \emph{must necessarily} hold, simply stating that the pan must be in exactly one place: $\neg\CA\wedge \neg \CB\wedge \neg \SI\twiddle_{\neg\CA\wedge \neg \CB\wedge \neg \SI}\bot$, $\CA\wedge \CB\twiddle_{\CA\wedge \CB} \bot$, $\CA\wedge \SI\twiddle_{\CA\wedge \SI} \bot$, $\CB\wedge \SI\twiddle_{\CB\wedge \SI} \bot$. Note that the conditionals $\alpha\twiddle_{\alpha}\bot$ impose that the valuations satisfying $\alpha$ can be placed only in rank $\tuple{\infty,\infty}$, that is, $\neg \alpha$ cannot be falsified, even in the counterfactual situations (see Example~\ref{Ex:PenguinDodo2}).
\end{example}


Let $\KB=\{\neg\CA\wedge \neg \CB\wedge \neg \SI\twiddle_{\neg\CA\wedge \neg \CB\wedge \neg \SI}\bot, \CA\wedge \CB\twiddle_{\CA\wedge \CB} \bot, \CA\wedge \SI\twiddle_{\CA\wedge \SI} \bot, \CB\wedge \SI\twiddle_{\CB\wedge \SI} \bot, \neg \CK \twiddle_{\top} \bot,\top \twiddle_{\top} \CL, \top \twiddle_{\top} \CA, \CB \twiddle_{\top} \ST, \top\twiddle_{\neg \CK} \SI, \top\twiddle_{\neg \CK} \neg \CL\}$ be an~SCKB formalising the scenario in Example~\ref{ex_kitchen}. We apply Algorithm~\ref{Func:Partition}, $\mathtt{Partition}$, to $\KB$:

\begin{itemize}
    \item The algorithm creates the conjunctive form $\KB^{\land}=\{\neg\CA\wedge \neg \CB\wedge \neg \SI\twiddle\bot, \CA\wedge \CB\twiddle \bot, \CA\wedge \SI\twiddle \bot, \CB\wedge \SI\twiddle \bot, \neg \CK \twiddle \bot,\top \twiddle \CL, \top \twiddle \CA, \CB \twiddle \ST, \neg \CK\twiddle \SI, \neg \CK\twiddle \neg \CL\}$ (we have simplified the formulas in the conditionals \wrt\ the definition of $\KB^{\land}$ in Section~\ref{Entailment}, for example substituting formulas $\alpha\land\alpha$ or $\alpha\land\top$ with $\alpha$).
    
    \item Calling algorithm $\mathtt{ComputeRanking}$, we rank $\KB^{\land}$ in $\E_0=\{\top \twiddle \CL, \top \twiddle \CA\}\cup \E_1$, $\E_1=\{\CB \twiddle \ST\}\cup \E_\infty$, $\E_\infty=\{\neg\CA\wedge \neg \CB\wedge \neg \SI\twiddle\bot, \CA\wedge \CB\twiddle \bot, \CA\wedge \SI\twiddle \bot, \CB\wedge \SI\twiddle \bot, \neg \CK \twiddle \bot,\neg \CK\twiddle \SI, \neg \CK\twiddle \neg \CL\}$.
    
    \item We then call the procedure $\mathtt{Rank}(\KB^{\land},\gamma)$ for every formula~$\gamma$ appearing in some conditional $\alpha\twiddle\beta$ in $\KB$. It turns out that $\mathtt{Rank}(\KB^{\land},\gamma)=\infty$ for $\gamma\in\{\neg\CA\wedge \neg \CB\wedge \neg \SI, \CA\wedge \CB, \CA\wedge \SI, \CB\wedge \SI, \neg \CK\}$. Consequently, $\KB_{\infty}=\{\neg\CA\wedge \neg \CB\wedge \neg \SI\twiddle_{\neg\CA\wedge \neg \CB\wedge \neg \SI}\bot, \CA\wedge \CB\twiddle_{\CA\wedge \CB} \bot, \CA\wedge \SI\twiddle_{\CA\wedge \SI} \bot, \CB\wedge \SI\twiddle_{\CB\wedge \SI} \bot, \top\twiddle_{\neg \CK} \SI, \top\twiddle_{\neg \CK} \neg \CL\}$.
    
    \item Eventually, the algorithm constructs the set $\KB^{\land}_{\infty\downarrow}$: first, from $\E_{\infty}$, we define~$\mu$ as $\bigwedge\{\CA\lor  \CB\lor  \SI, \neg\CA \lor \neg\CB, \neg\CA \lor \neg\SI, \neg\CB \lor \neg\SI, \CK\}$; then we set $\KB^{\land}_{\infty\downarrow}$ as $\{\neg\CA\wedge \neg \CB\wedge \neg \SI\twiddle\bot, \CA\wedge \CB\twiddle \bot, \CA\wedge \SI\twiddle \bot, \CB\wedge \SI\twiddle \bot,\neg \CK\twiddle \SI, \neg \CK\twiddle \neg \CL,\mu\twiddle\bot\}$.
\end{itemize}

Once we have $\KB^{\land}$ and $\KB^{\land}_{\infty\downarrow}$, we can give queries to  Algorithm \ref{Func:MinClosure} ($\mathtt{MinimalClosure}$). For example, we can check whether the agent should  expect  the pan to be in the sink ($\top\twiddle_{\top} \SI$).

\begin{itemize}
    \item Given $\KB^{\land}$, we define its materialisation $\overline{\KB^{\land}}$, which contains the implications $(\alpha\land\gamma)\limp \beta$ corresponding to the conditionals $\alpha\land\gamma\twiddle \beta$ in $\KB^{\land}$. Using $\overline{\KB^{\land}}$, the algorithm checks whether the  knowledge base~$\KB$ is inconsistent by checking whether $\overline{\KB^{\land}}\entails\bot$ (the reader can check that it is not the case.)
    \item We then have to check the rank of the situation $\top$ in $\top\twiddle_{\top} \SI$, which, being $\top$, must be~$0$. Hence, semantically, since $\top$ cannot be an exceptional proposition, $\top\twiddle_{\top} \SI$ is a conditional whose satisfaction needs to be checked \wrt\ the valuations in the finite ranks of the minimal epistemic model of $\KB$, in particular, \wrt\ the valuations in the rank $\tuple{\f,0}$. This corresponds to checking in Algorithm $\mathtt{MinimalClosure}$ whether $\top\twiddle \SI$ is in the rational closure of $\KB^{\land}$. That is, whether $\mathtt{RationalClosure}(\KB^{\land},\top\twiddle \SI)$ returns $\mathtt{true}$.
    
    In the procedure $\mathtt{RationalClosure}(\KB^{\land},\top\twiddle \SI)$, the rank $0$ is associated to $\top$, and $\E_0=\KB^{\land}$. Consequently, $\top\twiddle \SI$ is in the rational closure of $\KB^{\land}$ if and only if $\overline{\E_0}\models \SI$, which is not the case. Actually, we have that $\top\twiddle_{\top}\neg\SI$ is in the minimal closure of $\KB$, since, due to the presence of $\top \limp \CA$ and $(\CA\wedge \SI)\limp \bot$ in $\overline{\E_0}$, we have $\overline{\E_0}\models \neg \SI$.
\end{itemize}

We now consider a counterfactual situation, checking whether the agent believes that, in case the kitchen has not been cleaned, the pan is not in Cupboard2 ($\top\twiddle_{\neg \CK} \neg \CB$).

\begin{itemize}
    \item As for the previous query, the algorithm starts by checking whether $\KB$ is consistent.
    \item We then have to check the rank of the situation $\neg \CK$ in $\top\twiddle_{\neg \CK} \neg \CB$. Since in $\KB$ we have the conditional $\neg\CK\twiddle_{\top}\bot$, that is, the agent knows that the kitchen has been cleaned, the immediate conclusion is that $\mathtt{Rank}(\KB^{\land})=\infty$.
    \item Hence, semantically, $\top\twiddle_{\neg \CK} \neg \CB$ is a conditional that needs to be checked \wrt\ the valuations in the infinite ranks of the minimal epistemic model of $\KB$. 
    This corresponds to checking whether $ \neg\CK\twiddle  \neg \CB$ follows from $\KB^{\land}_{\infty\downarrow}$, that is, whether $\mathtt{RationalClosure}(\KB^{\land}_{\infty\downarrow},\neg\CK\twiddle \neg \CB)$ returns $\mathtt{true}$. 
     $\mathtt{RationalClosure}(\KB^{\land}_{\infty\downarrow},\neg\CK\twiddle \neg \CB)$ associates the rank $0$ to $\neg\CK$, and   $\E_0=\KB^{\land}_{\infty\downarrow}$. Consequently, $\neg\CK\twiddle \neg \CB$ is in the rational closure of $\KB^{\land}_{\infty\downarrow}$ if and only if $\overline{\KB^{\land}_{\infty\downarrow}}\cup\{\neg\CK\}\models \neg \CB$, which is the case, since $\overline{\KB^{\land}_{\infty\downarrow}}$ contains $\neg \CK\limp \SI$ and $\CB\land\SI\limp\bot$.
\end{itemize}

\subsection{Computational complexity of minimal entailment}

We now turn our attention to the computational complexity of deciding minimal entailment. We have seen that the entire procedure can be reduced to a sequence of classical propositional entailment tests, with propositional entailment known to be co-NP-complete. Therefore, we have to check, given an SCKB  as input, how many classical entailment tests are required in the worst case. We examine each algorithm in turn.

\begin{itemize}
    \item Given a set of defeasible conditionals $\C$, Algorithm $\mathtt{Exceptional}$ performs $|\C|$ propositional entailment tests.
    
    \item Given a set of defeasible conditionals $\C$, Algorithm $\mathtt{ComputeRanking}$ runs the algorithm $\mathtt{Exceptional}$ at most $|\C|$ times in the case where each conditional  from $\C$ has a distinct antecedent, and each rank contains exactly one conditional. In such a case, we have that the first iteration of the algorithm $\mathtt{Exceptional}$ performs $|\C|$ entailment checks, the second one $|\C|-1$ entailment checks, the third one $|\C|-2$ entailment checks, and so on. That is, the $i$-th iteration of $\mathtt{Exceptional}$ performs $|\C|-i+1$ propositional entailment checks. So there are fewer than $|\C|^2$ entailment checks and hence Algorithm $\mathtt{ComputeRanking}$ performs a polynomial number of propositional entailment checks. Note that, given a conditional knowledge base~$\C$, we need to run $\mathtt{ComputeRanking}$ only once. 
    
    \item Given a set of defeasible conditionals $\C$ and a formula $\alpha$, Algorithm $\mathtt{Rank}$ calls $\mathtt{ComputeRanking}$  (which performs at most $|\C|^2$ entailment checks), and then performs at most a number of entailment checks that corresponds to the number of ranks, which is $|\C|$ at most. Hence Algorithm $\mathtt{Rank}$ performs a polynomial number of propositional entailment checks. 
    
    \item Given a set of defeasible conditionals $\C$ and a conditional $\alpha\twiddle\beta$, Algorithm $\mathtt{RationalClosure}$ calls Algorithm $\mathtt{ComputeRanking}$ once and Algorithm $\mathtt{Rank}$ once, plus it makes a final entailment check. Hence, the algorithm  performs a polynomial number of propositional entailment checks.  
    
    \item Given an SCKB $\KB$, Algorithm $\mathtt{Partition}$ runs Algorithm $\mathtt{ComputeRanking}$ once and  Algorithm $\mathtt{Rank}$ at most $|\KB|$ times. Since $|\KB^{\land}|=|\KB|$, running $\mathtt{ComputeRanking}$ consists of $|\KB|^2$ entailment checks at most. The same holds for each run of $\mathtt{Rank}$. Hence running $\mathtt{Partition}$ consists of at most $|\KB|^2\cdot(|\KB|+1)= |\KB|^3+|\KB|^2$ entailment checks.
    
    \item Given an SCKB $\KB$ and a situated conditional $\alpha\twiddle_{\gamma}\beta$, Algorithm $\mathtt{MinimalClosure}$ runs Algorithm $\mathtt{Partition}$ once, followed by one entailment check (line 2), one call to Algorithm $\mathtt{Rank}$ and one call to algorithm $\mathtt{RationalClosure}$ (with either $\KB^{\land}$ or $\KB^{\land}_{\infty,\downarrow}$ as argument):
    \begin{itemize}
        \item $\mathtt{Partition}$ performs at most $|\KB|^3+|\KB|^2$ entailment checks.
        \item $\mathtt{Rank}$ performs at most $|\KB|^2$ entailment checks.
        \item $\mathtt{RationalClosure}$ performs at most $|\KB|^2$ entailment checks.
    \end{itemize}
     Hence Algorithm $\mathtt{MinimalClosure}$ performs a polynomial number of propositional entailments checks. 

\end{itemize}
In summary then, deciding minimal entailment using Algorithm $\mathtt{MinimalClosure}$ involves a polynomial number of propositional entailment checks, and is therefore in $\textsc{P}^{\textsc{coNP}}=\Delta^\textsc{P}_2$. Whether this decision problem is $\Delta^\textsc{P}_2$-complete is currently an open question.

\section{Related work}\label{RelatedWork}

With regard to the distinction between a plausible and an implausible state of affairs, a similar distinction has been used by Booth~\etal.~\cite{BoothEtAl14}, where some pieces of information are considered \emph{credible} while others are not,  and a new piece of information is accepted only if it is credible. Nonetheless, in case it is not credible, its plausibility increases every time such a piece of information is iterated. 
The distinction between plausible and implausible valuations links such an approach with our proposal, but the reasoning problems they model are different. Booth~\etal.~deal with the credibility of a new piece of information, also considering whether the agent is repeatedly exposed to such a piece of information. Here we deal with the distinction between expectations and counterfactuals: given an SC $\alpha\twiddle_{\gamma}\beta$, we could say that if $\gamma$ is \emph{credible}, then the defeasible conditional $\alpha\twiddle\beta$ is evaluated \wrt~one ranked interpretation (represented by the finite ranks), while it is evaluated \wrt~another ranked interpretation (represented by the infinite ranks) otherwise.

The connection between our conditional system and belief change is already made quite clear by the \emph{situated AGM postulates} (See Section \ref{CC}), but it still needs to be properly investigated. Such an investigation should proceed not only from the point of view of the possible definition of interesting revision operators corresponding to our situated conditionals (for example, via some modified version of the Ramsey test) but also from the point of view of the definition of appropriate revision operators modelling the dynamics of SCKB's, in line with what has been done for conditional knowledge bases \cite{CasiniMeyer2017,CasiniEtAl2018b}. With respect to this latter problem, the work of Booth~\etal.~\cite{BoothEtAl14} offers an interesting perspective on modelling the dynamics of a semantics with plausible and implausible state of affairs.
\myskip

The literature on the notion of context, which is akin to our use of situation, is vast, and several formalisations and applications of it have been studied across many areas within~AI~\cite{BikakisAntoniou2010,GhidiniGiunchiglia2001,HomolaSerafini2012,KlarmanGutierrez2013,PinoPerez99}. The role of context in conditional-like statements has been explored recently, in particular in defeasible reasoning over description logic ontologies and within semantic frameworks that are closely related to ours. Britz and Varzinczak~\cite{BritzVarzinczak2018-FoIKS,BritzVarzinczak2019-AMAI}, for example, have put forward a notion of defeasible class inclusion parameterised by atomic roles. Their semantics allows for multiple preference relations on objects, which is more general than our single-preference approach, and allows for objects to be compared in more than one way. This makes normality (or typicality) context-dependent and gives more flexibility from a modelling perspective. Giordano and Gliozzi~\cite{GiordanoGliozzi2018} consider reasoning about multiple aspects in defeasible description logics where the notion of aspect (or context) is linked to concept names (alias, atoms) also in a multi-preference semantics.

When compared with our framework, neither of the above-mentioned approaches allows for reasoning about objects that are `forbidden' by the background knowledge. In that respect, our proposal is complementary to theirs, and a contextual form of class inclusion along the lines of the ternary~$\twiddle$ here studied, with potential applications going beyond that of defeasible reasoning in ontologies, is worth exploring as future work.

\section{Concluding remarks}\label{Conclusion}

In this paper, we have made the case for the provision of a simple situated form of conditional. We have shown, using a number of representative examples, that it is sufficiently general to be used in several application domains. The proposed situated conditionals have an intuitive semantics which is based on a semantic construction that has proved to be quite useful in the area of belief change, and is more general and also more fine-grained than the standard preferential semantics. We also showed that the proposed conditionals can be described in terms of a set of postulates. We provided a representation result, showing that the postulates capture exactly the constructions obtained from the proposed semantics. An analysis in terms of the postulates shows that these situated conditionals are suitable for knowledge representation and reasoning, in particular when reasoning about information that is incompatible with background knowledge. 

With the basic semantic structures in place, we then proceeded to define a form of entailment for situated conditional knowledge bases that is based on the widely-accepted notion of rational closure for KLM-style reasoning. Moreover, we showed that, like rational closure, entailment for situated conditional knowledge bases is reducible to classical propositional reasoning. 
\myskip

\newcommand{\Un}{\ensuremath{\mathsf{u}}}

Note that the semantics we have proposed in the present work can easily be refined further. Our framework allows only for the distinction between the \emph{plausible} situations (the valuations with a finite rank that, with different degrees of expectation, define the agent's beliefs), and the \emph{implausible} ones (the valuations with an infinite rank that are not compatible with the agent's beliefs but are still conceivable with different levels of expectation). All other valuations have the \emph{inconceivable} rank $\tuple{\infty,\infty}$. 

There is a fairly straightforward way of refining the framework by allowing for different ranks of the kind $\tuple{\infty_1,i}$, $\tuple{\infty_2,i}$, etc., ($i\in \mathbb{N}$). To illustrate the point, assume we add \emph{unicorns} (\Un) to our vocabulary, and consider the interpretation in Figure \ref{Fig:Unicorn}. In such a model, we would be able to represent the fact that we believe that unicorns would not exist even if we move to situations in which dodos exist ($\Un\twiddle_{\dodo}\bot$), represented by the ranks $\tuple{\infty_{1},0}$ and $\tuple{\infty_{1},1}$. We would also be able to move to a further level of implausibility ($\infty_2$) in which the existence of unicorns is considered, making it possible to reason coherently about them. For now, we shall leave such a refinement of our semantic framework for future work.

\begin{figure}[t]
\begin{center}
\begin{TAB}(r,1cm,0.2cm)[3pt]{|c|c|}{|c|c|c|c|c|c|c|}
{$\tuple{\infty,\infty}$} & {$\states{\peng\land\lnot\bird}\cup\states{\dodo\land \lnot\bird}$} \\
{$\tuple{\infty_2,0}$} & {$\states{\Un}\setminus(\states{\peng\land\lnot\bird}\cup\states{\dodo\land \lnot\bird})$} \\
{ $\tuple{\infty_1,1}$} & { $\bar{\peng}\dodo\bird\flies\bar{\Un}$,\hspace{0.2cm} $\peng\dodo\bird\flies\bar{\Un}$}\\
{ $\tuple{\infty_1,0}$} & {$\bar{\peng}\dodo\bird\bar{\flies}\bar{\Un}$,\hspace{0.2cm} $\peng\dodo\bird\bar{\flies}\bar{\Un}$}\\
{ $\tuple{\f,2}$} & {$\peng\bar{\dodo}\bird\flies\bar{\Un}$ }\\
{ $\tuple{\f,1}$} & {$\bar{\peng}\bar{\dodo}\bird\bar{\flies}\bar{\Un}$,\hspace{0.2cm} $\peng\bar{\dodo}\bird\bar{\flies}\bar{\Un}$}\\
{ $\tuple{\f,0}$} & {$\bar{\peng}\bar{\dodo}\bird\flies\bar{\Un}$,\hspace{0.2cm} $\bar{\peng}\hspace{0.02cm}\bar{\dodo}\hspace{0.02cm}\bar{\bird}\flies\bar{\Un}$},\hspace{0.2cm} {$\bar{\peng}\hspace{0.02cm}\bar{\dodo}\hspace{0.02cm}\bar{\bird}\hspace{0.02cm}\bar{\flies}\hspace{0.02cm}\bar{\Un}$}
\end{TAB}
\end{center}
\caption{An example of an interpretation extending the epistemic interpretation with further infinite levels for more complex counterfactual reasoning.}
\label{Fig:Unicorn}
\end{figure}

The work described in this paper assumes classical propositional logic as the underlying logical formalism, but it is worthwhile to consider extending this to other, more expressive logics. In this regard, an extension to Description Logics is perhaps an obvious starting point, particularly since rational closure has already been reformulated for this case~\cite{GiordanoEtAl2015,Bonatti2019,CasiniStraccia2010,BritzVarzinczak2019-AMAI}. A different kind of extension of the work presented here is one in which other forms of entailment are investigated. For this, the obvious initial candidate is lexicographic closure~\cite{Lehmann1995} and its  variants~\cite{CasiniEtAl2014,CasiniEtAl2019-JELIA,CasiniStraccia2013}. More generally, we intend to investigate an extension to the class of entailment relations studied by Casini~\etal.~\cite{CasiniEtAl2019-JELIA}.

\section*{Acknowledgments}

The work of Giovanni Casini was partially supported by TAILOR (Foundations of Trustworthy AI Integrating Reasoning, Learning and Optimization), a project funded by EU Horizon 2020 research and innovation programme under GA No 952215.

Ivan Varzinczak has been partially supported by: (\emph{i})~the ANR Chaire~IA BE4musIA: BElief change FOR better MUlti-Source Information Analysis; (\emph{ii})~the CNRS/Royal Society project \emph{Non-Classical Reasoning for Enhanced Ontology-based Semantic Technologies}, and (\emph{iii})~the project AGGREEY ANR-22-CE23-0005 of the French National Research Agency~(ANR).

Thanks to the anonymous reviewers for their helpful suggestions.


\bibliographystyle{elsarticle-num}
\bibliography{References}


\appendix

\section{Proofs}

\restatableTheoremBSC*

\begin{proof}
Consider any epistemic interpretation~$\EI$ and pick any~$\gamma\in\Lang$. We consider three disjoint and covering cases. 

Case 1: If $\U^{\f}_{\EI}\cap\states{\gamma}\neq\emptyset$, then define~$\RM$ from~$\EI$ as follows: (\emph{i})~for all $u\in\U^{\f}_{\EI}\cap\states{\gamma}$, $\RM(u)\defined i$, where $\EI(u)=\tuple{\f,i}$; (\emph{ii})~for all $u\in\U\setminus\U^{\f}_{\EI}\cap\states{\gamma}$, $\RM(u)\defined\infty$. It follows from Definition~\ref{Def:SatisfactionEpistemicInterpretation} and the definition  of satisfaction of~$\twiddle$-statements in ranked interpretations that 
$\EI\sat\alpha\twiddle_{\gamma}\beta$ iff $\RM\sat\alpha\land\gamma\twiddle\beta$. From Theorem~\ref{Theorem:RepresentationResult}, it follows that the~$\twiddle$ generated by~$\RM$ satisfies the original~KLM postulates. Hence, it follows that~$\twiddle_{\gamma}$ satisfies the situated rationality postulates.

Case 2: If $\U^{\f}_{\EI}\cap\states{\gamma}=\emptyset$ but $\U^{\infty}_{\EI}\cap\states{\gamma}\neq\emptyset$, then define~$\RM$ from~$\EI$ as follows: (\emph{i})~for all $u\in\U^{\infty}_{\EI}\cap\states{\gamma}$, $\RM(u)\defined i$, where $\EI(u)=\tuple{\infty,i}$; (\emph{ii})~for all $u\in\U\setminus(\U^{\infty}_{\EI}\cap\states{\gamma})$, $\RM(u)\defined\infty$. 
It follows from Definition~\ref{Def:SatisfactionEpistemicInterpretation} and the definition of satisfaction for~$\twiddle$-statements in ranked interpretations that 
$\EI\sat\alpha\twiddle_{\gamma}\beta$ iff $\RM\sat\alpha\twiddle\beta$. From Theorem~\ref{Theorem:RepresentationResult}, it follows that the~$\twiddle$ generated by~$\RM$ satisfies the original~KLM postulates. For this specific $\gamma$ it then follows that~$\twiddle_{\gamma}$  satisfies the situated rationality postulates.

Case 3: If $\states{\gamma}\subseteq\U\setminus(\U^{\f}_{\EI}\cup\U^{\infty}_{\EI})$, then $\RM(u)\defined\infty$ for all $u\in\states{\gamma}$. Again, it follows from Definition~\ref{Def:SatisfactionEpistemicInterpretation} and the definition of satisfaction for~$\twiddle$ in ranked interpretations that 
$\EI\sat\alpha\twiddle_{\gamma}\beta$ iff $\RM\sat\alpha\twiddle\beta$. From Theorem~\ref{Theorem:RepresentationResult}, it follows that the~$\twiddle$ generated by~$\RM$ satisfies the original~KLM postulates.  For this specific $\gamma$, it then follows that~$\twiddle_{\gamma}$  satisfies the situated rationality postulates.

Putting the three cases above together, it then follows immediately that the situated conditional~$\twiddle_{\gamma}$ obtained from~$\EI$ satisfies the situated rationality postulates. 
\medskip

Now, in order to show that the converse does not hold, consider the language generated from (and only)~$\{\ce{p},\ce{q}\}$. Note first that there is a ranked interpretation~$\RM$ such that~$\RM\sat\alpha\twiddle\beta$ iff~$\ce{p}\land\ce{q}\land\alpha\entails\beta$. From Theorem~\ref{Theorem:RepresentationResult}, it follows that~$\twiddle$, defined in this way, is a rational conditional, and therefore satisfies the situated~KLM postulates. Similarly, there is a ranked interpretation~$\RM'$ such that~$\RM'\sat\alpha\twiddle\beta$ iff~$\ce{p}\land\ce{q}\land\alpha\entails\beta$. From Theorem~\ref{Theorem:RepresentationResult}, it follows that~$\twiddle$, defined in this way, is a rational conditional, and therefore satisfies the situated~KLM postulates. Now, define a situated conditional by letting~$\alpha\twiddle_{\ce{p}}\beta$ iff~$\ce{p}\land\ce{q}\land\alpha\entails\beta$, and~$\alpha\twiddle_{\gamma}\beta$ iff~$\alpha\entails\beta$, for every~$\gamma$ other than~$\ce{p}$. It then follows immediately that this situated conditional is a~BSC. However, it is easy to see that it cannot be generated by an epistemic interpretation. To see why, observe that~$\ce{p}\twiddle_{\ce{p}}\ce{q}$, but that~$\ce{p}\ntwiddle_{\ce{p\lor p}}\ce{q}$.
\end{proof}

\restatableTheoremFSC*

\begin{proof}
Let $\EI$ be an epistemic interpretation and let $\gamma\in\Lang$. Suppose~$\U^{\f}_{\EI}\cap\states{\gamma}\neq\emptyset$. Then if~$\EI\sat\alpha\twiddle_{\gamma}\beta$, it follows by Definition~\ref{Def:SatisfactionEpistemicInterpretation} that $\EI\sat\alpha\land\gamma\twiddle_{\top}\beta$. On the other hand, if $\U^{\f}_{\EI}\cap\states{\gamma}=\emptyset$, then $\EI\sat\alpha\land\gamma\twiddle_{\top}\beta$. This means that the situated conditional $\twiddle$ obtained from~$\EI$ as follows satisfies~Inc: $\alpha\twiddle_{\gamma}\beta$ iff $\EI\sat\alpha\twiddle_{\gamma}\beta$.

Suppose $\EI\nsat\top\twiddle_{\top}\lnot\gamma$. This means that~$\U^{\f}_{\EI}\cap\states{\gamma}\neq\emptyset$. Then if $\EI\sat\alpha\land\gamma\twiddle_{\top}\beta$, it follows by Definition~\ref{Def:SatisfactionEpistemicInterpretation} that $\EI\sat\alpha\twiddle_{\gamma}\beta$.
This means that the situated conditional $\twiddle$ obtained from~$\EI$ as follows satisfies~Vac: $\alpha\twiddle_{\gamma}\beta$ iff $\EI\sat\alpha\twiddle_{\gamma}\beta$.

That the situated conditional obtained from~$\EI$ as follows satisfies Ext follows immediately from Definition~\ref{Def:SatisfactionEpistemicInterpretation}: $\alpha\twiddle_{\gamma}\beta$ iff $\EI\sat\alpha\twiddle_{\gamma}\beta$.

For SupExp we consider two cases. For Case~1, if $\U^{\f}_{\EI}\cap\states{\gamma\land\delta}\neq\emptyset$, then the result follows easily. For Case~2, suppose $\U^{\f}_{\EI}\cap\states{\gamma\land\delta}=\emptyset$. If $\U^{\f}_{\EI}\cap\states{\delta}=\emptyset$, then the result follows easily. Otherwise the result follows from the fact that~$\U^{\f}_{\EI}\cap\states{\alpha\land\gamma\land\delta}=\emptyset$.

For SubExp, suppose that~$\EI\sat\delta\twiddle_{\top}\bot$. This means $\EI\sat\alpha\land\gamma\twiddle_{\delta}\beta$ implies that~$\U^{\infty}_{\EI}\cap\states{\alpha\land\gamma\land\delta}\subseteq\states{\beta}$, from which it follows that~$\EI\sat\alpha \twiddle_{\gamma\wedge\delta}\beta$.
\medskip

For the converse, consider any~FSC~$\twiddle$. We construct an epistemic interpretation~$\EI$ as follows. First, consider~$\twiddle_{\top}$. Since it satisfies the situated~KLM postulates, there is a ranked interpretation~$\RM$ such that~$\RM\sat\alpha\twiddle\beta$ iff~$\alpha\twiddle_{\top}\beta$. We set $\U^{\f}_{\EI}\defined\U^{\f}_{\RM}$, and for all $u\in\U^{\f}_{\EI}$, we let $\EI(u)\defined\tuple{\f,\RM(u)}$. Next, let $\U'\defined\U\setminus\U^{\f}_{\EI}$. Let $k^{\f}$ be a formula such that $\states{k^{\f}}=\U^{\f}_{\EI}$. Similarly, let $k^{\infty}$ be a formula such that $\states{k^{\infty}}=\U'$. Now, consider $\twiddle_{k^{\infty}}$. Since it satisfies the situated~KLM postulates, there is a ranked interpretation $\RM'$ such that $\RM'\sat\alpha\twiddle\beta$ iff $\alpha\twiddle_{k^{\infty}}\beta$. We let $\U^{\infty}_{\EI}\defined\{ u\in\U' \mid \RM'(u)\neq\infty\}$, and for all $u\in\U'$, we let $\EI(u)\defined\tuple{\infty,\RM'(u)}$. Observe that for some $u\in\U'$ it may be the case that $\EI(u)=\tuple{\infty,\infty}$, which means that for such a~$u$, $u\notin\U^{\infty}_{\EI}$. It is easily verified that~$\EI$ is indeed an epistemic interpretation. Next we show that~$\alpha\twiddle_{\gamma}\beta$ iff $\EI\sat\alpha\twiddle_{\gamma}\beta$. We do so by considering two cases.

Case~1: $\U^{\f}_{\EI}\cap\states{\gamma}\neq\emptyset$. Note first that it follows easily from the construction of~$\EI$ that~$\alpha\twiddle_{\top}\beta$ iff  $\EI\sat\alpha\twiddle_{\top}\beta$. Suppose~$\alpha\twiddle_{\gamma}\beta$. By~Inc, $\alpha\land\gamma\twiddle_{\top}\beta$ and therefore~$\EI\sat\alpha\land\gamma\twiddle_{\top}\beta$, and~$\EI\sat\alpha\twiddle_{\gamma}\beta$, by definition. Conversely, suppose~$\EI\sat\alpha\twiddle_{\gamma}\beta$. Then by definition, $\EI\sat\alpha\land\gamma\twiddle_{\top}\beta$, and therefore  $\alpha\land\gamma\twiddle_{\top}\beta$. Since~$\top\ntwiddle_{\top}\lnot\gamma$, it then follows from~Vac that~$\alpha\twiddle_{\gamma}\beta$.

Case~2: $\U^{\f}_{\EI}\cap\states{\gamma}\neq\emptyset$. By the construction of~$\EI$, it follows that~$\alpha\twiddle_{k^{\infty}}\beta$ iff $\EI\sat\alpha\twiddle_{k^{\infty}}\beta$. Suppose $\alpha\twiddle_{\gamma}\beta$. Note that~$\gamma\equiv k^{\infty}$. By~Ext, $\alpha\twiddle_{\gamma\land k^{\infty}}\beta$ and so, by~SupExp, $\alpha\land\gamma\twiddle_{k^{\infty}}\beta$. It then follows that~$\EI\sat\alpha\land\gamma\twiddle_{k^{\infty}}\beta$ and, by Definition~\ref{Def:SatisfactionEpistemicInterpretation}, that $\EI\sat\alpha\twiddle_{\gamma}\beta$. Conversely, suppose that $\EI\sat\alpha\twiddle_{\gamma}\beta$. Then $\EI\sat\alpha\land\gamma\twiddle_{k^{\infty}}\beta$, by Definition~\ref{Def:SatisfactionEpistemicInterpretation}, and, therefore, using Ext, we have~$\alpha\land\gamma\twiddle_{\gamma\land k^{\infty}}\beta$. Note that~$\EI\sat k^{\f}\twiddle_{\top}\bot$ and therefore~$k^{\f}\twiddle_{\top}\bot$. By~SubExp it then follows that~$\alpha\twiddle_{\gamma\land k^{\infty}}\beta$, and by~Ext that~$\alpha\twiddle_{\gamma}\beta$ holds.
\end{proof}

\restatableLemmaShifting*

\begin{proof}
In case $\alpha\land \gamma$ is not logically consistent, the lemma holds since $\EI\sat\alpha\twiddle_{\gamma}\beta$ and $\EI_{\downarrow}^{\infty}\sat\alpha\land\gamma\twiddle\beta$ for any $\beta$. Hence we assume that $\alpha\land \gamma$ is logically  consistent.

Let~$\EI\sat\lnot\gamma$, that is, there are no valuations in the finite ranks satisfying~$\gamma$. Then the satisfaction of the conditionals with situation~$\gamma$ must be checked, referring to the valuations that are ranked as infinite.  $\EI\sat\alpha\twiddle_{\gamma}\beta$ implies two possible situations: either there are some valuations among the ones in~$\states{\gamma}$ that are ranked as infinite and satisfy~$\alpha\land\gamma$, and among them, all the minimal ones satisfy also~$\beta$; or all the valuations satisfying~$\alpha\land\gamma$ have rank~$\tuple{\infty,\infty}$. $\gamma$ has finite rank in~$\EI_{\downarrow}^{\infty}$, or the rank~$\tuple{\infty,\infty}$. In the latter case, we have~$\EI_{\downarrow}^{\infty}\sat\alpha\land \gamma\twiddle\beta$. In the former case, the rank of~$\gamma$ in~$\EI$ is~$\tuple{\infty,i}$, with~$i<\infty$, that is, the rank of~$\gamma\land\alpha$ in~$\EI_{\downarrow}^{\infty}$ is~$\tuple{\f,j}$, for some~$j$ \st\ $i\leq j<\infty$, or~$\tuple{\infty,\infty}$. In the latter case, again, it is straightforward to conclude~$\EI_{\downarrow}^{\infty}\sat\alpha\land\gamma\twiddle\beta$. In the former case, we have $\EI\sat\alpha\twiddle_{\gamma}\beta$, and the construction of~$\EI_{\downarrow}^{\infty}$ imposes that the minimal valuations in~$\states{\alpha\land\gamma}$ satisfy also~$\beta$, that is, $\EI_{\downarrow}^{\infty}\sat\alpha\land\gamma\twiddle\beta$.


The proof is analogous in the opposite direction. Let~$\EI\sat\lnot\gamma$ and~$\EI{\downarrow}^{\infty}\sat\alpha\land\gamma\twiddle\beta$. Either the minimal valuations in~$\EI{\downarrow}^{\infty}$ satisfying~$\alpha\land \gamma$ are in rank~$\tuple{\f,i}$, for some $i<\infty$, and they all satisfy~$\beta$  (Case 1), or they are in~$\tuple{\infty,\infty}$  (Case 2). Since $\EI\sat\lnot\gamma$, in $\EI$ all the valuations satisfying $\gamma$ are in $\U^{\infty}_\EI \cup\tuple{\infty,\infty}$, and consequently the satisfaction of the SC $\alpha\twiddle_{\gamma}\beta$ needs to be judged considering the valuations in $\U^{\infty}_\EI \cup\tuple{\infty,\infty}$. If we are in Case 1, since a valuation $w$ has rank~$\tuple{\f,i}$ in~$\EI{\downarrow}^{\infty}$ iff $w$ has rank~$\tuple{\infty,i}$ in~$\EI$ (see the definition of \emph{counterfactual shifting} in Section \ref{Entailment}), we have that the minimal valuations in~$\EI$ satisfying~$\alpha\land \gamma$ have rank~$\tuple{\infty,i}$, for some $i<\infty$, and they all satisfy~$\beta$. If we are in Case 2, since $\EI\sat\lnot\gamma$, a valuation $w$ that satisfies $\gamma$ can have rank~$\tuple{\infty,\infty}$ in~$\EI{\downarrow}^{\infty}$ only if $w$ has rank~$\tuple{\infty,\infty}$ in~$\EI$ (again, see the definition of \emph{counterfactual shifting} in Section \ref{Entailment} and consider that no valuation with a finite rank satisfies $\gamma$ in $\EI$); hence  we have that the minimal valuations in~$\EI$ satisfying~$\alpha\land \gamma$ are in rank~$\tuple{\infty,\infty}$. 
 In both cases, we have $\EI\sat\alpha\twiddle_{\gamma}\beta$.
\end{proof}

\restatablePropUniqueMinimalModel*

\begin{proof}
We divide the proof into two parts. First, we prove that~$\EI_{\KB}$ is a minimal epistemic model, then that it is also the~\emph{only} minimal epistemic model.

Regarding minimality, we proceed by contradiction. We know, by Proposition~\ref{Prop:ModelOfE}, that~$\EI_{\KB}$ is an epistemic model of~$\KB$. Assume it is \emph{not} minimal, that is, assume there is an epistemic model~$\EI'$ of~$\KB$ \st, for every~$u\in\U$, $\EI'(u)\leq\EI_{\KB}(u)$, and there is a~$w\in\U$ \st~$\EI'(w)<\EI_{\KB}(w)$. Regarding the ranking of~$w$, we have two possibilities:

\begin{itemize}
\item[Case~1.] $\EI_{\KB}(w)=\tuple{\f,i}$, for some $i$, and~$\EI'(w)=\tuple{\f,j}$, for some~$j<i$. Let~$\KB^{\f}_{\EI'}=\{\alpha\twiddle_{\gamma}\beta\in\KB\mid\EI'\not\sat\lnot\gamma\}$. By Corollary~\ref{Corollary:SituationAsPremise}, $\EI'\sat\alpha\land \gamma\twiddle\beta$, for every~$\alpha\twiddle_{\gamma}\beta\in\KB^{\f}_{\EI'}$. Consider the ranked interpretation~$\RM'$ defined as:
\[
\RM'(u)=\left\{\begin{array}{ll}
        i, & \text{if }\EI'(u)=\tuple{\f,i},\text{ for some }i; \\
        \infty, & \text{otherwise.}
    \end{array}
\right.
\]

$\RM'$ above is clearly a ranked model of every~$\alpha\land\gamma\twiddle\beta$ \st~$\alpha\twiddle_{\gamma}\beta\in\KB^{\f}_{\EI'}$. Since~$\RM'$ has only one infinite rank, $\infty$, $\RM'$~is also a ranked model of every~$\alpha\land \gamma\twiddle\beta$ \st~$\alpha\twiddle_{\gamma}\beta\in\KB\setminus\KB^{\f}_{\EI'}$, since the minimal valuations satisfying their premises are in~$\states{\tuple{\infty,\infty}}$, and consequently they are trivially satisfied. Hence, $\RM'$~is a ranked model of~$\KB^{\land}$.

By Definition~\ref{Def:ConstructionMinimalModel}, $\EI_{\KB}$~has been built using the minimal ranked model~$\RM$ of~$\KB^{\land}$. But now we have also a ranked model~$\RM'$ of~$\KB^{\land}$ s.t. $\RM'$~is a ranked model of~$\KB^{\land}$ and, moreover, $\RM'$ is preferred to~$\RM$: by Definition~\ref{Def:ConstructionMinimalModel}, for every $u\in\U$, 
\[
\RM(u)=\left\{\begin{array}{ll}
        i, & \text{if }\EI_{\KB}(u)=\tuple{\f,i},\text{ for some }i; \\
        \infty, & \text{otherwise.}
    \end{array}
\right.
\]

Since we have assumed that $\EI'(u)\leq\EI_{\KB}(u)$ for every $u$, and there is a~$w$ \st~$\EI_{\KB}(w)=\tuple{\f,i}$, for some $i$, and~$\EI'(w)=\tuple{\f,j}$, for some~$j<i$, we have that $\RM'(u)\leq\RM(u)$ for every $u$, and there is a $w$ \st~$\RM(w)=i$, for some $i\in\mathbb{N}$, and~$\RM'(w)=j$, for some~$j<i$. That is, for every~$u\in\U$ $\RM'(u)\leq\RM(u)$, and $\RM'(w)<\RM(w)$. Hence $\RM$ is not the minimal model of $\KB^{\land}$, and this leads to a contradiction.

\item[Case~2.] $\EI_{\KB}$ and~$\EI'$ are identical \wrt~the finite ranks, and~$\EI_{\KB}(w)=\tuple{\infty,i}$, for some~$i$. We have two subcases: $\EI'(w)=\tuple{\infty,j}$, for some~$j<i$, or~$\EI'(w)=\tuple{\f,j}$, for some~$j$. The latter subcase leads to a contradiction: it can be proved analogously to Case~1. It remains to prove the first subcase.
    
The proof is still close to the one for Case~1 above; we simply have to refer to the counterfactual shiftings of~$\EI_{\KB}$ and~$\EI'$, $\EI_{\KB\downarrow}^\infty$ and $\EI'^\infty_\downarrow$ (see page~\pageref{countshift}). Since~$\EI_{\KB}$ and~$\EI'$ are epistemic models of~$\KB^\land$, $\EI_{\KB\downarrow}^\infty$ and~$\EI'^\infty_\downarrow$ are epistemic models of~$\KB^{\land}_{\infty}$, and~$\EI'^\infty_\downarrow$ is preferred to~$\EI_{\KB\downarrow}^\infty$. From~$\EI_{\KB\downarrow}^\infty$ and~$\EI'^\infty_\downarrow$, we can extract two ranked interpretations, $\RM_{\KB}^\infty$ and~$\RM'^\infty$ (see Definition~\ref{Def:ExtractedRanked}), that are both epistemic models of~$\KB^{\land}_{\infty}$. In the construction of~$\EI_{\KB}$, following Definition~\ref{Def:ConstructionMinimalModel}, we have used for the infinite ranks the ranked interpretation~$\RM_{\KB}^\infty$, which, also by Definition~\ref{Def:ConstructionMinimalModel}, must be the minimal ranked model of~$\KB^{\land}_{\infty}$. But in the present case, $\RM_{\KB}^\infty$ cannot be the minimal ranked model of~$\KB^{\land}_{\infty}$, since $\RM'^\infty$ is a ranked model of~$\KB^{\land}_{\infty}$ that is preferred to~$\RM_{\KB}^\infty$. This leads to a contradiction.
\end{itemize}

To conclude this part, in all the possible cases, if~$\EI_{\KB}$ is not a minimal epistemic model of~$\KB$, then we end up with a contradiction. Hence~$\EI_{\KB}$ must be a minimal epistemic model of~$\KB$.

The final step consists in proving that~$\EI_{\KB}$ is the \emph{only} minimal epistemic model of~$\KB$. The procedure is again by contradiction, assuming that~$\EI_{\KB}$ is not the only minimal epistemic model of~$\KB$. Hence, let~$\EI'$ be another minimal epistemic model of~$\KB$. The structure of the proof actually mirrors the one for the previous part, about the minimality of~$\EI_{\KB}$. Again, we can distinguish two main cases.
\begin{itemize}
\item[Case~1.] $\EI_{\KB}$ and~$\EI'$ differ \wrt~the ranking of some valuations among the ones ranked as finite. From~$\EI_{\KB}$ and~$\EI'$, we can extract, respectively, the ranked models~$\RM$ and~$\RM'$, which are both ranked models of~$\KB^{\land}$. But, by Definition~\ref{Def:MinimalEpistemicModel}, $\RM$ is the only minimal ranked model of~$\KB^{\land}$, that is, $\RM\prec\RM'$, which implies that~$\EI'$ cannot be a minimal epistemic model of~$\KB$.
\item[Case~2.] $\EI_{\KB}$ and~$\EI'$ do not differ \wrt~the ranking of the  valuations that are ranked as finite in both of them but differ \wrt~the ranking of some valuation, $w$, that is  ranked as infinite in one of the two. W.l.o.g., we assume that~$w$ is  ranked as infinite in~$\EI_{\KB}$. We  have two subcases: $\EI'(w)=\tuple{\infty,j}$, for some~$j$, or~$\EI'(w)=\tuple{\f,j}$, for some~$j$. The latter subcase leads to a contradiction: it can be proved analogously to Case~1 using the extracted ranked models. It remains to show the first subcase.
    
The proof is still analogous to Case~2 above. We refer to the counterfactual shiftings of~$\EI_{\KB}$ and~$\EI'$, $\EI_{\KB\downarrow}^\infty$ and~$\EI'^\infty_\downarrow$. Since~$\EI_{\KB}$ and~$\EI'$ are epistemic models of~$\KB^\land$ and they are identical \wrt~the finite ranks, $\EI_{\KB\downarrow}^\infty$ and~$\EI'^\infty_\downarrow$ are epistemic models of~$\KB^{\land}_{\infty}$. From~$\EI_{\KB\downarrow}^\infty$ and~$\EI'^\infty_\downarrow$, we can extract two ranked interpretations, $\RM_{\KB}^\infty$ and~$\RM'^\infty$ (see Definition~\ref{Def:ExtractedRanked}), that are both ranked models of~$\KB^{\land}_{\infty}$. In the construction of~$\EI_{\KB}$, following Definition~\ref{Def:MinimalEpistemicModel}, we have used for the infinite ranks the ranked interpretation~$\RM_{\KB}^\infty$, which, also by Definition~\ref{Def:MinimalEpistemicModel}, must be the minimal ranked model of~$\KB^{\land}_{\infty}$. If $\RM_{\KB}^\infty$ is the minimal ranked model of~$\KB^{\land}_{\infty}$, then $\RM_{\KB}^\infty$ is preferred to~$\RM'^\infty$, and, by construction, $\EI_{\KB}$ must be preferred to~$\EI'$. This leads to a contradiction.
\end{itemize}

To conclude, if we assume that there is another minimal epistemic model of~$\KB$ beyond~$\EI_{\KB}$, we end up with a contradiction. Hence, $\EI_{\KB}$ must be the only minimal epistemic model of~$\KB$.
\end{proof}

\restatableTheoremCompleteProcedure*

\begin{proof}
We already know that algorithms $\mathtt{Exceptional}$, $\mathtt{ComputeRanking}$, $\mathtt{ Rank}$ and $\mathtt{RationalClosure}$ are complete and correct \wrt\ the corresponding semantic notions.

As a first step, we need to prove that algorithm $\mathtt{Partition}$ returns the correct result, that is, the sets $\KB_{\infty}$ and $\KB^{\land}_{\infty\downarrow}$ correspond to the same sets introduced in Definition~\ref{Def:ConstructionMinimalModel}.

The correspondence of $\KB_{\infty}$ to the semantic notion introduced in Definition~\ref{Def:ConstructionMinimalModel} is guaranteed by the correctness of algorithm $\mathtt{ComputeRanking}$ \wrt~the semantic definition of ranks \wrt~the rational closure.

To prove the correspondence of $\KB^{\land}_{\infty\downarrow}$ to the semantic notion in Definition~\ref{Def:ConstructionMinimalModel}, we need to prove also that the defeasible conditionals $\mu\twiddle\bot$ and $\fm{\U^{\f}_{\RM}}\twiddle\bot$ are equivalent, which is an immediate consequence of Lemma~\ref{lemma_knowledge} and the LLE postulate.

Now we can check the correctness of algorithm $\mathtt{MinimalClosure}$. We consider the possible cases as presented in the algorithm.

\begin{itemize}
\item[Case 1.] $\overline{\KB^{\land}}\models\bot$. 

By Corollary~\ref{Corollary:Consistency}, $\overline{\KB^{\land}}\models\bot$ iff $\KB$ is inconsistent, and in such a case $\KB\miniment\alpha\twiddle_{\gamma}\beta$ for every $\alpha,\gamma,\beta$, and the algorithm behaves correctly.

\item[Case 2.]$\overline{\KB^{\land}}\not\models\bot$ and $\mathtt{Rank}(\KB^{\land},\gamma)<\infty$.

We have to prove that in this case, $\alpha\land\gamma\twiddle\beta$ is in the RC of $\KB^{\land}$ iff $\alpha\twiddle_{\gamma}\beta$ is in the minimal closure of $\KB$. 

Assume $\alpha\land\gamma\twiddle\beta$ is in the RC of $\KB^{\land}$, and let $\RM$ be the minimal ranked model of $\KB^\land$. That means that  $\min\states{\alpha\land\gamma}_{\RM}\subseteq\states{\beta}$.  Also, since $\mathtt{Rank}(\KB^{\land},\gamma)<\infty$, we have that $\states{\gamma}\cap\U^{\f}_{\RM}\neq\emptyset$. By construction of the minimal epistemic model of $\KB$, $\E_{\KB}$, $\U^{\f}_{\E_{\KB}}=\U^{\f}_{\RM}$, and the rank of each valuation is the same. Consequently, we have that $\states{\gamma}^{\f}_{\EI_{\KB}}\neq\emptyset$. According to Definition~\ref{Def:SatisfactionEpistemicInterpretation}, we have to check whether $\states{\gamma}^{\f}_{\EI_{\KB}}\subseteq\states{\neg\alpha}$ or $\states{\alpha\land\gamma}_{\EI_{\KB}}\subseteq\states{\beta}$. From $\RM\sat\alpha\land\gamma\twiddle\beta$, we single out two possible cases: 

\begin{itemize}
    \item $\mathtt{Rank}(\KB^{\land},\alpha)=\infty$. This implies that $\states{\gamma}^{\f}_{\EI_{\KB}}\subseteq\states{\neg\alpha}$.
    \item Otherwise, in $\RM$ we have $\min\states{\alpha\land\gamma}^{\f}\subseteq\states{\beta}$. Since $\EI_{\KB}$ preserves in~$\U^{\f}_{\EI_{\KB}}$ the same ranking as in $\U^{\f}_{\RM}$, we have $\states{\alpha\land\gamma}^{\f}_{\EI_{\KB}}\subseteq\states{\beta}$.
\end{itemize}

We can conclude that $\EI_{\KB}\sat\alpha\twiddle_{\gamma}\beta$.

Now we check the opposite direction: we assume $\EI_{\KB}\sat\alpha\twiddle_{\gamma}\beta$. Since $\mathtt{Rank}(\KB^{\land},\gamma)<\infty$, by Definition~\ref{Def:ConstructionMinimalModel} we have that $\states{\gamma}^{\f}_{\EI_{\KB}}\neq\emptyset$. The latter, together with $\EI_{\KB}\sat\alpha\twiddle_{\gamma}\beta$, implies  $\states{\alpha\land\gamma}^{\f}_{\EI_{\KB}}\subseteq\states{\beta}$. By Definition~\ref{Def:ConstructionMinimalModel}, this condition implies that $\states{\alpha\land\gamma}^{\f}_{\RM}\subseteq\states{\beta}$, which in turn implies $\RM\sat\alpha\land\gamma\twiddle\beta$.

\item[Case 3.]$\overline{\KB^{\land}}\not\models\bot$ and $\mathtt{Rank}(\KB^{\land},\gamma)=\infty$.

We have to prove that in this case $\alpha\land\gamma\twiddle\beta$ is in the RC of $\KB^{\land}_{\infty\downarrow}$ iff $\alpha\twiddle_{\gamma}\beta$ is in the minimal closure of $\KB$. 

Since $\mathtt{Rank}(\KB^{\land},\gamma)=\infty$, we have that $\states{\gamma}\cap\U^{\f}_{\RM}=\emptyset$. By construction of the minimal epistemic model of $\KB$, $\EI_{\KB}$, $\U^{\f}_{\EI_{\KB}}=\U^{\f}_{\RM}$. Consequently, we have that $\states{\gamma}^{\f}_{\EI_{\KB}}=\emptyset$ and $\states{\gamma}_{\EI_{\KB}}$ all have rank $\tuple{\infty,j}$, for some $j$.

Assume $\alpha\land\gamma\twiddle\beta$ is in the RC of $\KB^{\land}_{\infty\downarrow}$, and let $\RM'$ be the minimal ranked model of $\KB^\land_{\infty\downarrow}$. 
According to Definition~\ref{Def:SatisfactionEpistemicInterpretation}, we have to check whether $\states{\alpha\land\gamma}_{\EI_{\KB}}\subseteq\states{\beta}$. Assume this is not the case, that is, $\EI_{\KB}\not\sat\alpha\twiddle_{\gamma}\beta$. Since $\states{\gamma}^{\f}_{\EI_{\KB}}=\emptyset$, all the valuations in $\states{\gamma}_{\EI_{\KB}}$ are  ranked as infinite, and $\EI_{\KB}\not\sat\alpha\twiddle_{\gamma}\beta$ implies that there is a valuation~$w$ in $\states{\alpha\land\gamma}_{\EI_{\KB}}$ \st\ $w\not\sat\beta$. Let $w\in\states{\tuple{\infty,i}}$, for some $i<\infty$, and $w\preceq v$, for every $v\in\states{\alpha\land\gamma}$. By Definition~\ref{Def:MinimalEpistemicModel}, in~$\RM'$ we have $w\in\states{i}$, for some $i<\infty$, and $w\preceq v$, for every $v\in\states{\alpha\land\gamma}$. Hence, we would have $\RM'\not\sat\alpha\land\gamma\twiddle\beta$, which is against our hypothesis that $\alpha\land\gamma\twiddle\beta$ is in the~RC of~$\KB^{\land}_{\infty\downarrow}$.

Now we assume $\EI_{\KB}\sat\alpha\twiddle_{\gamma}\beta$. Again, since $\states{\gamma}^{\f}_{\EI_{\KB}}=\emptyset$, all the valuations in $\states{\gamma}_{\EI_{\KB}}$ are  ranked as infinite. The latter, together with Definition~\ref{Def:MinimalEpistemicModel}, implies that $\states{\gamma}_{\EI_{\KB}}=\min\states{\alpha\land\gamma}_{\RM'}$, and consequently  $\states{\gamma}_{\EI_{\KB}}\subseteq\states{\beta}$ implies  $\min\states{\alpha\land\gamma}_{\RM'}\subseteq\states{\beta}$. We can conclude $\RM'\sat\alpha\land\gamma\twiddle\beta$, that is, $\alpha\land\gamma\twiddle\beta$ is in the~RC of~$\KB^{\land}_{\infty\downarrow}$.





\end{itemize}

We have proved that in all possible cases $\mathtt{MinimalClosure}(\KB,\alpha\twiddle_{\gamma}\beta)$ returns $\mathtt{true}$ iff $\KB\miniment\alpha\twiddle_{\gamma}\beta$.
\end{proof}

\end{document}


\maketitle

In this appendix we present the proofs of the propositions in the paper. In Section~\ref{sect_procedure} we also present the procedure to decide minimal entailment that is mentioned at the end of Section 4 in the paper.


\section{Proofs}

{\noindent\bf Corollary 1.} \emph{Let $\EI$ be an epistemic interpretation. Then $\RI^{\EI}\sat\alpha\twiddle\beta$ iff $\EI\sat\alpha\twiddle_{\top}\beta$. }

\begin{proof}
By definition, $\EI\sat\alpha\twiddle_{\top}\beta$ iff  $\states{\alpha}_{\EI}\subseteq\states{\beta}$ or $\emptyset\neq\states{\top}^f_{\EI}\subseteq\states{\lnot\alpha}$. If $\states{\alpha}^f_{\EI}\neq\emptyset$ the result follows since $\U^f_{\RI^{\EI}}=\U^f_{\EI}$. If $\states{\alpha}^f_{\EI}=\emptyset$ then $\EI\sat\alpha\twiddle_{\top}\bot$ and $\RI^{\EI}\sat\alpha\twiddle\bot$. 
\end{proof}

{\noindent\bf Theorem 2.} \emph{Every epistemic interpretation generates a BCC, but the converse does not hold.}  

\begin{proof}
Consider any epistemic interpretation $\EI$ and pick any $\gamma\in\Lang$. We consider three disjoint and covering cases. 

Case 1: If $\U^f_{\EI}\cap\states{\gamma}\neq\emptyset$, define $\RI$ from $\EI$ as follows: 
(i) for all $u\in\U^f_{\EI}\cap\states{\gamma}$, $\RI(u)\defined i$ where $\EI(u)=\tuple{f,i}$; (ii) for all $u\in\U\setminus\U^f_{\EI}\cap\states{\gamma}$,
$\RI(u)\defined\infty$. If follows from Definition 4 and the definition (in Section 2) of satisfaction for $\twiddle$ in ranked interpretations that 
$\EI\sat\alpha\twiddle_{\gamma}\beta$ iff $\RI\sat\alpha\twiddle\beta$. From Theorem 1 it follows that the $\twiddle$ generated by $\RI$ satisfies the original KLM properties, and therefore, so does $\twiddle_{\gamma}$ (where $\twiddle_{\gamma}$ is the defeasible conditional obtained from $\EI$ by fixing $\gamma$).  

Case 2: If $\U^f_{\EI}\cap\states{\gamma}=\emptyset$ but $\U^{\infty}_{\EI}\cap\states{\gamma}\neq\emptyset$, define $\RI$ from $\EI$ as follows: (i) for all $u\in\U^{\infty}_{\EI}\cap\states{\gamma}$, $\RI(u)\defined i$ where $\EI(u)=\tuple{\infty,i}$; (ii) for all $u\in\U\setminus(U^{\infty}_{\EI}\cap\states{\gamma})$, $\RI(u)\defined\infty$. 
If follows from Definition 4 and the definition (in Section 2) of satisfaction for $\twiddle$ in ranked interpretations that 
$\EI\sat\alpha\twiddle_{\gamma}\beta$ iff $\RI\sat\alpha\twiddle\beta$. From Theorem 1 it follows that the $\twiddle$ generated by $\RI$ satisfies the original KLM properties, and therefore, so does $\twiddle_{\gamma}$ (where $\twiddle_{\gamma}$ is the defeasible conditional obtained from $\EI$ by fixing $\gamma$).  

Case 3: If $\states{\gamma}\subseteq\U\setminus(\U^f_{\EI}\cup\U^{\infty}_{\EI})$ then $\RI(u)\defined\infty$ for all $u\in\U$. Again, it follows from Definition 4 and the definition (in Section 2) of satisfaction for $\twiddle$ in ranked interpretations that 
$\EI\sat\alpha\twiddle_{\gamma}\beta$ iff $\RI\sat\alpha\twiddle\beta$. From Theorem 1 it follows that the $\twiddle$ generated by $\RI$ satisfies the original KLM properties, and therefore, so does $\twiddle_{\gamma}$ (where $\twiddle_{\gamma}$ is the defeasible conditional obtained from $\EI$ by fixing $\gamma$).    

From this it follows that for all $\gamma\in\Lang$, $\twiddle_{\gamma}$ viewed as a defeasible conditional, satisfies the original KLM properties. It then follows immediately that the contextual conditional $\twiddle$ obtained from $\EI$ satisfies the contextual KLM properties. 

To show that the converse does not hold, consider the language generated from $\{\ce{p},\ce{q}\}$. Note firstly that there is a ranked interpretation $\RI$ such that $\RI\sat\alpha\twiddle\beta$ iff 
$\ce{p}\land\ce{q}\land\alpha\entails\beta$. From Theorem 1 it therefore follows that $\twiddle$, defined in this way, is a rational conditional, and therefore satisfies the contextual KLM properties. 
Similarly, there is a ranked interpretation $\RI'$ such that $\RI'\sat\alpha\twiddle\beta$ iff $\ce{p}\land\ce{q}\land\alpha\entails\beta$. From Theorem 1 it therefore follows that $\twiddle$, defined in this way, is a rational conditional, and therefore satisfies the contextual KLM properties. Now, define the contextual conditional $\twiddle$ by letting $\alpha\twiddle_{\ce{p}}\beta$ iff $\ce{p}\land\ce{q}\land\alpha\entails\beta$, and $\alpha\twiddle_{\gamma}\beta$ iff $\alpha\entails\beta$ for every $\gamma$ other than \ce{p}.  It then follows immediately that $\twiddle$ is a BCC. However, it is easy to see that it cannot be generated by an epistemic interpretation. To see why, observe that $\ce{p}\twiddle_{\ce{p}}\ce{q}$ but that $\ce{p}\ntwiddle_{\ce{p\vee p}}\ce{q}$.

\end{proof}

{\noindent\bf Theorem 3.} \emph{Every epistemic interpretation generates an FCC. Every FCC can be generated by an epistemic interpretation.}

\begin{proof}
Let $\EI$ be an epistemic interpretation. Suppose $\U^f_{\EI}\cap\states{\gamma}\neq\emptyset$. Then if $\alpha\twiddle_{\gamma}\beta$ it follows by Definition 4 that  
$\alpha\land\gamma\twiddle_{\top}\beta$. On the other hand, if  $\U^f_{\EI}\cap\states{\gamma}=\emptyset$ then $\alpha\land\gamma\twiddle_{\top}\beta$, which means that the contextual conditional generated by $\EI$ satisfies Inc. 

Suppose $\EI\nsat\top\twiddle_{\top}\lnot\gamma$. This means that $\U^f_{\EI}\cap\states{\gamma}\neq\emptyset$. Then if $\alpha\land\gamma\twiddle_{\bot}\beta$ it follows by Definition 4 that  
$\alpha\twiddle_{\gamma}\beta$, which means that the contextual conditional generated by $\EI$ satisfies Vac.

That the contextual conditional generated by $\EI$ satisfies Ext follows immediately from Definition 4.

For SupExp we consider two cases. For case 1, if $\U^f_{\EI}\cap\states{\gamma\land\delta}\neq\emptyset$ the result follows easily. For case 2, 
suppose $\U^f_{\EI}\cap\states{\gamma\land\delta}=\emptyset$. If $\U^f_{\EI}\cap\states{\delta}=\emptyset$ the result follows easily. Otherwise the result follows from the fact that  
$\U^f_{\EI}\cap\states{\alpha\land\gamma\land\delta}=\emptyset$. 

For SubExp, suppose that $\EI\sat\delta\twiddle_{\top}\bot$. This means $\EI\sat\alpha\land\gamma \twiddle_{\delta}\beta$ implies that 
$\U^{\infty}_{\EI}\cap\states{\alpha\land\gamma\land\delta}\subseteq\beta$, from which it follows that $\EI\sat\alpha \twiddle_{\gamma\wedge\delta}\beta$. 

For the converse, consider any FCC $\twiddle$. We construct an epistemic interpretation $\EI$ as follows. First, consider $\twiddle_{\top}$. Since it satisfies the contextual KLM properties, there is a ranked interpretation $\RI$ such that $\RI\sat\alpha\twiddle\beta$ iff $\alpha\twiddle_{\top}\beta$. We set $\U^f_{\EI}\defined\U^{\RI}$ and for all $u\in\U^f_{\EI}$, we let 
$\EI(u)\defined\tuple{f,\RI(u)}$. Next, let $\U'\defined\U\setminus\U^f_{\EI}$. Let $k^f$ be a formula such that $\states{k^f}=\U^f_{\EI}$. Similarly, let 
$k^{\infty}$ be a formula such that $\states{k^{\infty}}=\U'$.
 Now, consider $\twiddle_{k^{\infty}}$. Since it satisfies the contextual KLM properties, there is a ranked interpretation $\RI'$ such that 
 $\RI'\sat\alpha\twiddle\beta$ iff $\alpha\twiddle_{k^{\infty}}\beta$. We let $\U^{\infty}_{\EI}\defined\{ u\in\U' \mid \RI'(u)\neq\infty\}$ and for all $u\in\U'$, we let $\EI(u)\defined\tuple{\infty,\RI'(u)}$. 
 Observe that for some $u\in\U'$ it may be the case  $\EI(u)=\tuple{\infty,\infty}$, which means that for such a $u$, $u\notin\U^{\infty}_{\EI}$. 
 It is easily verified that $\EI$ is indeed an epistemic interpretation. Next we show that $\alpha\twiddle_{\gamma}\beta$ iff $\EI\sat\alpha\twiddle_{\gamma}\beta$. We do so by considering two cases. 
 Case 1: $\U^f_{\EI}\cap\states{\gamma}\neq\emptyset$. Note firstly that it follows easily from the construction of $\EI$ that $\alpha\twiddle_{\top}\beta$ iff  $\EI\sat\alpha\twiddle_{\top}\beta$. Suppose  $\alpha\twiddle_{\gamma}\beta$. By Inc,  
 $\alpha\land\gamma\twiddle_{\top}\beta$ and therefore $\EI\sat\alpha\land\gamma\twiddle_{\top}\beta$, and  $\EI\sat\alpha\twiddle_{\gamma}\beta$, by definition. Conversely,  
suppose $\EI\sat\alpha\twiddle_{\gamma}\beta$. Then by definition, $\EI\sat\alpha\land\gamma\twiddle_{\top}\beta$, and therefore  $\alpha\land\gamma\twiddle_{\top}\beta$. Since 
$\top\ntwiddle_{\top}\lnot\gamma$, it then follows from Vac that $\alpha\twiddle_{\gamma}\beta$.  Case 2:  $\U^f_{\EI}\cap\states{\gamma}\neq\emptyset$. By the construction of $\EI$ it follows that $\alpha\twiddle_{k^{\infty}}\beta$ iff  $\EI\sat\alpha\twiddle_{k^{\infty}}\beta$. Suppose  $\alpha\twiddle_{\gamma}\beta$. Note that $\gamma\equiv k^{\infty}$. By Ext, 
$\alpha\twiddle_{\gamma\land k^{\infty}}\beta$ and so, by SupExp, $\alpha\land\gamma\twiddle_{k^{\infty}}\beta$. It then follows that $\EI\sat\alpha\land\gamma\twiddle_{k^{\infty}}\beta$ and, by Definition 4, that $\EI\sat\alpha\twiddle_{\gamma}\beta$. Conversely, suppose that $\EI\sat\alpha\twiddle_{\gamma}\beta$. Then $\EI\sat\alpha\land\gamma\twiddle_{k^{\infty}}\beta$ by Definition 4, and therefore, using Ext,  
that $\alpha\land\gamma\twiddle_{\gamma\land k^{\infty}}\beta$. Note that $\EI\sat k^f\twiddle_{\top}\bot$ and therefore $k^f\twiddle_{\top}\bot$. By SubExp it then follows that 
$\alpha\twiddle_{\gamma\land k^{\infty}}\beta$, and by Ext that $\alpha\twiddle_{\gamma}\beta$.
\end{proof}

{\noindent\bf Corollary 2.} \emph{
Every FCC satisfies Succ, but there are FCCs for which Cons does not hold.}

\begin{proof}
To prove that Succ holds it suffices, by Theorem 2, to show that $\EI\sat\alpha\twiddle_{\gamma}\gamma$ for all epistemic interpretations $\EI$, and all $\alpha,\gamma$. To see that this holds, observe that $\states{\alpha\land\gamma}_{\EI}\subseteq\states{\gamma}$. 

To prove that Cons does not hold it  suffices, by Theorem 2, to show that there is an epistemic interpretation $\EI$ such that $\EI\sat\top_{\gamma}\bot$ but $\gamma\neq\bot$.  To construct such an $\EI$, let $\U^f_{\EI} = \U^{\infty}_{\EI}=\emptyset$ (and so $\EI(u)=\tuple{\infty,\infty}$ for all $u\in\U)$.
\end{proof}

{\noindent\bf Proposition 1.} \emph{
Every FCC satisfies Incons and Cond.}

\begin{proof}
To prove that Incons holds it suffices, by Theorem 2, to show that $\EI\sat\alpha\twiddle_{\bot}\beta$ for all epistemic interpretations $\EI$, and all $\alpha,\beta$. To see that this holds, observe that $\states{\alpha\land\bot}_{\EI}=\emptyset$. 

To prove that Cond holds it suffices, by Theorem 2, to show that if $\EI\nsat\gamma\twiddle_{\top}\bot$ then $\EI\sat\alpha\land\gamma\twiddle_{\top}\beta$ iff $\EI\sat\alpha\twiddle_{\gamma}\beta$ for all epistemic interpretations $\EI$, and all $\alpha,\beta,\gamma$. So, suppose that 
$\EI\nsat\gamma\twiddle_{\top}\bot$. By Definition 4 this means that $\U^f_{\EI}\cap\states{\gamma}\neq\emptyset$. From this it follows that 
$\states{\alpha\land\gamma}_{\EI}=\states{\alpha\land\gamma\land\top}_{\EI}$, that $\emptyset\neq\states{\gamma}^f_{\EI}$ and  $\emptyset\neq\states{\top}^f_{\EI}$, and  that
$\states{\gamma}^f_{\EI}\subseteq\states{\lnot\alpha}$ iff $\states{\top}^f_{\EI}\subseteq\states{\lnot\alpha}$, from which the results follows by Definition 4.
\end{proof}

{\noindent\bf Corollary 3.} \emph{For any epistemic interpretation $\EI$, if $\U^f_{\EI}\cap\states{\gamma}\neq\emptyset$ then $\EI\sat\alpha\twiddle_{\gamma}\beta$  iff $\EI\sat\alpha\land\gamma\twiddle_{\top}\beta$.}

\begin{proof}

Since it is just a semantic reformulation of the property (Cond), it follows directly from the proof  in Proposition 1 that (Cond) holds.
\end{proof}

{\noindent\bf Proposition 2.} \emph{Let $\KB$ be a conditional base, and let $\EI_{\RM}$ be defined as in Definition 10. $\EI_{\RM}$ is an epistemic model of $\KB$.}

\begin{proof}
Let $\alpha\twiddle_{\gamma} \beta\in\KB$. Since $\EI_{\RM}$ is an epistemic model of $\KB^{\land}$ and we have Corollary 3,  if  $\states{\gamma}\cap \U^f_{\EI_{\RM}}\neq\emptyset$ then we conclude  $\EI_{\RM}\sat\alpha\twiddle_{\gamma} \beta$. Otherwise, we have  $\states{\gamma}\subseteq \states{\tuple{\infty,\infty}}$, that implies $\states{\alpha\land\gamma}\subseteq \states{\tuple{\infty,\infty}}$, that in turn  implies $\EI_{\RM}\sat\alpha\twiddle_{\gamma} \beta$.
\end{proof}

{\noindent\bf Proposition 3.} \emph{Let $\KB$ be a conditional base. $\KB$ has an epistemic model iff $\KB^{\land}$ has a ranked model.}

\begin{proof}
Proposition 2 proves that if $\KB^{\land}$ has a ranked model, then $\KB$ has an epistemic model. For the opposite direction, assume that $\KB$ has an epistemic model $\EI$. From $\EI$ we define an epistemic model $\EI_{rk}$ in the following way:

\[\EI_{rk}(u)=
\left\{\begin{array}{ll}
    \EI(u) & \text{if } \EI(u)=\tuple{f,i} \text{ for some }i;\\
    \tuple{\infty, \infty} & \text{otherwise.}
\end{array}\right.
\]

It is easy to check that $\EI_{rk}$ is still an epistemic model of $\KB$, and moreover, because of Corollary 3, it is an epistemic model of $\KB^{\land}$: For every $\alpha\twiddle_{\gamma}\beta\in\KB$, if $\EI_{rk}\not\sat\neg\gamma$, then $\EI_{rk}\sat\alpha\land\gamma\twiddle\beta$ by Corollary 3; if $\EI_{rk}\sat\neg\gamma$, then $\states{\alpha\land\gamma}\subseteq\states{\tuple{\infty,\infty}}$, and we can conclude $\EI_{rk}\sat\alpha\land\gamma\twiddle\beta$.

Let $\R$ be the ranked model corresponding to $\EI_{rk}$, that is, 

\[\R(u)=
\left\{\begin{array}{ll}
    i & \text{if } \EI_{rk}(u)=\tuple{f,i} \text{ for some }i;\\
    \infty & \text{otherwise.}
\end{array}\right.
\]

Since the preferences between all the valuations in $\U$ are the same in $\EI_{rk}$ and $R$, it is easy to see that if $\EI_{rk}$ is an epistemic model of $\KB^{\land}$,  $\R$ is a ranked model of $\KB^{\land}$.
\end{proof}

{\noindent\bf Corollary 4.} \emph{A CCKB $\KB$ is consistent iff $\overline{\KB^\land}\not\entails\bot$.}

\begin{proof}
It is a known fact that a set of classical conditionals $\C$ has a ranked model iff its materialisation $\overline{\C}$ is consistent w.r.t. propositional logic \cite[Lemma 5.21]{LehmannMagidor1992}. This fact, together with Proposition 3, proves the corollary.
\end{proof}

{\noindent\bf Lemma 1.} \emph{For any epistemic interpretation $\EI$, if $\EI\sat\neg\gamma$ then $\EI\sat\alpha\twiddle_{\gamma}\beta$  iff $\EI^{\infty}\sat\alpha\land\gamma\twiddle\beta$.}

\begin{proof}
Let $\EI\sat\lnot\gamma$, that is, there are not valuations in the finite ranks that satisfy $\gamma$; hence the satisfaction of the conditionals with context $\gamma$ must be checked referring to infinitely ranked valuations.  $\EI\sat\alpha\twiddle_{\gamma}\beta$ imposes that, among the infinitely ranked valuations in $\states{\gamma}$ there are minimal infinitely ranked valuations satisfying $\alpha\land\gamma$ and that all of them satisfy also $\beta$, or that the minimal valuations satisfying $\alpha\land\gamma$ have rank $\tuple{\infty,\infty}$. $\gamma$ has finite rank in $\EI^{\infty}$, or rank $\tuple{\infty,\infty}$. In the latter case, we have $\EI^{\infty}\sat\alpha\land \gamma\twiddle\beta$. In the former case, the rank of $\gamma$ in $\EI$ is $\tuple{\infty,i}$, with $i<\infty$, that is, the rank of $\gamma\land \alpha$ in $\EI^{\infty}$ is $\tuple{f,j}$, for some $j$, $i\leq j<\infty$, or $\tuple{\infty,\infty}$. In the latter case, again, it is straightforward to conclude $\EI^{\infty}\sat\alpha\land \gamma\twiddle\beta$. In the former case, $\EI\sat\alpha\twiddle_{\gamma}\beta$ and the construction of $\EI^{\infty}$ impose that the minimal valuations in $\states{\alpha\land\gamma}$ satisfy also $\beta$, that is, $\EI^{\infty}\sat\alpha\land \gamma\twiddle\beta$.

The proof is analogous in the opposite direction. If $\EI\sat\lnot\gamma$, then there are finitely ranked valuations in $\EI^\infty$ satisfying $\gamma$. Let $\EI^{\infty}\sat\alpha\land \gamma\twiddle\beta$. Either the minimal valuations in $\EI^{\infty}$ satisfying $\alpha\land \gamma$ are in rank $\tuple{f,i}$, for some $i<\infty$,  and satisfy $\beta$, or they are in $\tuple{\infty,\infty}$: in the former case, it means that the minimal valuations in $\EI^{\infty}$ satisfying $\alpha\land \gamma$ are in rank $\tuple{f,i}$, for some $i<\infty$,  and satisfy $\beta$, or they are in $\tuple{\infty,\infty}$. In both cases, since the minimal valuations in $\EI$ satisfying $\gamma$ are infinitely ranked, $\EI\sat\alpha\twiddle_{\gamma}\beta$.
\end{proof}

{\noindent\bf Proposition 4.} \emph{Let $\KB$ be a consistent CCKB, and let $\EI_{\KB}$ be an epistemic interpretation built as in Definition 12. $\EI_{\KB}$ is an epistemic model of $\KB$.}

\begin{proof}
Let $\KB_{\infty}$ be defined as in Definition 12. We  distinguish two possible cases.

\begin{itemize}
    \item $\alpha\twiddle_{\gamma}\beta\in \KB\setminus \KB_{\infty}$, that is, $\EI_{\KB}(\gamma)=\tuple{f,i}$ for some $i$. Due to the construction of $\EI_{\KB}$ (Definition 12), $\EI_{\KB}$ is an epistemic model of $\KB^{\land}$, that is, it is an epistemic model of $\alpha\land\gamma\twiddle\beta$; due to Corollary 3, $\EI_{\KB}\sat \alpha\twiddle_{\gamma}\beta$.
    \item $\alpha\twiddle_{\gamma}\beta\in \KB_{\infty}$, that is, $\EI_{\KB}(\gamma)=\tuple{\infty,i}$ for some $i$. Due to the construction of $\EI_{\KB}$ (Definition 12), $\EI_{\KB}$ is an epistemic model of $\KB^{\land}$, that is, is an epistemic model of $\alpha\land\gamma\twiddle\beta$. Let $\EI_{\KB\downarrow}^{\infty}$ be the counterfactual shifting of $\EI_{\KB}$. Due to Lemma 1 we know that, since  $\EI_{\KB\downarrow}^{\infty}\sat\alpha\land\gamma\twiddle\beta$, $\EI_{\KB\downarrow}^{\infty}\sat\alpha\twiddle_{\gamma}\beta$ holds. Since $\states{\alpha\land\gamma}_{\EI_{\KB}}=\states{\alpha\land\gamma}_{\EI_{\KB}}^{\infty}=\states{\alpha\land\gamma}_{\EI_{\KB}^\infty}$, for every $u\in\U$, $u\in \states{\alpha\land\gamma}_{\EI_{\KB}^\infty}$ iff $u\in \states{\alpha\land\gamma}_{\EI_{\KB}}$, that is, $\EI_{\KB}\sat \alpha\twiddle_{\gamma}\beta$.
\end{itemize}

We can conclude that for very $\alpha\twiddle_{\gamma}\beta\in\KB$, $\EI_{\KB}\sat \alpha\twiddle_{\gamma}\beta$

\end{proof}

{\noindent\bf Proposition 5.} \emph{
Let $\KB$ be a consistent CCKB, and let $\EI_{\KB}$ be an epistemic interpretation built as in Definition 12. $\EI_{\KB}$ is the only minimal epistemic model of $\KB$.}

\begin{proof}
We divide the proof in two parts: first we prove that $\EI_{\KB}$ is a minimal epistemic model, then that it is the \emph{only} minimal epistemic model.

Regarding minimality, we proceed by contradiction. We know by Proposition 4 that $\EI_{\KB}$ is an epistemic model of $\KB$. We assume it is not minimal, that is, there is an epistemic model $\EI'$ of $\KB$ s.t., for every $u\in\U$, $\EI'(u)\leq\EI_{\KB}(u)$, and there is a $w\in\U$ s.t. $\EI'(w)<\EI_{\KB}(w)$.
Regarding the ranking of $w$, we have two possibilities:

\begin{itemize}
    \item[Case 1.] $\EI_{\KB}(w)=\tuple{f,i}$ for some $i$, and $\EI'(w)=\tuple{f,j}$ for some $j<i$. Let $\KB^f_{\EI'}=\{\alpha\twiddle_{\gamma}\beta\in\KB\mid \EI'\not\sat\neg\gamma\}$. By Corollary 3, $\EI'\sat\alpha\land \gamma\twiddle\beta$ for every $\alpha\twiddle_{\gamma}\beta\in \KB^f_{\EI'}$. Consider the ranked interpretation $\R'$ defined as:
    \[\R'(u)=\left\{\begin{array}{ll}
        i & \text{if }\EI'(u)=\tuple{f,i}\text{ for some }i; \\
        \infty & \text{otherwise.}
    \end{array}
    \right.\]
    
    $\R'$ is clearly a ranked model for every $\alpha\land \gamma\twiddle\beta$ s.t. $\alpha\twiddle_{\gamma}\beta\in \KB^f_{\EI'}$; since  $\R'$ has only an infinite rank, $\infty$,  $\R'$ is also a ranked model for every $\alpha\land \gamma\twiddle\beta$ s.t. $\alpha\twiddle_{\gamma}\beta\in \KB\setminus\KB^f_{\EI'}$, since the minimal valuations satisfying their premises are in $\states{\tuple{\infty,\infty}}$ and consequently they are trivially satisfied. Consequently, $\R'$ is  a ranked model of $\KB^{\land}$. 
    
    By Definition 12, $\EI_{\KB}$ has been built using the minimal ranked model $\R$ of $\KB^{\land}$. However, now we end up with a ranked model $\R'$ of $\KB^{\land}$ that is  preferred to $\R$, since for every $u\in\U$, $\R'(u)\leq\R_{\KB}(u)$, and $\R'(w)<\R_{\KB}(w)$. This is a contradiction.
    
    \item[Case 2.] $\EI_{\KB}$ and $\EI'$ are identical w.r.t. the finite ranks, and $\EI_{\KB}(w)=\tuple{\infty,i}$ for some $i$. We can have two subcases:  $\EI'(w)=\tuple{\infty,j}$ for some $j<i$, or $\EI'(w)=\tuple{f,j}$ for some $j$. The latter subcase takes to a contradiction:  it can be proved analogously to Case 1. We have to prove the first subcase.
    
    The proof is still close to the one for Case 1, simply we have to refer to the conterfactual shiftings of $\EI_{\KB}$ and $\EI'$, $\EI_{\KB\downarrow}^\infty$ and $\EI'^\infty_\downarrow$. Since $\EI_{\KB}$ and $\EI'$ are epistemic models of $\KB^\land$, $\EI_{\KB\downarrow}^\infty$ and $\EI'^\infty_\downarrow$ are epistemic models of $\K^{\land}_{\infty}$, and $\EI'^\infty_\downarrow$ is preferred to $\EI_{\KB\downarrow}^\infty$. From $\EI_{\KB\downarrow}^\infty$ and $\EI'^\infty_\downarrow$ we can extract two ranked interpretations, $\R_{\KB}^\infty$ and $\R'^\infty$ (see Definition 5), that are both epistemic models of $\K^{\land}_{\infty}$. In the construction of $\EI_{\KB}$, following Definition 12, we have used for the infinite ranks the ranked interpretation $\R_{\KB}^\infty$, that, still by Definition 12, must be the minimal ranked model of $\K^{\land}_{\infty}$. But in the present case $\R_{\KB}^\infty$ cannot be the minimal ranked model of $\K^{\land}_{\infty}$, since $\R'^\infty$ is a ranked model of $\K^{\land}_{\infty}$ that is preferred to $\R_{\KB}^\infty$. This is a contradiction.

\end{itemize}
To conclude this part, in all the possible cases if     $\EI_{\KB}$ is not a minimal epistemic model of $\KB$, we end up with a contradiction. Hence $\EI_{\KB}$ must be a minimal epistemic model of $\KB$.


The final step consists in proving that $\EI_{\KB}$ is the \emph{only} minimal epistemic model of $\KB$. The procedure is again by contradiction, assuming that $\EI_{\KB}$ is not the only minimal epistemic model of $\KB$. Hence, let $\EI'$ be another minimal epistemic model of $\KB$. The structure of the proof actually mirrors the one for the previous part, about the minimality of $\EI_{\KB}$. Again, we can distinguish two main cases.

\begin{itemize}
    \item[Case 1.] $\EI_{\KB}$ and $\EI'$ differ w.r.t. the ranking of some valuations among the finitely ranked ones. From $\EI_{\KB}$ and $\EI'$ we can extract, respectively, the ranked models $\RM$ and $\RM'$, that are both ranked models of $\KB^{\land}$. But, by Definition 12, $\RM$ is the only minimal ranked model of $\KB^{\land}$, that is, $\RM\prec\RM'$, that implies that $\EI'$ cannot be a minimal epistemic model of $\KB$.

    \item[Case 2.] $\EI_{\KB}$ and $\EI'$ do not differ w.r.t. the ranking of the  valuations that are finitely ranked in both of them, but differ w.r.t. the ranking of some valuation, $w$, that is infinitely ranked in one of the two. W.l.o.g., we assume that $w$ is  infinitely ranked in $\EI_{\KB}$. We  have two subcases:  $\EI'(w)=\tuple{\infty,j}$ for some $j$, or $\EI'(w)=\tuple{f,j}$ for some $j$. The latter subcase takes to a contradiction:  it can be proved analogously to Case 1 using the extracted ranked models. We have to prove the first subcase.
    
   The proof is still analogous to the Case 2 above. We refer to the conterfactual shiftings of $\EI_{\KB}$ and $\EI'$, $\EI_{\KB\downarrow}^\infty$ and $\EI'^\infty_\downarrow$. Since $\EI_{\KB}$ and $\EI'$ are epistemic models of $\KB^\land$ and they are identical w.r.t the finite ranks, $\EI_{\KB\downarrow}^\infty$ and $\EI'^\infty_\downarrow$ are epistemic models of $\K^{\land}_{\infty}$. From $\EI_{\KB\downarrow}^\infty$ and $\EI'^\infty_\downarrow$ we can extract two ranked interpretations, $\R_{\KB}^\infty$ and $\R'^\infty$ (see Definition 5), that are both ranked models of $\K^{\land}_{\infty}$. In the construction of $\EI_{\KB}$, following Definition 12, we have used for the infinite ranks the ranked interpretation $\R_{\KB}^\infty$, that, still by Definition 12, must be the minimal ranked model of $\K^{\land}_{\infty}$. If $\R_{\KB}^\infty$ is the minimal  ranked model of $\K^{\land}_{\infty}$,$\R_{\KB}^\infty$ is preferred to $\R'^\infty$, and by construction $\EI_{\KB}$ must be preferred to $\EI'$. This is a contradiction.

\end{itemize}
To conclude, if we assume that there is another minimal epistemic model of $\KB$ beyond $\EI_{\KB}$  we end up with a contradiction. Hence $\EI_{\KB}$ must be the only minimal epistemic model of $\KB$.

\end{proof}

\section{Decision Procedure for Minimal Entailment}\label{sect_procedure}

As mentioned at the end of Section 4, it is possible to define a procedure to decide whether a conditional is in the minimal closure of a CCKB $\KB$. Not being possible to add this material to the paper because of the length limit, we add it here for the sake of completeness. The  procedure to decide whether a conditional is in the minimal closure of a CCKB is described by  Algorithm \ref{Func:MinClosure}, and it relies on a series of propositional decision problems, hence it can be implemented on top of any propositional reasoner. In what follows we assume that the  reader has a certain familiarity with  propositional RC. 

Algorithms \ref{Func:Exceptional}-\ref{Func:RationalClosure} are known procedures (see the works by Freund \cite{Freund1998} and by Casini and Straccia \cite[Section 2]{CasiniStraccia2010}), that together define a decision procedure for rational closure (RC). As indicated in the paper (Section 2), on the semantic side the RC of a KB $\KB$ of defeasible conditionals can be characterised using the \emph{minimal ranked model} $\RM^{\KB}_{RC}$ \cite{GiordanoEtAl2015}, that is, $\alpha\twiddle\beta$ is in the RC of a set of defeasible conditionals $\C$ iff $\RM^{\KB}_{RC}\sat\alpha\twiddle\beta$ (Definition 2 in the paper).

It has been proved \cite{Freund1998,CasiniStraccia2010} that $\alpha\twiddle\beta$ is in the RC of $\C$, that is, $\RM^{\KB}_{RC}\sat\alpha\twiddle\beta$, iff $\mathtt{RationaClosure}(\C,\alpha\twiddle\beta)$ returns $\mathtt{true}$. Let's quickly explain all the involved algorithms. We refer to Figure 1 from the paper (present also in this document), that is the minimal ranked model of the KB $\mathcal{P}=\{b\twiddle f, p\twiddle\neg f, p\land \neg b\twiddle\bot\}$.

\begin{itemize}
    \item $\mathtt{Exceptional}(\C)$ takes as input a finite set $\C$ of defeasible conditionals and gives back the exceptional elements, that is, the conditionals $\alpha\twiddle \beta$ s.t. $\top\twiddle\neg\alpha$ holds in the minimal ranked model of $\C$. 
    For example, from Figure 1 you can check that the conditionals $p\twiddle\neg f$ and $p\land \neg b\twiddle\bot$ are exceptional, since all the valuations in the layer $0$ do not satisfy  $p$, and in fact $\mathtt{Exceptional}(\mathcal{P})=\{p\twiddle\neg f, p\land \neg b\twiddle\bot\}$. The procedure fully relies on propositional logic, since it uses the \emph{materialisation} of the KB $\C$ (see Section 4 in the paper).
    \item $\mathtt{ComputeRanking}(\C)$ ranks each conditional in the KB $\C$ w.r.t. its exceptionality level. $\E_{0}$ contains all the conditionals, $\E_{1}$ the exceptional ones w.r.t. $\E_{0}$, and so on... $\E_{\infty}$ contains the fixed point of the exceptionality procedure, that is, the conditionals that have antecedents that cannot be satisfied in any finitely ranked valuation in any ranked model of $\C$. $\mathtt{ComputeRanking}(\mathcal{P})$ returns $\E_{0}=\C=\{b\twiddle f, p\twiddle\neg f, p\land \neg b\twiddle\bot\}$, $\E_{1}=\{ p\twiddle\neg f, p\land \neg b\twiddle\bot\}$, $\E_{\infty}=\{  p\land \neg b\twiddle\bot\}$.
    \item $\mathtt{Rank}(\C,\alpha)$ decides the rank of a proposition, that is, the lower rank in the minimal ranked model containing a valuation that satisfies the proposition. For example, the reader can check that $\mathtt{Rank}(\mathcal{P}, \neg p)=0$, $\mathtt{Rank}(\mathcal{P},  p)=1$, $\mathtt{Rank}(\mathcal{P}, p\land f)=2$, $\mathtt{Rank}(\mathcal{P}, b\land\neg f)=\infty$, values that, for each of the propositions, corresponds exactly to the lower layer in the minimal ranked model in which there is a valuation satisfying the proposition (see Figure 1).
    \item $\mathtt{RationalClosure}(\C,\alpha\twiddle\beta)$ tells us whether $\alpha\twiddle\beta$ is in the RC of $\C$, that is, whether $\RM^\KB_{RC}\sat\alpha\twiddle\beta$. For example, $\mathtt{RationalClosure}(\mathcal{P},p\twiddle\neg f)$ is $\mathtt{true}$, since: $\mathtt{Rank}(\mathcal{P},  p)=1$,  $\E_{1}=\{ p\twiddle\neg f, p\land \neg b\twiddle\bot\}$, and $\E_{1}\cup\{p\}\models\neg f$.
\end{itemize}

 Note that {\bf all the procedures fully rely on propositional logic}.

\begin{figure}[ht]
\begin{center}
\scalebox{0.7}{
\begin{TAB}(r,1cm,0.2cm)[3pt]{|c|c|}{|c|c|c|c|}%
 {$\infty$} & {$\bar{\bird}\bar{\flies}\p$, \quad $\bar{\bird}\flies\p$} \\
 {$2$} & {$\bird\flies\p$} \\
 {$1$} & {$\bird\bar{\flies}\bar{\p}$, \quad $\bird\bar{\flies}\p$}\\ 
 {$0$} & {$\bar{\bird}\bar{\flies}\bar{\p}$, \quad $\bar{\bird}\flies\bar{\p}$, \quad $\bird\flies\bar{\p}$} \\
\end{TAB}
}
\end{center}
\vspace*{-0.2cm}
\caption{A ranked interpretation for~$\Prp=\{\bird,\flies,\peng\}$.}
\label{Figure:RankedInterpretation}
\end{figure}

\begin{algorithm}[ht]
\SetAlgoLined
\SetKwData{Left}{left}\SetKwData{This}{this}\SetKwData{Up}{up}
\SetKwFunction{Union}{Union}\SetKwFunction{FindCompress}{FindCompress}
\SetKwInOut{Input}{input}\SetKwInOut{Output}{output}
\SetKw{Return}{return}

\Input{$\text{A set  of defeasible conditionals } \C$}
\Output{$\E\subseteq\C\text{ such that }\E\text{ is exceptional w.r.t.\ }\C$}
\BlankLine
 $\E\gets\emptyset$\;
 
$\overline{\C}\gets\{\alpha\rightarrow\beta\mid \alpha\twiddle \beta\in\C\}$\;

\ForEach{$\alpha\twiddle \beta\in\C$}{	
  	\If{$\overline{\C}\entails\neg\alpha$}{$\E\gets\E\cup\{\alpha\twiddle \beta\}$}
 }
\Return{$\E$}
\caption{Exceptional($\C$)}\label{Func:Exceptional}
\end{algorithm}

\begin{algorithm}[ht]
\caption{ComputeRanking($\C$)\label{Func:Ranking}}

\SetAlgoLined
\SetKwData{Left}{left}\SetKwData{This}{this}\SetKwData{Up}{up}
\SetKwFunction{Union}{Union}\SetKwFunction{FindCompress}{FindCompress}
\SetKwInOut{Input}{input}\SetKwInOut{Output}{output}
\SetKw{Return}{return}

\Input{$\text{A set of defeasible conditionals }\C$}
\Output{An exceptionality ranking $r_{\C}$}
\BlankLine
	$i\gets 0$\;
	
	$\E_{0}\gets\C$\;
	
	$\E_{1}\gets \mathtt{Exceptional}(\E_{0}$)\;
	
	\While{$\E_{i+1}\neq\E_{i}$}{
		$i\gets i + 1$\;
		
		$\E_{i+1}\gets \mathtt{Exceptional}(\E_{i}$)\;
	}
	$\E_{\infty}\gets\E_{i}$\;
	
	$r_{\C}\gets (\E_0,\ldots,\E_{i-1}, \E_{\infty})$\;
	
\Return{$r_{\C}$}
\end{algorithm}

\begin{algorithm}[ht]
\SetAlgoLined
\SetKwData{Left}{left}\SetKwData{This}{this}\SetKwData{Up}{up}
\SetKwFunction{Union}{Union}\SetKwFunction{FindCompress}{FindCompress}
\SetKwInOut{Input}{input}\SetKwInOut{Output}{output}
\SetKw{Return}{return}

\caption{Rank($\C$, $\alpha$)\label{Func:InfiniteRank}}
\Input{\text{A set of defeasible conditionals }\C \text{, a proposition } $\alpha$}
\Output{the rank $rk_{\C}(\alpha)$ of $\alpha$}
\BlankLine
$r_{\C}=(\E_0,\ldots,\E_n,\E_{\infty})\gets \mathtt{ComputeRanking}(\C)$\;

$i\gets 0$\;

\While {$\E_{i}\entails\neg\alpha \text{ and }i\leq n$}
{$i\gets i + 1$\;
}
\If{$i\leq n$}
	{$rk_{\C}(\alpha)\gets i$\;}
\Else{\If{$\E_{\infty}\not\entails\neg\alpha$}{$rk_{\C}(\alpha)\gets (i+1)$\;}	
        \Else{$rk_{\C}(\alpha)\gets\infty$\;}
        }
\Return{$rk_{\C}(\alpha)$}
\end{algorithm}

\begin{algorithm}[ht]
\SetAlgoLined
\SetKwData{Left}{left}\SetKwData{This}{this}\SetKwData{Up}{up}
\SetKwFunction{Union}{Union}\SetKwFunction{FindCompress}{FindCompress}
\SetKwInOut{Input}{input}\SetKwInOut{Output}{output}
\SetKw{Return}{return}

\caption{RationalClosure($\C$, $\alpha\twiddle\beta$)\label{Func:RationalClosure}}
\Input{\text{A set of defeasible conditionals }\C\text{, a query }$\alpha\twiddle\beta$}
\Output{$\mathtt{true}$ if $\C\entails_{RC}\alpha\twiddle\beta$, $\mathtt{false}$ otherwise}
\BlankLine
$r_{\KB}=(\E_0,\ldots,\E_n,\E_{\infty})\gets \mathtt{ComputeRanking}(\C)$\;

$r\gets \mathtt{Rank}(\C,\alpha)$\;



\Return{$\E_{r}\cup\{\alpha\}\entails\beta$\;}
\end{algorithm}

Algorithms \ref{Func:Partition} and \ref{Func:MinClosure} are new. They define a procedure to decide minimal entailment $\miniment$, given a CCKB, and they are built on top of $\mathtt{ComputeRanking}$, $\mathtt{Rank}$, and $\mathtt{RationalClosure}$. Let us go through them:

\begin{itemize}
    \item $\mathtt{Partition}(\KB)$ takes as input a CCKB $\KB$ and identifies the set $\KB_{\infty}$ and the set of defeasible conditionals $\KB^{\land}_{\infty\downarrow}$, in a way that, we will prove, corresponds to Definition 12 in the paper. That is, $\KB_{\infty}$ is the set of conditionals which context is infinitely ranked w.r.t. $\KB^{\land}$.
    \item $\mathtt{MinimalClosure}(\KB,\alpha\twiddle_{\gamma}\beta)$ tells us whether $\alpha\twiddle_{\gamma}\beta$ is in the minimal closure of $\KB$. First the algorithm checks if $\KB$ is a consistent CCKB. Then, in case it is consistent, it checks the rank of the context $\gamma$. If the context is finite, then it checks whether the conjunctive form  $\alpha\land\gamma\twiddle\beta$ is in the RC of $\KB^{\land}$. Otherwise, it checks whether the conjunctive form $\alpha\land\gamma\twiddle\beta$ is in the RC of $\KB^{\land}_{\infty\downarrow}$.
\end{itemize}

\begin{algorithm}[ht]
\caption{Partition($\KB$)\label{Func:Partition}}

\SetAlgoLined
\SetKwData{Left}{left}\SetKwData{This}{this}\SetKwData{Up}{up}
\SetKwFunction{Union}{Union}\SetKwFunction{FindCompress}{FindCompress}
\SetKwInOut{Input}{input}\SetKwInOut{Output}{output}
\SetKw{Return}{return}

\Input{$\text{A CCKB } \KB$}
\Output{The set $\KB_{\infty}$, and the conjunctive forms $\KB^{\land}\text{ and }\KB_{\infty\downarrow}^{\land}$.}
\BlankLine
$\KB^{\land}\gets\{\alpha\land\gamma\twiddle \beta\mid\alpha\twiddle_{\gamma} \beta\in\KB\}$\;

	$r_{\KB^{\land}}=(\E_0,\ldots,\E_n,\E_{\infty})\gets \mathtt{ComputeRanking}(\KB^{\land})$\;
	
	$\KB_{\infty}\gets\emptyset$
	
	\ForEach{$\alpha\twiddle_{\gamma} \beta\in\KB$}{	
  	\If{$\mathtt{Rank} (\KB^{\land},\gamma)=\infty $}{$\KB_{\infty}\gets\KB_{\infty}\cup\{\alpha\twiddle_{\gamma}\beta\}$}
 }
	

$\mu\gets\bigwedge\{\neg \alpha\mid \alpha\twiddle\beta\in\E_{\infty}\} $\;

$\KB^{\land}_{\infty\downarrow}\gets\{\alpha\land\gamma\twiddle \beta\mid\alpha\twiddle_{\gamma} \beta\in\KB_{\infty}\}\cup\{\mu\twiddle\bot\}$\;

 \Return{$\KB_{\infty},\KB^{\land},\KB_{\infty\downarrow}^{\land}$}
\end{algorithm}




	
	




\begin{algorithm}[ht]
\caption{MinimalClosure($\KB,\alpha\twiddle_{\gamma}\beta$)\label{Func:MinClosure}}

\SetAlgoLined
\SetKwData{Left}{left}\SetKwData{This}{this}\SetKwData{Up}{up}
\SetKwFunction{Union}{Union}\SetKwFunction{FindCompress}{FindCompress}
\SetKwInOut{Input}{input}\SetKwInOut{Output}{output}
\SetKw{Return}{return}

\Input{$\text{A CCKB } \KB\text{, a query }\alpha\twiddle_{\gamma}\beta$}
\Output{$\mathtt{true}$ if $\KB\miniment\alpha\twiddle_{\gamma}\beta$, $\mathtt{false}$ otherwise}
\BlankLine
$\KB^{\land},\KB_{\infty\downarrow}^{\land}\gets \mathtt{Partition(\KB)}$\;
$\overline{\KB^{\land}}=\{(\alpha\land\gamma)\rightarrow\beta\mid\alpha\land\gamma\twiddle\beta\in\KB^{\land}\}$\;

\If{$\overline{\KB^{\land}}\entails\bot$}
    {\Return{$\mathtt{true}$}\;}
\Else{
    \If{$\mathtt{Rank}(\KB^{\land},\gamma)<\infty$}
    {\Return{$\mathtt{RationalClosure}(\KB^{\land},\alpha\land\gamma\twiddle\beta)$}\;}
    \Else{\Return{$\mathtt{RationalClosure}(\KB^{\land}_{\infty\downarrow},\alpha\land\gamma\twiddle\beta)$}\;}
    }
\end{algorithm}


We need to prove that Algorithm \ref{Func:MinClosure} is complete and correct w.r.t. minimal entailment $\miniment$. Before the main theorem, we need to prove the following lemma.

\newtheorem*{lemma1}{Lemma 1*}
\newtheorem*{theorem1}{Theorem 1*}

\begin{lemma1}\label{lemma_knowledge}
Let $\KB$ be a consistent CCKB. $\KB^\land$ is its conjunctive classical form and $\RM$ is the minimal ranked model of $\KB^\land$. Let $\mu$ be defined as in Algorithm \ref{Func:Partition}, and let $\fm{\U^f_\RM}$ be defined as in Definition 12. $\mu$ is logically equivalent to $\fm{\U^f_\RM}$.
\end{lemma1}

\begin{proof}
First, we prove that $\fm{\U^f_\RM}\entails\mu$. Let $\alpha\twiddle\beta\in\E_{\infty}$. This implies that $rk_{\KB^\land}(\alpha)=\infty$, that is, all the valuations that satisfy $\alpha$ are in  $\states{\infty}$. That is, $\U^f_\RM\subseteq\states{\neg\alpha}$ for every $\alpha$ s.t. $\alpha\twiddle\beta\in\E_{\infty}$. That implies 
\[\U^f_\RM\subseteq\bigcap\{\states{\neg\alpha}_{\RM}\mid\alpha
\twiddle\beta\in\E_{\infty}\},\]

and, consequently,

\[\fm{\U^f_\RM}\entails\mu.\]

Now we prove that $\mu\entails\fm{\U^f_\RM}$. Assume that is not the case, that is, there is a valuation $w\in\U^\infty_\RM$ s.t. $w\sat\mu$. Let $n$ be the highest finite rank in $\RM$, and consider the ranked model $\RM'$ obtained from $\RM$ just by assigning to the valuation $w$ the rank $n+1$. $\RM'$ is preferred to $\RM$, and it is easy to see that $\RM'$ is a ranked model of $\KB$: for every $\alpha\twiddle \beta\in\E_i$, for some $i<\infty$, there is a valuation in a lower rank satisfying $\alpha\land\beta$, while for every $\alpha\twiddle\beta\in\E_\infty$, $w\sat\neg\alpha$, and consequently $w$ is irrelevant w.r.t. the satisfaction of $\alpha\twiddle\beta$ by $\RM'$, since it is not in $\min\states{\alpha}^f_{\RM'}$. Hence we have that $\RM'\prec_{\KB}\RM$, against the hypothesis that $\RM$ is the minimal element in $\prec_{\KB}$. We have a contradiction, and consequently \[\mu\models\fm{\U^f_{\RM}}.\]

We can conclude that the two formulas are logically equivalent.
\end{proof}

Now we can prove the main theorem.

\begin{theorem1}\label{th_complete_procedure}
Let $\KB$ be a CCKB. $\mathtt{MinimalClosure}(\KB,\alpha\twiddle_{\gamma}\beta)$ returns $\mathtt{true}$ iff $\KB\miniment\alpha\twiddle_{\gamma}\beta$.

\end{theorem1}

\begin{proof}
We already know that the algorithms $\mathtt{Exceptional}$, $\mathtt{ ComputeRanking}$, $\mathtt{ Rank}$ and $\mathtt{RationalClosure}$ are complete and correct w.r.t. the correspondent semantic notions.

As a first step, we need to prove that the algorithm $\mathtt{Partition}$ returns the correct result, that is, the sets $\KB_{\infty}$ and $\KB^{\land}_{\infty\downarrow}$  correspond to the same sets introduced in Definition 12 in the paper.

The correctness of $\KB_{\infty}$ is guaranteed by the correctness of the algorithm $\mathtt{ComputeRanking}$.

To prove the correctness of $\KB^{\land}_{\infty\downarrow}$ we need to prove also that the defeasible conditionals $\mu\twiddle\bot$ and $\fm{\U^f_{\RM}}\twiddle\bot$ are equivalent, that is an immediate consequence of Lemma 1*. 

Now we can move to check the correctness of the algorithm $\mathtt{MinimalClosure}$. We will consider the possible cases as presented in the algorithm.

\begin{itemize}

\item[Case 1.] $\overline{\KB^{\land}}\models\bot$. 

By Corollary 4, $\overline{\KB^{\land}}\models\bot$ iff $\KB$ is inconsistent, and in such a case $\KB\miniment\alpha\twiddle_{\gamma}\beta$ for every $\alpha,\gamma,\beta$, and the algorithm is correct.

\item[Case 2.]$\overline{\KB^{\land}}\not\models\bot$ and $\mathtt{Rank}(\KB^{\land},\gamma)<\infty$.

We have to prove that in this case $\alpha\land\gamma\twiddle\beta$ is in the RC of $\KB^{\land}$ iff $\alpha\twiddle_{\gamma}\beta$ is in the minimal closure of $\KB$. 

Assume $\alpha\land\gamma\twiddle\beta$ is in the RC of $\KB^{\land}$, and let $\RM$ be the minimal ranked model of $\KB^\land$. 
That means that  $\min\states{\alpha\land\gamma}_{\RM}\subseteq\states{\beta}$.  Also, since $\mathtt{Rank}(\KB^{\land},\gamma)<\infty$, we have that $\states{\gamma}\cap\U^f_{\RM}\neq\emptyset$. By construction of the minimal epistemic model of $\KB$, $\E_{\KB}$, $\U^f_{\E_{\KB}}=\U^f_{\RM}$, and the rank of each valuation is the same. Consequently we have that $\states{\gamma}^f_{E_{\KB}}\neq\emptyset$. According to definition 4 in the paper, we have to check whether $\states{\gamma}^f_{E_{\KB}}\subseteq\states{\neg\alpha}$ or $\states{\alpha\land\gamma}_{E_{\KB}}\subseteq\states{\beta}$. From $\RM\sat\alpha\land\gamma\twiddle\beta$ we can identify two possible cases: 

\begin{itemize}
    \item $\mathtt{Rank}(\KB^{\land},\alpha)=\infty$. That implies that $\states{\gamma}^f_{E_{\KB}}\subseteq\states{\neg\alpha}$.
    \item Otherwise, in $\RM$ we have $\min\states{\alpha\land\gamma}^f\subseteq\states{\beta}$. Since $\E_{\KB}$ preserves in $\U^f_{\E_{\KB}}$ the same ranking as in   $\U^f_{\RM}$, we have        $\states{\alpha\land\gamma}^f_{E_{\KB}}\subseteq\states{\beta}$.
\end{itemize}

We can conclude that $\E_{\KB}\sat\alpha\twiddle_{\gamma}\beta$.

Now we check the opposite direction: we assume $\E_{\KB}\sat\alpha\twiddle_{\gamma}\beta$. Since $\mathtt{Rank}(\KB^{\land},\gamma)<\infty$, by Definition 12 we have that $\states{\gamma}^f_{\E_{\KB}}\neq\emptyset$. The latter, together with $\E_{\KB}\sat\alpha\twiddle_{\gamma}\beta$, implies  $\states{\alpha\land\gamma}^f_{E_{\KB}}\subseteq\states{\beta}$. By Definition 12, this condition implies that $\states{\alpha\land\gamma}^f_{\RM}\subseteq\states{\beta}$, that in turn implies $\RM\sat\alpha\land\gamma\twiddle\beta$.

\item[Case 3.]$\overline{\KB^{\land}}\not\models\bot$ and $\mathtt{Rank}(\KB^{\land},\gamma)=\infty$.

We have to prove that in this case $\alpha\land\gamma\twiddle\beta$ is in the RC of $\KB^{\land}_{\infty\downarrow}$ iff $\alpha\twiddle_{\gamma}\beta$ is in the minimal closure of $\KB$. 

Since $\mathtt{Rank}(\KB^{\land},\gamma)=\infty$, we have that $\states{\gamma}\cap\U^f_{\RM}=\emptyset$. By construction of the minimal epistemic model of $\KB$, $\E_{\KB}$, $\U^f_{\E_{\KB}}=\U^f_{\RM}$. Consequently we have that $\states{\gamma}^f_{E_{\KB}}=\emptyset$ and $\states{\gamma}_{E_{\KB}}$ have rank $\tuple{\infty,j}$ for some $j$.

Assume $\alpha\land\gamma\twiddle\beta$ is in the RC of $\KB^{\land}_{\infty\downarrow}$, and let $\RM'$ be the minimal ranked model of $\KB^\land_{\infty\downarrow}$. 
According to definition 4 in the paper, we have to check whether $\states{\alpha\land\gamma}_{E_{\KB}}\subseteq\states{\beta}$.

Assume that the thesis does not hold, that is, $\E_{\KB}\not\sat\alpha\twiddle_{\gamma}\beta$. Since $\states{\gamma}^f_{E_{\KB}}=\emptyset$, all the valuations in $\states{\gamma}_{E_{\KB}}$ are infinitely ranked, and $\E_{\KB}\not\sat\alpha\twiddle_{\gamma}\beta$ implies that there is a valuation $w$ in $\states{\alpha\land\gamma}_{E_{\KB}}$ s.t. $w\not\sat\beta$. Let $w\in\states{\tuple{\infty,i}}$ for some $i<\infty$, and $w\preceq v$ for every $v\in\states{\alpha\land\gamma}$. By Definition 12, in $\RM'$ we have  $w\in\states{i}$ for some $i<\infty$, and $w\preceq v$ for every $v\in\states{\alpha\land\gamma}$, hence we would have $\RM'\not\sat\alpha\land\gamma\twiddle\beta$, against our hypothesis that $\alpha\land\gamma\twiddle\beta$ is in the RC of $\KB^{\land}_{\infty\downarrow}$.

Now we assume $\E_{\KB}\sat\alpha\twiddle_{\gamma}\beta$. Again, since $\states{\gamma}^f_{E_{\KB}}=\emptyset$, all the valuations in $\states{\gamma}_{E_{\KB}}$ are infinitely ranked. The latter, together with Definition 12, implies that $\states{\gamma}_{E_{\KB}}=\min\states{\alpha\land\gamma}_{\RM'}$, and consequently  $\states{\gamma}_{E_{\KB}}\subseteq\states{\beta}$ implies  $\min\states{\alpha\land\gamma}_{\RM'}\subseteq\states{\beta}$. We can conclude $\RM'\sat\alpha\land\gamma\twiddle\beta$, that is, $\alpha\land\gamma\twiddle\beta$ is in the RC of $\KB^{\land}_{\infty\downarrow}$.





\end{itemize}

We have proved that in all the possible cases $\mathtt{MinimalClosure}(\KB,\alpha\twiddle_{\gamma}\beta)$ returns $\mathtt{true}$ iff $\KB\miniment\alpha\twiddle_{\gamma}\beta$.
\end{proof}

To conclude, in this section we have introduced a  procedure that decides minimal entailment and that reduces such a decision problem to a series of steps relying only on propositional logic.

\bibliographystyle{plain}
\bibliography{References}


%% file: AIJ-SituatedConditionals-Macros.tex
\newcommand{\ie}{\mbox{i.e.}}
\newcommand{\eg}{\mbox{e.g.}}
\newcommand{\cf}{\mbox{cf.}}

\newcommand{\wrt}{\mbox{w.r.t.}}
\newcommand{\st}{\mbox{s.t.}}

\newcommand{\etal}{\mbox{et al}}

\newcommand{\myskip}{\smallskip} 
\newcommand{\df}[1]{\textbf{#1}} 

\newcommand{\tuple}[1]{\langle #1 \rangle}
\newcommand{\defined}{\:\raisebox{1ex}{\scalebox{0.5}{\ensuremath{\mathrm{def}}}}\hskip-1.65ex\raisebox{-0.1ex}{\ensuremath{=}}\:}
\newcommand{\assigned}{\mathrel{\mathop:}=}

\newtheorem{definition}{Definition}[section]
\newtheorem{lemma}{Lemma}[section]
\newtheorem{theorem}{Theorem}[section] 
\newtheorem{proposition}{Proposition}[section]
\newtheorem{corollary}{Corollary}[section]
\newtheorem{example}{Example}[section]

\newcommand{\Prp}{\ensuremath{\mathcal{P}}}
\newcommand{\Lang}{\ensuremath{\mathcal{L}}} 

\newcommand{\KB}{\ensuremath{\mathcal{KB}}} 

\newcommand{\U}{\ensuremath{\mathcal{U}}} 

\newcommand{\Mod}[1]{\llbracket#1\rrbracket} 
\newcommand{\fm}[1]{\ensuremath{{\sf sent}({#1})}} 

\newcommand{\limp}{\rightarrow} 
\newcommand{\liff}{\leftrightarrow} 
\renewcommand{\bar}[1]{\overline{#1}} 

\newcommand{\sat}{\Vdash}
\newcommand{\nsat}{\not\sat}
\newcommand{\entails}{\models}

\newcommand{\twiddle}{\mathrel|\joinrel\sim}
\newcommand{\ntwiddle}{\not\twiddle}

\newcommand{\C}{\ensuremath{\mathcal{C}}}

\newcommand{\RM}{\mathscr{R}} 
\newcommand{\states}[1]{\llbracket#1\rrbracket} 
\newcommand{\R}{\ensuremath{\text{\it Rk}}}
\newcommand{\ei}{\ensuremath{\text{\it e}}}
\newcommand{\EI}{\mathscr{E}} 

\newcommand{\E}{\ensuremath{\mathcal{E}}} 

\newcommand{\miniment}{\entails_{m}}

\newcommand{\f}{\ensuremath{\mathtt{f}}}

\newcommand{\ce}[1]{\ensuremath{\mathsf{#1}}}

\newcommand{\bird}{\ensuremath{\mathsf{b}}}
\newcommand{\flies}{\ensuremath{\mathsf{f}}}
\newcommand{\peng}{\ensuremath{\mathsf{p}}}
\newcommand{\dodo}{\ensuremath{\mathsf{d}}}
\newcommand{\mc}{\ensuremath{\mathsf{mc}}}
\newcommand{\epeng}{\ensuremath{\mathsf{ep}}}
\newcommand{\rob}{\ensuremath{\mathsf{r}}}

\newcommand{\p}{\sf p}

